%% file: Oliehoek11arXiv_Update14-nocomments.tex
\documentclass{article}

\input{preamble_U14}
\usepackage{url}
\usepackage{hyperref}

\begin{document}

\title{Exploiting Agent and Type Independence in \\ Collaborative Graphical Bayesian Games}

\author[1]{Frans A.\ Oliehoek}
\affil[1]{CSAIL, Massachusetts Institute of Technology, Cambridge,
  MA, USA\\
  \protect\href{mailto:fao@csail.mit.edu}{\url{fao@csail.mit.edu}}}
\author[2]{Shimon Whiteson}
\affil[2]{Informatics Institute, University of Amsterdam, Amsterdam,
The Netherlands\\\protect\href{mailto:s.a.whiteson@uva.nl}{\url{s.a.whiteson@uva.nl}}}
\author[3]{Matthijs T.J.\ Spaan}
\affil[3]{Institute for Systems and Robotics, Instituto Superior
T{\'e}cnico, Lisbon, Portugal\\ \protect\href{mailto:mtjspaan@isr.ist.utl.pt}{\url{mtjspaan@isr.ist.utl.pt}}}

\date{}

\maketitle

\begin{abstract}
    Efficient collaborative decision making is an important challenge for multiagent systems.  Finding optimal joint actions is especially challenging when each agent has only imperfect information about the state of its environment.  Such problems can be modeled as \emph{collaborative Bayesian games} in which each agent receives private information in the form of its \emph{type}.  However, representing and solving such games requires space and computation time exponential in the number of agents.  This article introduces \emph{collaborative graphical Bayesian games (CGBGs)}, which facilitate more efficient collaborative decision making by decomposing the global payoff function as the sum of local payoff functions that depend on only a few agents.  We propose a framework for the efficient solution of CGBGs based on the insight that they posses two different types of independence, which we call \emph{agent independence} and \emph{type independence}.  In particular, we present a \emph{factor graph} representation that captures both forms of independence and thus enables efficient solutions.  In addition, we show how this representation can provide leverage in sequential tasks by using it to construct a novel method for \emph{decentralized partially observable Markov decision processes}.  Experimental results in both random and benchmark tasks demonstrate the improved scalability of our methods compared to several existing alternatives.\footnote{
        Parts of this paper serve as a basis for the papers presented at UAI'12~\citep{Oliehoek12UAI} and AAMAS'13~\citep{Oliehoek13AAMAS}.
    }\\
\textbf{keywords:} reasoning under uncertainty, decision-theoretic planning, multiagent decision making, collaborative Bayesian games, decentralized partially observable Markov decision processes

\end{abstract}

\section{Introduction}

Collaborative multiagent systems are of significant scientific interest, not
only because they can tackle inherently distributed problems, but also because
they facilitate the decomposition of problems too complex to be tackled by a
single agent
\citep{Huhns:87:DAI,Sycara:98:AIMag,Panait:05:JAAMAS,Vlassis:07:MASbook,Busoniu:08:IEEE_SMC}.
As a result, a fundamental question in artificial intelligence is how best to
design control systems for collaborative multiagent systems.  In other words,
how should teams of agents act so as to most effectively achieve common goals?
When uncertainty and many agents are involved, this question is particularly
challenging, and has not yet been answered in a satisfactory way.

A key challenge of collaborative multiagent decision making is the presence of \emph{imperfect information} \citep{Harsanyi:67-68:incompleteInfo,Kaelbling:98:AI}.  Even in single-agent settings, agents may have incomplete knowledge of the state of their environment, e.g., due to noisy sensors.  However, in multiagent settings this problem is often greatly exacerbated, as agents have access only to their own sensors, typically a small fraction of those of the complete system.  In some cases, imperfect information can be overcome by sharing sensor readings.  However, due to bandwidth limitations and synchronization issues, communication-based solutions are often brittle and scale poorly in the number of agents.

\begin{figure}
\centering
\includegraphics[width=0.8\columnwidth]{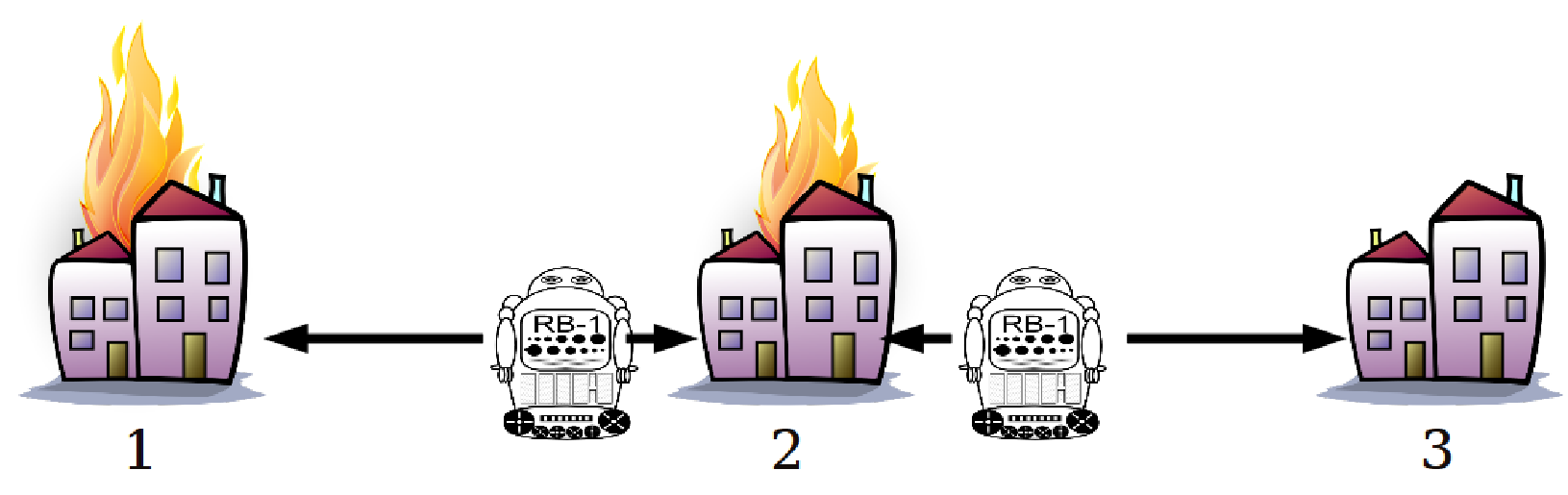}
\caption{Illustration of multiagent decision making with imperfect
  information.  Both agents are located near house 2 and know
  that it is on fire. However, each agent receives only a noisy
  observation of the single neighboring house it can observe in the
  distance.}
\label{fig:example-2agent-ff}
\end{figure}

As an example, consider the situation depicted in \fig{example-2agent-ff}.
After an emergency call by the owner of house 2, two firefighting agents are
dispatched to fight the fire.  While each agent knows there is a fire at house
2, the agents are not sure whether fire has spread to the neighboring houses.
Each agent can potentially observe flames at one of the neighboring houses
(agent 1 observes house 1 and agent 2 observes house 3) but neither has perfect
information about the state of the houses.  As a result, effective decision
making is difficult.  If agent 1 observes flames in house 1, it may be tempted
to fight fire there rather than at house 2.  However, the efficacy of doing so
depends on whether agent 2 will stay to fight fire in house 2, which in turn
depends on whether agent 2 observes flames in house 3, a fact unknown to agent
1.

\emph{Strategic games}, the traditional models of game theory, are poorly suited to modeling such problems because they assume that there is only one state, which is known to all the agents. In contrast, in the example of \fig{example-2agent-ff}, each agent has only a partial view of the state, i.e., from each agent's individual perspective, multiple states are possible. 
Such problems of multiagent decision making with imperfect information can be modeled with \emph{Bayesian games (BGs)} \citep{Harsanyi:67-68:incompleteInfo,Osborne+Rubinstein:94}. 
In a BG, each agent has a \emph{type} that specifies what private information it holds. For example, an agent's type may correspond
to an observation that it makes but the other agents do not. Before the agents select actions, their types are drawn from a distribution.
Then, the payoffs they receive depend not only on the actions they choose, but also on their types. Problems in which the agents have a common goal can be modeled as \emph{collaborative Bayesian games (CBGs)}, in which all agents share a single global payoff function. Unfortunately, solving CBGs efficiently is difficult, as both the space needed to represent the payoff function and the computation time needed to find optimal joint actions scale exponentially with the number of agents.

In this article, we introduce \emph{collaborative graphical Bayesian games (CGBGs)}, a new framework designed to facilitate more efficient collaborative decision making with imperfect information.  As in strategic games \citep{Guestrin:01:NIPS,Kok:06:JMLR}, global payoff functions in Bayesian games can often be decomposed as the sum of \emph{local payoff functions}, each of which depends on the actions of only a few agents. We call such games graphical because this decomposition can be expressed as an \emph{interaction hypergraph} that specifies which agents participate in which local payoff functions.

Our main contribution is to demonstrate how this graphical structure can be exploited to solve CGBGs more efficiently.  Our approach is based on the critical insight that CGBGs contain two fundamentally different types of independence.  Like graphical strategic games, CGBGs possess \emph{agent independence}: each local payoff function depends on only a subset of the agents.  However, we identify that  GBGs also possess \emph{type independence}: since only one type per agent is realized in a given game, the expected payoff decomposes as the sum of \emph{contributions} that depend on only a subset of types.  

We propose a \emph{factor graph} representation that captures both agent and type independence.  Then, we show how such a factor graph can be used to find optimal joint policies via \emph{nonserial dynamic programming} \citep{Bertele:72:NDPbook,Guestrin:01:NIPS}. While this approach is faster than a na\"ive alternative, we prove that its computational complexity remains exponential in the number of types. However, we also show how the same factor graph facilitates even more efficient, scalable computation of approximate solutions via $\<MP>$ \citep{Kok:06:JMLR}, a message-passing algorithm.  In particular, we prove that each iteration of max-plus is tractable for small local neighborhoods.

Alternative solution approaches for CGBGs can be found among existing techniques.  
For example, a CGBG can be converted to a multiagent influence diagram (MAID) \citep{Koller+Milch:03:GEB:MAIDs}.
However, since the resulting MAID has a single strongly connected component, the divide and conquer technique proposed 
by \citeauthor{Koller+Milch:03:GEB:MAIDs} reduces to brute-force search.
Another approach is to convert CGBGs to 
non-collaborative graphical strategic games, for which efficient
solution algorithms exist \citep{Vickrey:02:AAAI,Ortiz:03:NIPS,Daskalakis:06:EC}.  However, the conversion process essentially strips away the CGBG's type independence, resulting in an exponential increase in the worst-case size of the payoff function.  CGBGs can also be modeled as constraint optimization problems \citep{Modi:05:AI}, for which some methods implicitly exploit type independence~\citep{Oliehoek:10:AAMAS,Kumar:10:AAMAS}.  However, these methods do not explicitly identify type independence and do not exploit agent independence.

Thus, the key advantage of the approach presented in this article is the simultaneous exploitation of both agent and type independence.  We present a range of experimental results that demonstrate that this advantage leads to better scalability than several alternatives with respect to the number of agents, actions, and types.

While CGBGs model an important class of collaborative decision-making problems, they apply only to \emph{one-shot} settings, i.e., each agent needs to select only one action.  However, CGBG solution methods can also provide substantial leverage in \emph{sequential} tasks, in which agents take a series of actions over time.  We illustrate the benefits of CGBGs in such settings by using them to construct a novel method for solving  \emph{decentralized partially observable Markov
decision processes (Dec-POMDPs)}~\citep{Bernstein:02:Complexity}.  Our method extends an existing approach in which each stage of the Dec-POMDP is modeled as a CBG.  In particular, we show how approximate inference and factored value functions can be used to reduce the problem to a set of CGBGs, which can be solved using our novel approach. Additional experiments in multiple Dec-POMDP benchmark tasks demonstrate better scalability in the number of agents than several alternative methods. In particular, for a sequential version of a firefighting task as described above, we were able to scale to 1000 agents, where previous approaches to Dec-POMDPs have not been demonstrated beyond 20 agents.

The rest of this paper is organized as follows. \sec{background} provides
background by introducing collaborative (Bayesian) games and
their solution methods. \sec{CGBGs} introduces CGBGs, which capture both agent and type independence.  This section also presents solution methods that exploit such independence, analyzes their computational complexity, and empirically evaluates their performance. In \sec{DecPOMDPs}, we show that the impact of our work extends to sequential tasks by presenting and evaluating a new Dec-POMDP method based on CGBGs. \sec{relatedWork} discusses related work,
\sec{future} discusses possible directions for future work, and \sec{conclusions} concludes.

\section{Background}
\label{sec:background}
In this section, we provide background on various game-theoretic models for collaborative
decision making. We start with the well-known framework of \emph{strategic games} and
discuss their graphical counterparts, which allow for compact representations of problems
with many agents. Next, we discuss \emph{Bayesian games}, which take into account
different private information available to each agent.  These models provide a foundation for understanding 
\emph{collaborative graphical Bayesian games}, the framework we propose in Section \ref{sec:CGBGs}.

\subsection{Strategic Games}
\label{sec:SGs}
The \emph{strategic game} (SG) framework \citep{Osborne+Rubinstein:94} is probably the most studied of all game-theoretic models.  Strategic games are also called \emph{normal form games} or \emph{matrix games}, since two-agent games can be represented by matrices. We first introduce the formal model and then discuss solution methods and compact representations.

\subsubsection{The Strategic Game Model}
\label{sec:SGmodel}
In a strategic game, a set of agents participate in a one-shot interaction in which they each select an action. The outcome of the game is determined by the combination of selected actions, which leads to a payoff for each agent.
\begin{definition} 
A \emph{strategic game (SG)} is a tuple $\left\langle \agentS,\jaS,\left\langle \utA1,...\utA\nrA\right\rangle \right\rangle $,
where
\begin{itemize}
\item $\agentS=\{\agentI1,\dots,\agentI\nrA\}$ is the set of $\nrA$ agents,
\item $\jaS=\times_i \mathcal{A}_i$ is the set of joint actions $\ja=\langle\aA1,\dots,\aA\nrA\rangle$,
\item $\utA i:\jaS\rightarrow\reals$ is the payoff function of agent~$i$.
\end{itemize}
\end{definition}
This article focuses on collaborative decision making: settings in which the agents have the same goal, which is modeled by the fact that the payoffs the agents receive are identical.
\begin{definition} A \emph{collaborative strategic game (CSG)} is a strategic game in which each agent has the same payoff function:
$\forall_{i,j}\forall_{\ja}\;\utA i(\ja)=\utA j(\ja).$
\end{definition}
In the collaborative case, we drop the subscript on the payoff function and simply write $\utF$. 
CSGs are also called \emph{identical payoff games} or \emph{team games}.

\subsubsection{Solution Concepts}
\label{sec:SGsolutions}
A solution to an SG is a description of what actions each agent should take. While many solution concepts have been proposed, one of central importance is the equilibrium introduced by \citet{Nash:50}.
\begin{definition} A joint action $\ja=\left\langle \aA1,\dots,\aA i,\dots,\aA\nrA\right\rangle $
is a \emph{Nash equilibrium (NE)} if and only if 
\begin{equation}
\utA i(\left\langle \aA1,\dots,\aA i,\dots,\aA\nrA\right\rangle )
\geq\utA i(\left\langle \aA1,\dots,\aA i',\dots,\aA\nrA\right\rangle )
,\quad\forall_{i\in\agentS}
,\ \forall_{\aA i'\in\aAS i}.
\label{eq:def:NashEquilibrium}
\end{equation}
\end{definition}
Intuitively, an NE is a joint action such that no agent can improve its payoff by changing its own action.
A game may have zero, one or multiple NEs.\footnote{Nash proved that every finite game contains at least one NE if actions are
allowed to be played with a particular probability, i.e., if \emph{mixed
strategies} are allowed.} When there are multiple NEs, the concept of \emph{Pareto optimality} can help distinguish between them.
\begin{definition} A joint action $\ja$ is \emph{Pareto
optimal} if there is no other joint action $\ja'$ that specifies
at least the same payoff for every agent and a higher payoff for at
least one agent, i.e., there exists no $\ja'$ such that\begin{equation}
\forall_{\agentI i}\;\utA i(\ja')\geq\utA i(\ja)\quad\wedge\quad\exists_{\agentI i}\;\utA i(\ja')>\utA i(\ja).\label{eq:Pareto-dominated}\end{equation}
If there does exist an $\ja'$ such that \eqref{eq:Pareto-dominated}
holds, then $\ja'$ \emph{Pareto dominates} $\ja$.
\end{definition}

\begin{definition} A joint action $\ja$ is a \emph{Pareto-optimal Nash equilibrium  (PONE)} if and only if it is an NE and there is
no other \emph{$\ja'$} such that $\ja'$ is an NE \emph{and}
Pareto dominates~$\ja$.
\end{definition}
Note that this definition does not require that $\ja$ is Pareto optimal.  On the contrary, there may exist an $\ja'$ that dominates $\ja$ but is not an NE.

\subsubsection{Solving CSGs}
\label{sec:SGsolving}
In collaborative strategic games, each maximizing entry of the payoff function is a PONE. Therefore, finding a
PONE requires only looping over all the entries in $\utF$ and
selecting a maximizing one, which takes time linear in the size of
the game. However, coordination issues can arise when searching for a
PONE with a decentralized algorithm, e.g., when there are multiple
maxima. Ensuring that the agents select the same PONE can be
accomplished by imposing certain social conventions or through repeated
interactions~\citep{Boutilier:96:TARK}. In this article, we assume that the game is solved in an off-line \emph{centralized} planning phase 
and that the joint strategy is then distributed to the agents, who merely execute the actions in the on-line phase.
We focus on the design of cooperative teams of agents, for which this is a reasonable assumption.

\subsection{Collaborative Graphical Strategic Games}
\label{sec:CGSGs}
Although CSGs are conceptually easy to solve, the game description scales exponentially with the number of agents. That is, the size
of the payoff function and thus the time required for the trivial algorithm is 
$
O(\left|\aAS*\right|^{\nrA})
$,
where $\left|\aAS*\right|$ denotes the size of the largest individual action set.  This is a major obstacle in the representation and solution of SGs for large values of $\nrA.$
Many games, however, possess independence because not all agents need to
coordinate directly \citep{Guestrin:01:NIPS,Kearns:01:UAI,Kok:06:JMLR}. This idea is formalized by collaborative graphical strategic games.

\subsubsection{The Collaborative Graphical SG Model}
In collaborative graphical SGs,
the payoff function is decomposed into
\emph{local payoff functions}, each having limited \emph{scope}, i.e., only
subsets of agents participate in each local payoff function.  

\begin{definition}
\label{def:CGSG}
A \emph{collaborative graphical strategic game (CGSG)} 
is a CSG whose payoff function 
$\utF$ decomposes over a number $\nrR$ of \emph{local payoff functions} 
$\utS=\{\utI1,\dots,\utI\nrEd\}$:
\begin{equation}
    \utF(\ja)=\sum_{\ed=1}^{\nrEd}\utI \ed(\jaG{\ed}).
\label{eq:edge-based-decomposition}
\end{equation}
Each local payoff function $\utI \ed$ has scope $\agSC{\utI\ed}$, the
subset of agents that participate in $\utI \ed$. 
Here $\jaG{\ed}$ denotes the \emph{local joint action}, i.e., the profile of actions of the agents in $\agSC{\utI\ed}$.
\end{definition} 

Each local payoff component can be interpreted as a hyper-edge in an\emph{
interaction hyper-graph } $IG=\langle\agentS,\edS\rangle$ in which the nodes $\agentS$
are agents and the hyper-edges $\edS$ are local payoff functions
\citep{Nair:05:AAAI,Oliehoek:08:AAMAS}. Two (or more) agents are connected by
such a (hyper-)edge $\ed\in\edS$ if and only if they participate in the
corresponding local payoff function $\utI\ed$.%
\footnote{This constitutes an
\emph{edge-based decomposition}, which stands in contrast to \emph{agent-based
decompositions} \citep{Kok:06:JMLR}. 
We focus on edge-based decompositions because they are more
general.} 
Note that we shall abuse notation in that $\ed$ is used as an index into the set of local
payoff functions and as an element of the set of scopes.
\fig{EdgeDecomposition}a shows the interaction hyper-graph of a five-agent CGSG. 
If only two agents participate in each local payoff function, the
interaction hyper-graph reduces
to a regular graph and the framework is identical to that of \emph{coordination
graphs} \citep{Guestrin:01:NIPS,Kok:06:JMLR}.

CGSGs are also similar to \emph{graphical games}
\citep{Kearns:01:UAI,Kearns:07:GraphicalGames,Soni:07:AAMAS}. However, there is
a crucial difference in the meaning of the term `graphical'.  In CGSGs, it
indicates that the single, common payoff function $(\utF=\utA1=\dots=\utA\nrA)$
decomposes into local payoff functions, each involving subsets of agents.
However, all agents participate in the common payoff function (otherwise they
would be irrelevant to the game). In contrast, graphical games are typically
not collaborative.  Thus, in that context, the term indicates that the
\emph{individual} payoff functions $\utA1,\dots,\utA\nrA$ involve subsets of
agents.  However, these individual payoff functions do not decompose into sums
of local payoff functions.

\subsubsection{Solving CGSGs}
\label{sec:SolvingCGSGs}
Solving a collaborative graphical strategic game entails finding a maximizing joint action. However, if the representation of a
particular problem is \emph{compact}, i.e.~exponentially smaller than its non-graphical (i.e., CSG) representation,
then the trivial algorithm of
\sec{SGsolving} runs in exponential time. \emph{Non-serial dynamic
programming (NDP)} \citep{Bertele:72:NDPbook}, also known as \emph{variable elimination} \citep{Guestrin:01:NIPS,Vlassis:07:MASbook} and \emph{bucket elimination} \citep{Dechter:99:AI}, 
can find an optimal solution much faster by exploiting the structure of the problem. 
We will explain NDP in more detail in \sec{ndp}.

\begin{figure}
{
\psfrag{ag1}[cc][cc]{$\agentI 1$}
\psfrag{ag2}[cc][cc]{$\agentI 2$}
\psfrag{ag3}[cc][cc]{$\agentI 3$}
\psfrag{ag4}[cc][cc]{$\agentI 4$}
\psfrag{ag5}[cc][cc]{$\agentI 5$}
\psfrag{f1}[cc][cc]{$\utI 1$}
\psfrag{f2}[cc][cc]{$\utI 2$}
\psfrag{f3}[cc][cc]{$\utI 3$}
\psfrag{f4}[cc][cc]{$\utI 4$}
\subfloat[An interaction hyper-graph.]{
\includegraphics[width=0.45\columnwidth]{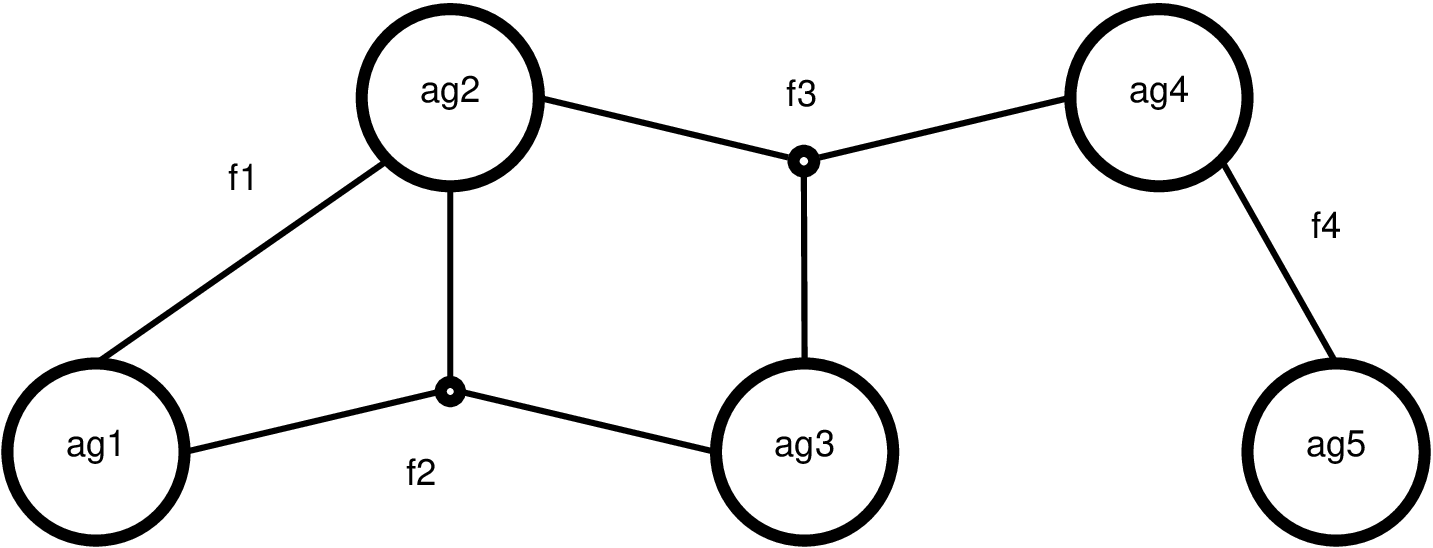}
}
\hfill
\subfloat[The corresponding factor graph.]{
\includegraphics[width=0.45\columnwidth]{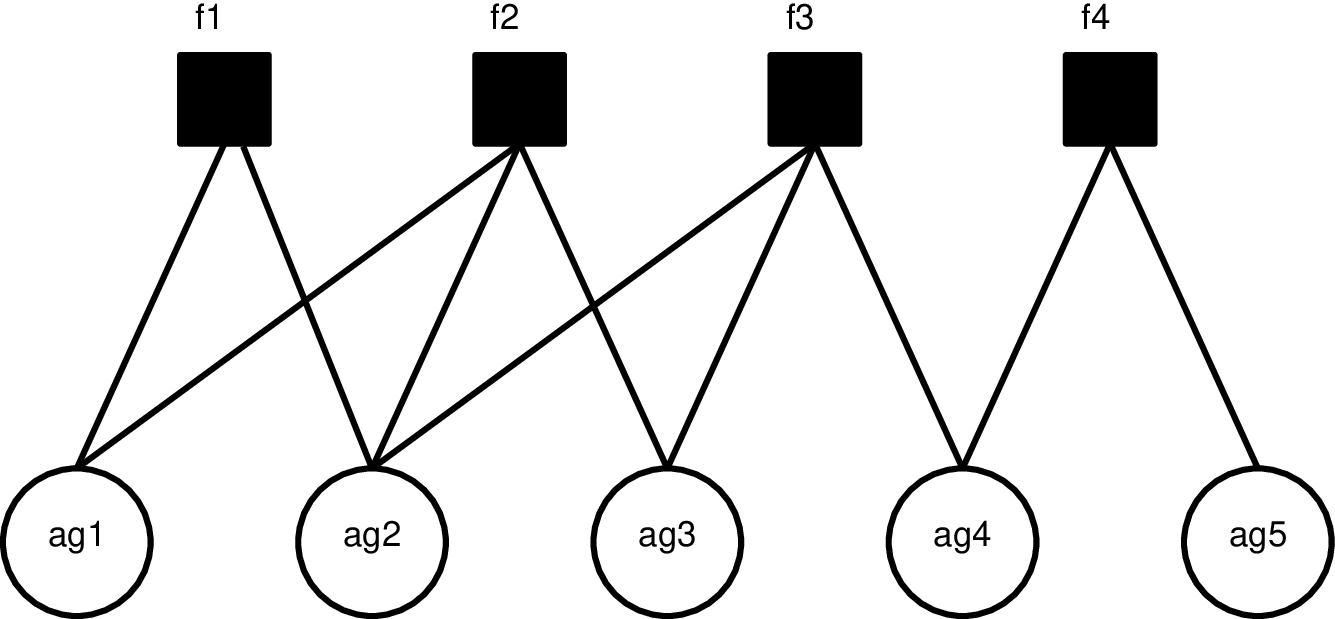}
\label{fig:CGSGFG}
}
\caption{A CGSG with five agents. In (a), each node is an agent and each hyper-edge is a local payoff function.  In (b), the circular nodes are agents and the square nodes are local payoff functions, with edges indicating in which local payoff function each agent participates.}
\label{fig:EdgeDecomposition}
}
\end{figure}

Alternatively,
$\<MP>$, a message-passing algorithm described further in \sec{max-plus}, can be applied to the interaction graph
\citep{Kok:05:robocup,Kok:06:JMLR}.   In practice, $\<MP>$ is often much faster than NDP \citep{Kok:05:robocup,Kok:06:JMLR,Farinelli:08:AAMAS,Kim:2010:DCR}. 
However, when more than two agents participate in
the same hyper-edge  (i.e., when the interaction graph is a hyper-graph),
message passing cannot be conducted on the hyper-graph itself. 
Fortunately, an interaction hyper-graph can be translated into a \emph{factor graph}
\citep{Kschischang:01:IEEE_IT,Loeliger:04:IEEE_SPM} to which $\<MP>$ is
applicable. The resulting factor graph is a bipartite graph containing one set
of nodes for all the local payoff functions and another for all the
agents.\footnote{In the terminology of factor graphs, the local payoff
functions correspond to factors and the agents to variables whose domains
are the agents' actions.} A local payoff function $\utI \ed$  is
connected to an agent $i$ if and only if $i \in \agSC{\utI\ed}$.
\fig{EdgeDecomposition} illustrates the relationship between an interaction
hyper-graph and a factor graph. 

It is also possible to convert a CGSG into a (non-collaborative) graphical SG by
combining all payoff functions in which an agent participates into one
normalized, individual payoff function.\footnote{This corresponds to converting
an edge-based representation to an agent-based representation 
\citep{Kok:06:JMLR}.}
Several methods for solving graphical SGs are then applicable
\citep{Vickrey:02:AAAI,Ortiz:03:NIPS,Daskalakis:06:EC}.  Unfortunately, 
the individual payoff functions resulting from this transformation are exponentially larger
in the worst case.

\subsection{Bayesian Games}
\label{sec:BGs}
Although strategic games provide a rich model of interactions between agents,
they assume that each agent has complete knowledge of all relevant information and
can therefore perfectly predict the payoffs that 
result from each joint action.  As such, they cannot explicitly represent cases
where agents possess private information that influences the effects of
actions.  For example, in the firefighting example depicted in
\fig{example-2agent-ff}, there is no natural way in a strategic game to
represent the fact that each agent has different information about the state of
the houses. In more complex problems with a large number of agents, modeling
private information is even more important, since assuming that so many agents
have perfect knowledge of the complete state of a complex environment is rarely
realistic.
In this section, we describe \emph{Bayesian games}, which augment the strategic game framework to explicitly model private information.  As before, we focus on the collaborative case.

\subsubsection{The Bayesian Game Model}
\label{sec:BGmodel}

A Bayesian game, also called a \emph{strategic game of imperfect information}
\citep{Osborne+Rubinstein:94} is an augmented strategic game in which
the players hold private information. The private information of agent $i$ defines its \emph{type} $\typeA i\in\typeAS i$. The payoffs the agents receive depend not only on their actions, but also on their types. Formally, a Bayesian game is defined as follows:

\begin{definition} A \emph{Bayesian game (BG)} is a tuple $\left\langle \agentS,\jaS,\jtypeS,\Pr(\jtypeS),\left\langle \utA1,...\utA\nrA\right\rangle \right\rangle $,
where
\begin{itemize}
\item $\agentS,\jaS$ are the sets of agents and joint actions as in an SG,
\item $\jtypeS=\times_{i\in\agentS}~\typeAS i$ is the set of joint types $\jtype=\langle\typeA1,\dots,\typeA\nrA\rangle$,
\item $\Pr(\jtypeS)$ is the distribution over joint types, and
\item $\utA i:\jtypeS\times\jaS\rightarrow\reals$ is the payoff function
of agent~$i$.
\end{itemize}
\end{definition}

In many problems, the types are a probabilistic function of a hidden state, i.e., based on a hidden state, there is some probability $\Pr(\jtype)$ for each joint type. This is typically the case, as in the example below, when an agent's type corresponds to a private observation it makes about such a state. However, this hidden state is not a necessary component of a BG. On the contrary, BGs can also model problems where the types correspond to intrinsic properties of the agents. For instance, in a employee recruitment game, a potential employee's type could correspond to whether or not he or she is a hard worker.

\begin{definition} A \emph{collaborative Bayesian game (CBG) }is
a Bayesian game with identical payoffs: \\
$\forall_{i,j}\forall_{\jtype}\forall_{\ja}\;\utA i(\jtype,\ja)=\utA j(\jtype,\ja). $
\end{definition}

In a strategic game, the agents simply select actions. However, in a BG, the agents can condition their actions on their types. Consequently, agents in BGs select policies instead of actions. A joint policy $\jpolBG=\left\langle \polBG1,...,\polBG\nrA\right\rangle $,
consists of individual policies $\polBG i$ for each agent~$i$. Deterministic (pure) individual policies are mappings from types to actions $\polBG i:\typeAS i\rightarrow\aAS i$, while stochastic policies map each type $\typeA i$ to a probability distribution over actions $\Pr({\aAS i})$.

\subsubsection*{Example: Two-Agent Fire Fighting}
\label{sec:2ffExample}
As an example, consider a formal model of the situation depicted in
\fig{example-2agent-ff}.  The agents each have two
actions available: agent~1 can fight fire at the first two houses ($\ffA{1}$
and $\ffA{2}$) and agent~2 can fight fire at the last two houses  ($\ffA{2}$
and $\ffA{3}$).  Both agents are located near $\ffA{2}$ and therefore know
whether it is burning. However, they are uncertain whether $\ffA{1}$  and
$\ffA{3}$ are burning or not. Each agent gets a noisy observation of one of
these houses, which defines its type. In particular, agent~1 can observe flames
($\ffOfl_1$) or not ($\ffOnf_1$) at $\ffA{1}$ and agent~2 can observe ($\ffOfl_2$) or not ($\ffOnf_2$) at $\ffA{3}$. 
\begin{table}
\noindent \begin{center}
\begin{tabular}{c||c||c|c|c|c}
& & \multicolumn{4}{c}{$\Pr(\jtype|s)$}\tabularnewline
\hline
state $s$ & $\Pr(s)$ & $\left\langle \ffOfl_1,\ffOfl_2\right\rangle $ & $\left\langle \ffOfl_1,\ffOnf_2\right\rangle $ & $\left\langle \ffOnf_1,\ffOfl_2\right\rangle $ & $\left\langle \ffOnf_1,\ffOnf_2\right\rangle $\tabularnewline
\hline
\hline
no neighbors on fire & $0.7$ & $0.01$ & $0.09$ & $0.09$ & $0.81$\tabularnewline
\hline
house 1 on fire & $0.10$ & $0.09$ & $0.81$ & $0.01$ & $0.09$\tabularnewline
\hline
house 3 on fire & $0.15$ & $0.09$ & $0.01$ & $0.81$ & $0.09$\tabularnewline
\hline
both on fire & $0.05$ & $0.81$ & $0.09$ & $0.09$ & $0.01$\tabularnewline
\hline
\hline
& $\Pr(\jtype)$ & $0.07$ & $0.15$ & $0.19$ & $0.59$\tabularnewline
\end{tabular}
\par\end{center}
\caption{The conditional probabilities of the joint types given states and the resulting distribution
over joint types for the two-agent firefighting problem.}
\label{tab:2ffTypeProbs}
\end{table}
The probability of making the correct observation is 0.9. 
\tab{2ffTypeProbs} shows the resulting probabilities of joint types conditional on
the state. The table also shows the \emph{a priori} state distribution---it is most
likely that none of the neighboring houses are on fire and $\ffA{3}$ has a slightly
higher probability of being on fire than $\ffA{1}$---and the resulting probability
distribution over joint types, computed by marginalizing over states:
$
\Pr( \jtype ) =
\sum_{s} \Pr( \jtype | s) \Pr(s).
$
\begin{table}
\begin{center}
\begin{tabular}{c||c|c|c|c}
& \multicolumn{4}{c}{Payoff of joint actions $\utF(\s,\ja)$}\tabularnewline
\hline
state $s$ & $\left\langle \ffA1,\ffA2\right\rangle $ & $\left\langle \ffA1,\ffA3\right\rangle $ & $\left\langle \ffA2,\ffA2\right\rangle $ & $\left\langle \ffA2,\ffA3\right\rangle $\tabularnewline
\hline
\hline
no neighbors on fire & $+2$ & $0$ & $+3$ & $+2$\tabularnewline
\hline
house 1 on fire & $+4$ & $+2$ & $+3$ & $+2$\tabularnewline
\hline
house 3 on fire & $+2$ & $+2$ & $+3$ & $+4$\tabularnewline
\hline
both on fire & $+4$ & $+4$ & $+3$ & $+4$\tabularnewline
\end{tabular}
\par\end{center}
\caption{The payoffs as a function of the joint actions and hidden state for the two-agent firefighting problem.}
\label{tab:2ffRewards}
\end{table}
Finally, each agent generates a +2 payoff for the team by fighting fire at a burning
house. However, payoffs are sub-additive: if both agents fight fire at the same
house (i.e., at $\ffA{2}$), a payoff of +3 is generated. 
Fighting fire at a house that is not burning does not generate any payoff. 
\tab{2ffRewards} summarizes all the possible payoffs.

These rewards can be converted to the $\utF(\jtype, \ja)$ format by computing
the conditional state probabilities $ \Pr(\s|\jtype)$ using Bayes' rule
and taking the expectation over states:
\begin{equation}
\utF(\jtype,\ja) = \sum_{\s} \utF(\s,\ja) \mult \Pr(\s|\jtype).
\end{equation}
The result is a fully specified Bayesian game whose payoff matrix is shown in \tab{2ffBG}.

\begin{table}
{\noindent
\renewcommand{\multirowsetup}{\centering}
\setlength\arrayrulewidth{\thickertableline}\arrayrulecolor{black}
\setlength{\tabcolsep}{4pt}
\newlength{\firstcol}
\setlength{\firstcol}{1.5cm}
\noindent \begin{center}
\begin{tabular}{cc|cc|cc}
& $\typeA2$ & \multicolumn{2}{c|}{$\ffOfl_2$} & \multicolumn{2}{c}{$\ffOnf_2$}\tabularnewline
$\typeA1$ & & $\ffA2$ & $\ffA3$ & $\ffA2$ & $\ffA3$\tabularnewline
\hline
\multirow{2}{\firstcol}{$\ffOfl_1$} & $\ffA1$ & $3.414$ & $2.032$ & $3.14$ & $1.22$\tabularnewline
& $\ffA2$ & $3$ & $3.543$ & $3$ & $2.08$\tabularnewline
\hline
\multirow{2}{\firstcol}{ $\ffOnf_1$ } & $\ffA1$ & $2.058$ & $1.384$ & $2.032$ & $0.079$\tabularnewline
& $\ffA2$ & $3$ & $3.326$ & $3$ & $2.047$\tabularnewline
\end{tabular}
\par\end{center}
\noindent }
\caption{The Bayesian game payoff matrix for the two-agent firefighting problem.}
\label{tab:2ffBG}
\end{table}

\subsubsection{Solution Concepts}
\label{sec:NEinBGs}

In a BG, the concept of NE is replaced by a  \emph{Bayesian
Nash equilibrium (BNE).} A profile
of policies $\jpolBG=\left\langle \polBGA1,...,\polBGA\nrA\right\rangle $
is a BNE when no agent $i$ has an incentive to switch its policy
$\polBG i$, given the policies of the other agents $\jpolBGG{\excl i}$.
This occurs when, for each agent~$i$ and each of its types~$\typeA i$,
$\polBG i$ specifies the action that maximizes its expected value.
When a Bayesian game is collaborative, the characterization of a BNE is simpler. Let the \emph{value} of a joint policy be its expected payoff:
\begin{equation}
V(\jpolBG)=
\sum_{\jtype\in\jtypeS}\Pr(\jtype)\utF(\jtype,\jpolBG(\jtype)),
\label{eq:BGIP_value}
\end{equation}
where $\jpolBG(\jtype)=\left\langle \polBG1(\typeA1),...,\polBG\nrA(\typeA\nrA)\right\rangle $
is the joint action specified by $\jpolBG$ for joint type $\jtype$.
Furthermore, let the \emph{contribution} of a joint type be:
\begin{equation}
\CoJT{\jtype}(\ja)\defas\Pr(\jtype)\utF(\jtype,\ja).
\label{eq:CoJT}
\end{equation}
The value of a joint policy  $\jpolBG$ can be interpreted as a sum of 
contributions, one for each joint type.
The BNE of a CBG maximizes a sum of such contributions.
\begin{theorem}
\label{thm:BGIPsolution}
The Bayesian
Nash equilibrium of a CBG is:
\begin{equation}
\jpolBG^{*}=
\argmax_{\jpolBG}
V(\jpolBG) =
\argmax_{\jpolBG}
\sum_{\jtype\in\jtypeS} \CoJT{\jtype}(\jpolBG(\jtype)),
\label{eq:BGIPsolution}
\end{equation}
which is a Pareto-optimal (Bayesian) Nash equilibrium (PONE).
\begin{proof}
A CBG $G$ can be reduced to a CSG $G'$ where each action of
$G'$ corresponds to a policy of $G$. Furthermore, in $G'$, a joint action $\ja'$
corresponds to a joint policy of $G$ and the payoff of a joint action
$\utF'(\ja')$ corresponds to the value of the joint policy.
As explained in \sec{SGsolutions}, a PONE
for a CSG is a maximizing entry, which corresponds to
\eq{BGIPsolution}. 
For a more formal proof, see \citep{Oliehoek:08:JAIR}.
\end{proof}\end{theorem}

\subsubsection{Solving CBGs}
\label{sec:SolvingCBGs}
Although the characterization of a PONE is simple, finding one is intractable
in general. In fact, a CBG is equivalent to a \emph{team decision problem},
which is NP-hard \citep{Tsitsiklis:85:TAC}.

Since a CBG is an instance of a (non-collaborative) BG, solution methods for the
latter apply.  A common approach is to convert a BG~$G$ to an SG~$G'$, as in the proof
of \thm{BGIPsolution}.  
An action $\aA i'$ in $G'$ correspond to a policy $\polBG i$ in~$G$, $\aA i' \defas \polBG i$,
and the
payoff of a joint action in $G'$ equals the expected payoff of the
corresponding joint BG policy $\utF'(\ja')\defas V(\jpolBG)$. 
However, since the number of policies for an
agent in a BG is exponential in the number of types, the conversion to an SG
leads to an exponential blowup in size.  When applying this procedure
in the collaborative case (i.e., to a CBG),
the result is a CSG to which the trivial algorithm applies. In effect, since
joint actions correspond to joint BG-policies, this procedure corresponds to
brute-force evaluation of all joint BG-policies.

A different approach to solving CBGs is \emph{alternating maximization (AM)}.
Starting with a random joint policy, each
agent iteratively computes a best response policy 
for each of its types.
In this way the agents hill-climb
towards a local optimum.  While the method guarantees finding an NE, it
can not guarantee finding a PONE and there is no bound on the quality of the
approximation. By starting from a specially constructed starting point, it is possible to give some guarantees on the quality of approximation \citep{Cogill:06:SIAM-JCO}. These guarantees, however, degrade exponentially as the number of agents increases.

Finally, recent work shows that the additive structure of the value function 
\eq{BGIPsolution} can be exploited by heuristic search to greatly speed up the computation of optimal solutions  \citep{Oliehoek:10:AAMAS}.
Furthermore, the point-based backup operation in a Dec-POMDP---which can be
interpreted as a special case of CBG---can be solved using state-of-the-art weighted constraint
satisfaction methods \citep{Kumar:10:AAMAS}, also providing significant increases in performance.

\section{Exploiting Independence in Collaborative Bayesian Games}
\label{sec:CGBGs}

The primary goal of this work is to find ways to efficiently solve large CBGs, i.e., CBGs with many agents, actions and types. None of the models presented in the previous section are adequate for the task.  CGSGs, by representing independence between agents, allow solution methods that scale to many agents, but they do not model private information.  In contrast, CBGs model private information but do not represent independence between agents.  Consequently, CBG solution methods scale poorly with respect to the number of agents. 

In this section, we propose a new model to address these issues. In particular, we make
three main contributions. First, \sec{independence} distinguishes between two types of independence that can occur in CBGs: in addition to the \emph{agent independence} that occurs in CGSGs, all CBGs possess \emph{type independence}, an inherent consequence of imperfect information. Second, \sec{CGBGmodel} proposes \emph{collaborative graphical Bayesian games}, a new framework that models both these types of independence. Third, \sec{sol-met} describes solution methods for this model that use a novel factor graph representation to capture both agent and type independence such that they can be exploited by NDP and $\<MP>$.  
We prove that, while the computational cost of NDP applied to such a factor graph remains exponential in the number of individual types, $\<MP>$ is tractable for small local neighborhoods.

\subsection{Agent and Type Independence}
\label{sec:independence}

As explained in \sec{CGSGs}, in many CSGs, agent interactions are sparse. The resulting independence, which we call \emph{agent independence}, has long been exploited to compactly represent and more efficiently solve games with many agents, as in the CGSG model.

While many CBGs also possess agent independence, the CBG framework provides no way to model or exploit it.  In addition, regardless of whether they have agent independence, \emph{all} CBGs possess a second kind of independence, which we call \emph{type independence}, that is an inherent consequence of imperfect information.  Unlike agent independence, type independence is captured in the CBG model and can thus be exploited.

Type independence, which we originally identified in \citep{Oliehoek:10:PhD}, is a result of the additive structure of a joint policy's value (shown in \eq{BGIPsolution}).  The key insight is that each of the contribution terms from \eq{CoJT} depends only on the action selected for some individual types. In particular, the action $\polBGA i(\typeA i)$ selected for type~$\typeA i$ of agent~$i$ affects only the contribution terms whose joint types involve~$\typeA i$. 

For instance, in the two-agent firefighting problem, one possible joint type is
$\jtype = \langle \ffOnf, \ffOnf \rangle$ (neither agent observes flames).
Clearly, the action $\polBGA1 (\ffOfl)$ that agent~$1$ selects when it has type
$\ffOfl$ (it observes flames), has no effect on the contribution of this joint
type. 

\begin{figure}
\begin{center}
{
\psfrag{a1t1}[cc][cc]{$ \ffOnf_1 $}
\psfrag{a1t2}[cc][cc]{$ \ffOfl_1 $}
\psfrag{a2t1}[cc][cc]{$ \ffOnf_2 $}
\psfrag{a2t2}[cc][cc]{$ \ffOfl_2 $}
\psfrag{t1t1}[cc][cc]{$ \CoJT{\langle \ffOnf_1, \ffOnf_2 \rangle }$}
\psfrag{t1t2}[cc][cc]{$ \CoJT{\langle \ffOnf_1, \ffOfl_2 \rangle }$}
\psfrag{t2t1}[cc][cc]{$ \CoJT{\langle \ffOfl_1, \ffOnf_2 \rangle }$}
\psfrag{t2t2}[cc][cc]{$ \CoJT{\langle \ffOfl_1, \ffOfl_2 \rangle }$}
\includegraphics[scale=0.33]{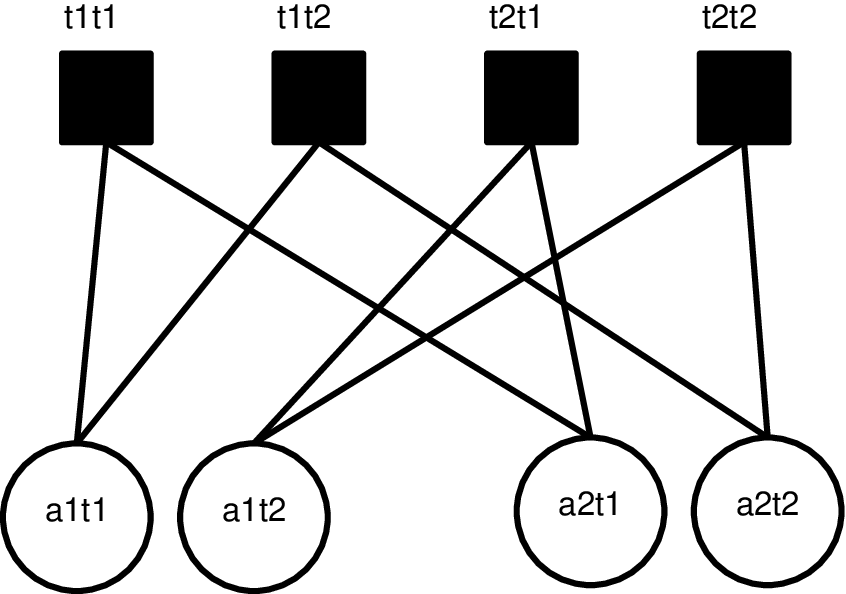}
}
\end{center}
\caption{A factor graph of the two-agent firefighting problem, illustrating the type independence inherent in CBGs. The action chosen for an individual type $\typeA i$ affects only a subset of contribution factors. For instance, the action that agent~1 selects when it has type $\ffOnf$ affects only the contribution factors $ \CoJT{\langle \ffOnf_1, \ffOnf_2 \rangle}$ and $ \CoJT{\langle \ffOnf_1, \ffOfl_2 \rangle}$ in which it has that type.
}
\label{fig:CBGstructure}
\end{figure}

As illustrated in \fig{CBGstructure}, this type of structure can also be
represented by a factor graph with one set of nodes for all the contributions
(corresponding to joint types) and another set for all the individual types of all
the agents.  Unlike the representation that results from reducing a BG to an SG played by
\emph{agent-type} combinations \citep{Osborne+Rubinstein:94}, this factor graph does not completely `flatten'
the utility function.  On the contrary, it explicitly represents the contributions of each joint
type, thereby capturing type independence. The distinction between agent and type independence is summarized in the following observation.

\begin{observation}
\label{obs:TypesOfStructure}
CBGs can possess two different types of independence:
\begin{enumerate}
\item Agent independence: the payoff function is additively decomposed over local payoff functions, each specified over only a subset
of agents, as in CGSGs.
\item Type independence:  only one type per 
agent is actually realized, leading to a value function that is additively decomposed over contributions, each specified over only a subset of
types. 
\end{enumerate}
\end{observation}

The consequence of this distinction is that neither the CGSG nor CBG model is adequate to model complex games with imperfect
information.  To scale to many agents, we need a new model that expresses (and therefore makes it possible
to exploit) both types of independence.  In the rest of this section, we propose
a model that does this and show how both types of independence can be
represented in a factor graph which, in turn, can be solved using NDP and $\<MP>$.

\subsection{The Collaborative Graphical Bayesian Game Model}
\label{sec:CGBGmodel}

A collaborative graphical Bayesian game  is a CBG whose common payoff function decomposes over a number of local payoff functions (as in a CGSG).

\begin{definition}
\label{dfn:CGBG}
A \emph{collaborative graphical Bayesian game (CGBG)} is a tuple 
$\left\langle \agentS,\jaS,\jtypeS,\mathcal{P},\mathcal{\utS}\right\rangle$,
with:
\begin{itemize}
\item $\agentS,\jaS,\jtypeS$ as in a Bayesian game, 
\item $\mathcal{P} = \{\Pr(\jtypeS_1),...,\Pr(\jtypeS_\nrR)\}$ is a set of consistent \emph{local probability distributions},
\item $\utS=\{u^1,\ldots,u^\nrR\}$ is the set of $\nrR$ \emph{local payoff functions}.
These correspond to a set $\edS$ of hyper-edges of an interaction graph, 
such that the total team payoff can (with some abuse of notation) be written as
$\utF(\theta,\ja)=\sum_{\ed\in\edS}\utI\ed(\theta_e,\jaG{\ed}).$
\end{itemize}

\end{definition} 

A CGBG is collaborative because all agents share the common payoff function $\utF(\theta,\ja)$. It is also graphical because this payoff function decomposes into a sum of local payoff functions, each of which depends on only a subset of agents.\footnote{Arguably, since all CBGs have type independence, they are in some sense already graphical, as illustrated in \fig{CBGstructure}.  However, to be consistent with the literature, we use the term graphical here to indicate agent independence.} As in CGSGs, each local payoff function $\utI \ed$ has scope $\agSC{\utI\ed}$, which can be expressed in an interaction hyper-graph $IG=\left\langle \agentS,\edS\right\rangle $ with one hyper-edge for each $\ed\in\edS$. Strictly speaking, an edge corresponds to the scope of a local payoff function, i.e., the set of agents that participate in it (as in `$\jaG\ed$'), but we will also use $\ed$ to index the sets of hyper-edges and payoff functions (as in $\utI\ed$).

Each local payoff function depends not only on the local joint action $\jaG{\ed}$, but also on the \emph{local joint type} $\theta_e$, i.e., the types of the agents in  $\ed$ (i.e., in $\agSC{\utI\ed}$). Furthermore, each local probability function $\Pr(\jtypeG{\age})$ specifies the probability of each local joint type. The goal is to maximize the expected sum of rewards:
\begin{equation}
\jpolBG^{*}  =  \argmax_{\jpolBG}
\sum_{\jtypeG{}}
\Pr(\jtypeG{})
u(\jtypeG{},\jpolBGG{}(\jtypeG{})) = 
\argmax_{\jpolBG}
\sum_{\ed\in\edS}
\sum_{\jtypeG{\age}}
\Pr(\jtypeG{\age})
\utI\ed(\jtypeG{\age},\jpolBGG{\age}(\jtypeG{\age}))
\label{eq:CGBGsolution}
\end{equation}
where $\jpolBGG{\age}(\jtypeG{\age})$ is the local joint action under policy $\jpolBGG{}$ given local joint type $\jtypeG{\age}$.

In principle, the local probability functions can be computed from the full joint probability function $\Pr(\jtypeS)$.  However, doing so is generally intractable as it requires marginalizing over the types that are not in scope.  By including $\mathcal{P}$ in the model, we implicitly assume that $\Pr(\jtypeS)$ has a compact representation that allows for efficient computation of $\Pr(\jtypeG{\age})$, e.g.,  by means of Bayesian networks \citep{Pearl:88, Bishop:06} or other graphical models. 

Not all probability distributions over joint types will admit such a compact representation.  However, those that do not will have a size exponential in the number of agents and thus cannot even be represented, much less solved, efficiently.  Thus, the assumption that these local probability functions exist is minimal in the sense that it is a necessary condition for solving the game efficiently.  Note, however, that it is not a sufficient condition.  On the contrary, the computational advantage of the methods proposed below results from the agent and type independence captured in the resulting factor graph, not the existence of local probability functions.

\subsubsection*{Example: Generalized Fire Fighting}
\label{sec:nffExample}

As an example, consider $\GFF$, which is like the two-agent firefighting problem
of \sec{2ffExample} but with $\nrA$ agents. In this version there are $\nrHouses$ 
houses and the agents are physically distributed over the area. 
Each agent gets an
observation of the $\ffnrObs$ nearest houses and may choose to fight fire at any
of the $\ffnrAcs$ nearest houses. 
For each house~$\house$, there is a local payoff
function involving the agents in its
neighborhood (i.e., of the agents that can choose to fight fire at~$\house$). These
payoff functions yield sub-additive rewards similar to those in \tab{2ffRewards}. 
The type of each agent~$i$ is defined by the $\ffnrObs$ observations it receives
from the surrounding houses: $\typeA{i} \in \{\ffOfl_i, \ffOnf_i\}^\ffnrObs $. The
probability of each type depends on the probability that the surrounding houses
are burning. As long as those probabilities can be compactly represented, the probabilities over types can be too.
\begin{figure}
{
\hfill \includegraphics[width=0.4\columnwidth]{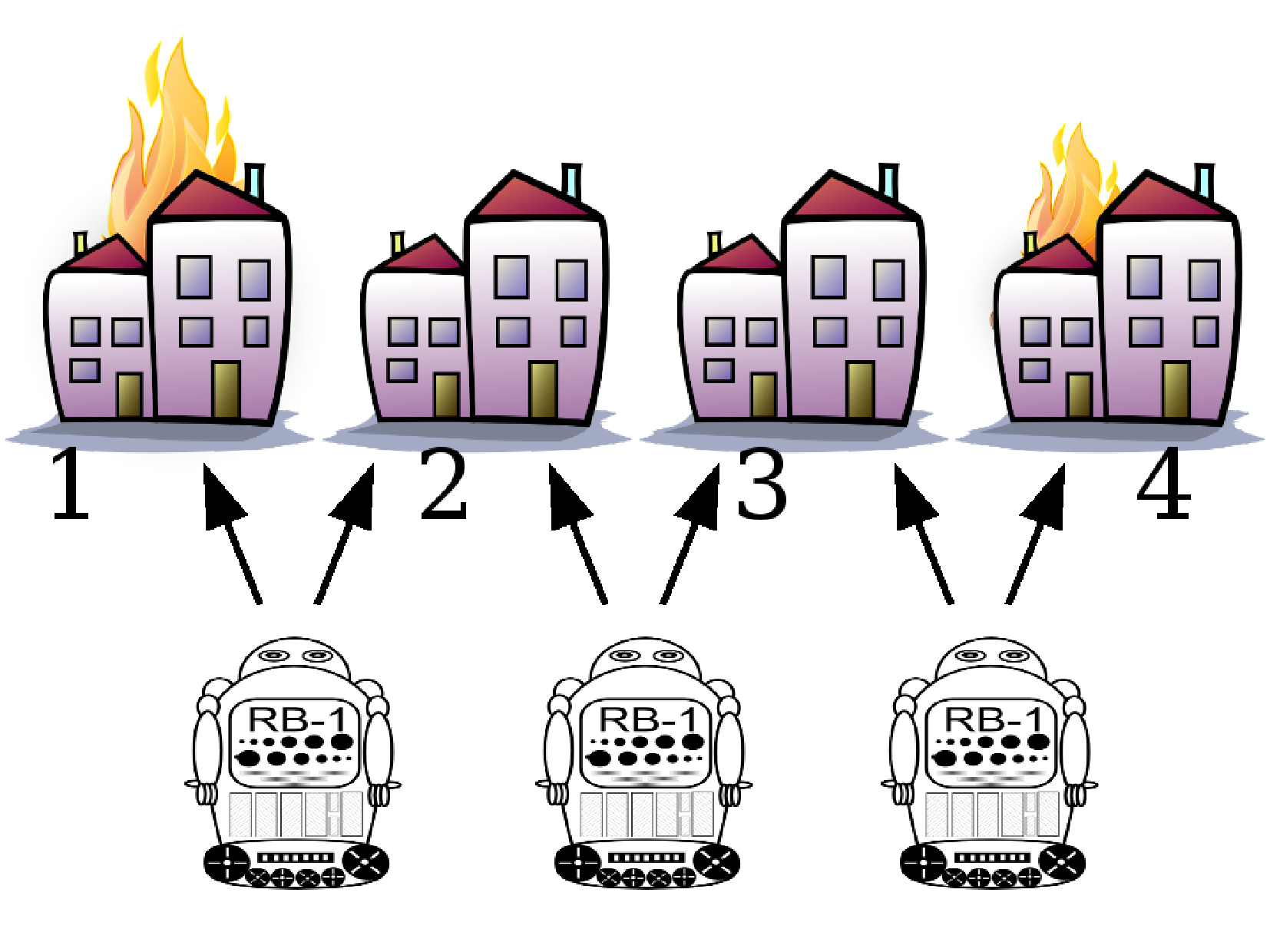} \hfill
}
\caption{
Illustration of $\GFF$ with $\nrHouses=4$ and $\nrA=3$.
}
\label{fig:nffExample}
\end{figure}
\fig{nffExample} illustrates the case where 
$\nrHouses=4$ and $\nrA=3$. Each agent can go to the $\ffnrAcs=2$
closest houses. In this problem, there are 4 local payoff functions,
one for each house, each with limited scope. Note that the payoff functions for the first and the last house
include only one agent, which means their scopes are proper subsets of the scopes
of other payoff functions (those for houses 2 and 3 respectively). Therefore, they can be included in those functions, reducing the number of local payoff functions in this example to two: one in which agents 1 and 2 participate, and one in which agents 2 and 3 participate.

\subsection{Relationship to Other Models}
\label{sec:related_models}
To provide a better understanding of the CGBG model, we elaborate on its relationship with existing models.
Just as CGSGs are related to graphical games, CGBGs are related to graphical BGs \citep{Soni:07:AAMAS}.  However, as before, there is a crucial difference in the meaning of the term `graphical'.  In CGBGs, all agents participate in a common payoff function that decomposes into local payoff functions, each involving subsets of agents. In contrast, graphical BGs are not necessarily collaborative and the individual payoff functions involve subsets of agents.  Since these individual payoff functions do not decompose, CGBGs are not a special case of GBGs but rather a unique, novel formalism.  In addition, the graphical BGs considered by \citet{Soni:07:AAMAS} make much more restrictive assumptions on the type probabilities, allowing only \emph{independent type distributions} (i.e., $\Pr(\jtypeS)$ is defined as the product of individual type probabilities $\Pr(\typeAS i)$) and assuming \emph{conditional utility independence} (i.e., the payoff of an agent~$i$  depends only on its own type, not that of other agents: $\utA{i}(\typeA i, \ja)$).

More closely related is the multiagent influence diagram (MAID) framework
that extends decision diagrams to multiagent settings \citep{Koller+Milch:03:GEB:MAIDs}.
In particular, a MAID represents a decision problem with a Bayesian network that
contains a set of chance nodes and,
for each agent, a set of decision and utility nodes. As in a CGBG, the individual payoff function
for each agent is defined as the sum of local payoffs (one for each utility node of that agent).
On the one hand, MAIDs are more general than CGBGs because they can represent non-identical payoff settings (though it would be straightforward to extend CGBGs to such problems).  On the other hand, CGBGs are more general than MAIDs since they allow any representation of the distribution over joint types (e.g., a Markov random field), as long as 
the local probability distributions can be computed efficiently.

\begin{figure}
{
\psfrag{t1}[cc][cB]{$\typeA 1$}
\psfrag{t2}[cc][cB]{$\typeA 2$}
\psfrag{t3}[cc][cB]{$\typeA 3$}
\psfrag{t12}[cc][cB]{$\jtypeG {12}$}
\psfrag{t23}[cc][cB]{$\jtypeG {23}$}
\psfrag{a1}[cc][cB]{$\aA 1$}
\psfrag{a2}[cc][cB]{$\aA 2$}
\psfrag{a3}[cc][cB]{$\aA 3$}
\psfrag{u1}[cc][cB]{$\utI {1}$}
\psfrag{u2}[cc][cB]{$\utI {2}$}
\begin{minipage}[c]{0.50\linewidth}\centering
\subfloat{%
\includegraphics[width=0.9\columnwidth]
{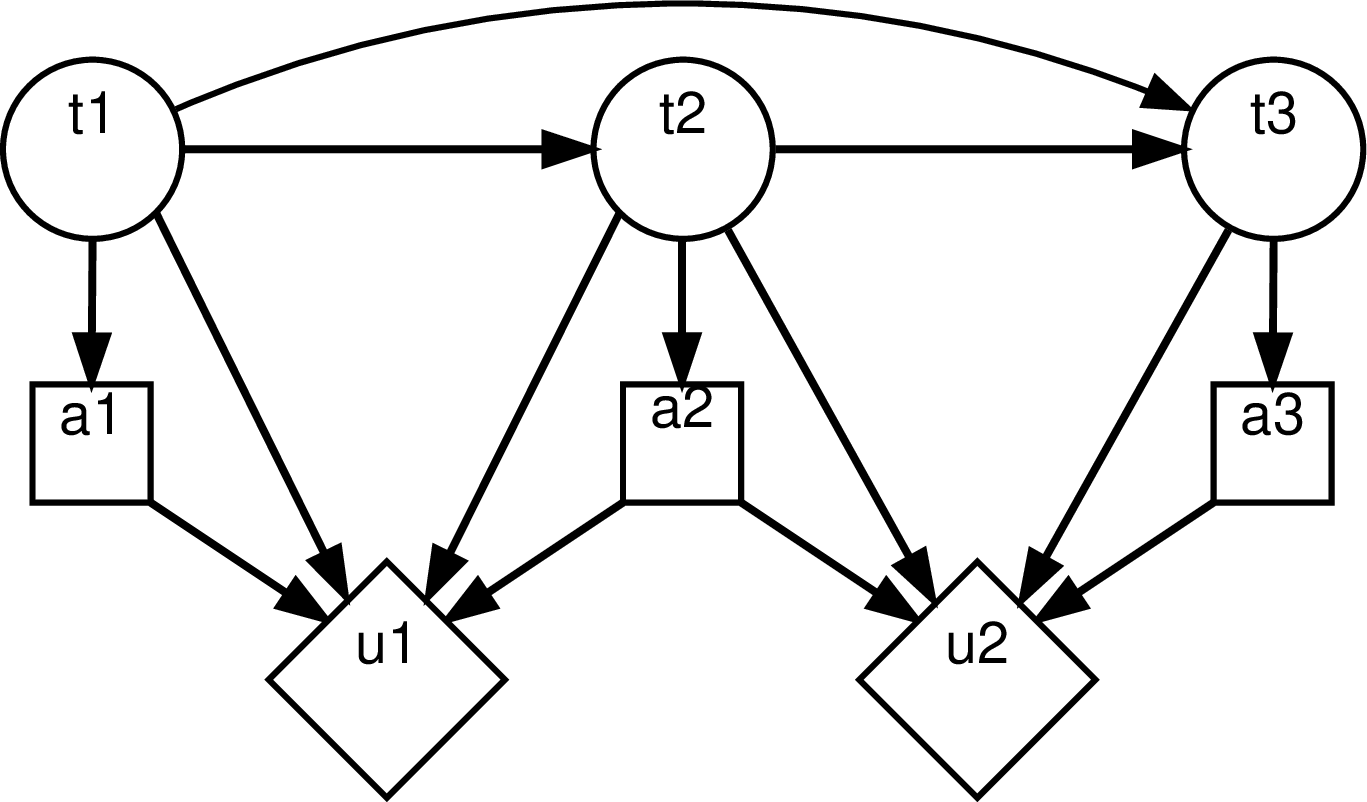}
\label{fig:MAID_rep}
}
\end{minipage}
\hfill
\begin{minipage}[c]{0.45\linewidth}\centering
\subfloat{%
\includegraphics[width=0.9\columnwidth]
{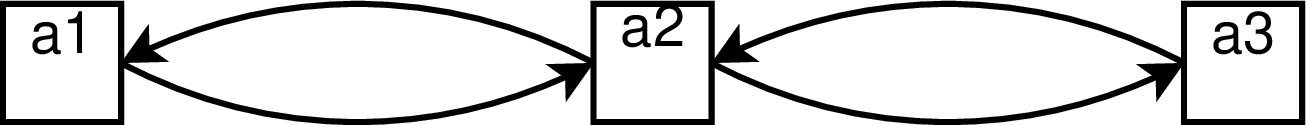} 
}
\label{fig:MAID_rg}
\end{minipage}
}
\caption{
At left, a three-agent CGBG represented as a MAID.  At right, the resulting relevance graph, which has only
one strongly connected component containing all decision variables.
}
\label{fig:MAID}
\end{figure}

A CGBG can be represented as a MAID, as illustrated in \fig{MAID}.
It is important to note that, in this MAID, both utility nodes are associated with all agents, such that each agent's goal is to optimize the sum $\utI1 + \utI2$ and the MAID is collaborative.
However, the resulting MAID's \emph{relevance graph} \citep{Koller+Milch:03:GEB:MAIDs}, which
indicates which decisions influence each other, consists of
a single strongly connected component. Consequently, the divide and conquer
solution method proposed by \citeauthor{Koller+Milch:03:GEB:MAIDs} offers no speedup over brute-force evaluation of all the joint
policies.
In the following section, we propose methods to overcome this problem and solve CGBGs efficiently.

\subsection{Solution Methods}
\label{sec:sol-met}

Solving a CGBG amounts to finding the maximizing joint policy as expressed by
\eqref{eq:CGBGsolution}. 
As mentioned in \sec{SolvingCBGs}, it is possible to convert a BG to an SG in
which the actions correspond to BG policies.  In previous work
\citep{Oliehoek:08:AAMAS}, we applied similar transformations to CGBGs, yielding
CGSGs to which all the solution methods mentioned in \sec{CGSGs} are applicable.
Alternatively, it is possible to convert to a (non-collaborative) graphical BG
\citep{Soni:07:AAMAS} and apply the proposed solution method. Under the hood,
however, this method converts to a graphical SG.

The primary limitation of all of the options mentioned above is that they
exploit only agent independence, not type independence. In fact, converting a
CGBG to a CGSG has the effect of stripping all type independence from the
model. To see why, note that type independence in a CBG 
occurs as a result of the
form of the payoff function $\utF(\jtype,\jpolBGG{}(\jtypeG{}))$.  In other
words, the payoff depends on the joint action, which in turn depends only on the
joint type that occurs.  Converting to a CSG produces a payoff function that
depends on the joint action selected for \emph{all possible} joint types,
effectively ignoring type independence.  A direct result is that the solution
methods have an exponential dependence on the number of types. 

In this section, we propose a new approach to solving CGBGs that avoids this
problem by exploiting both kinds of independence.  The main idea is to
represent the CGBG using a novel factor graph formulation that neatly captures both agent
and type independence.  The resulting factor graph can then be solved using methods such
as NDP and $\<MP>$.

To enable this factor graph formulation, we define a \emph{local contribution} as follows:
\begin{equation}
\CoJTGE {\jtypeG{\age}} {\ed} (\jaG\age)
\defas
\Pr(\jtypeG{\age})\utI\ed(\jtypeG{\age},\jaG\age)
\label{eq:CoJTGE}
\end{equation}
Using this notation, the solution of the CGBG is 
\begin{equation}
\jpolBG^{*}  =  
\argmax_{\jpolBG}
\sum_{\ed\in\edS}
\sum_{\jtypeG{\age}}
\CoJTGE{\jtypeG{\age}}{\age} (\jpolBGG{\age}(\jtypeG{\age})).
\label{eq:CGBGsolutionContris}
\end{equation}

\begin{figure}
\begin{center}
{
\psfrag{b1}[cc][tc]{ $ \polBGA{1} $ }
\psfrag{b2}[cc][tc]{ $ \polBGA{2} $ }
\psfrag{b3}[cc][tc]{ $ \polBGA{3} $ }
\psfrag{f2}[cc][Bc]{ $ \utI{1} $ }
\psfrag{f3}[cc][Bc]{ $ \utI{2} $ }

\psfrag{a1t1}[cc][Bc]{ $ \typeAI{1}{1} $ }
\psfrag{a1t2}[cc][Bc]{ $ \typeAI{1}{2} $ }
\psfrag{a2t1}[cc][Bc]{ $ \typeAI{2}{1} $ }
\psfrag{a2t2}[cc][Bc]{ $ \typeAI{2}{2} $ }
\psfrag{a3t1}[cc][Bc]{ $ \typeAI{3}{1} $ }
\psfrag{a3t2}[cc][Bc]{ $ \typeAI{3}{2} $ }
\psfrag{f2t1t1}[[cc][cc]{ $ \CoJTGE{\langle 1,1 \rangle} {1} $ }
\psfrag{f2t1t2}[[cc][cc]{ $ \CoJTGE{\langle 1,2 \rangle} {1} $ }
\psfrag{f2t2t1}[[cc][cc]{ $ \CoJTGE{\langle 2,1 \rangle} {1} $ }
\psfrag{f2t2t2}[[cc][cc]{ $ \CoJTGE{\langle 2,2 \rangle} {1} $ }
\psfrag{f3t1t1}[[cc][cc]{ $ \CoJTGE{\langle 1,1 \rangle} {2} $ }
\psfrag{f3t1t2}[[cc][cc]{ $ \CoJTGE{\langle 1,2 \rangle} {2} $ }
\psfrag{f3t2t1}[[cc][cc]{ $ \CoJTGE{\langle 2,1 \rangle} {2} $ }
\psfrag{f3t2t2}[[cc][cc]{ $ \CoJTGE{\langle 2,2 \rangle} {2} $ }
\includegraphics[scale=0.35]{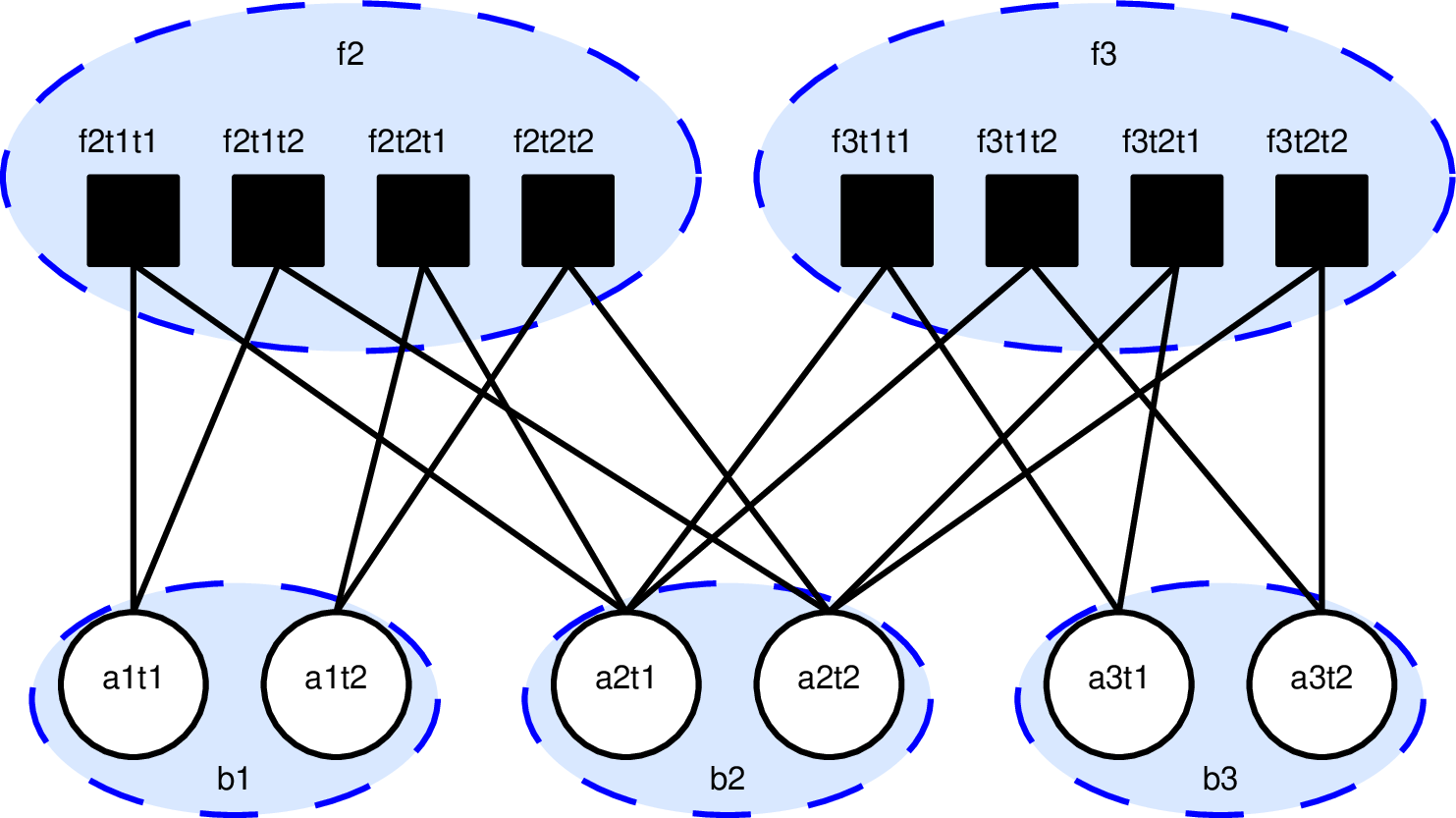}
}
\end{center}
\caption{
Factor graph for $\GFF$ with three agents, two types per agent, and two local payoff functions. Agents 1 and 2 participate in payoff function $u^1$ (corresponding to the first four contributions), while agents 2 and 3
participate in $u^2$ (corresponding to the last four contributions). The factor graph expresses both agent independence (e.g., agent 1 does not participate in $u^2$) and type independence (e.g, the action agent 1 selects when it receives observation $ \typeAI{1}{1} $ affects only the first 2 contributions).
}
\label{fig:CGBGfactorgraph}
\end{figure}

Thus, the solution corresponds to the maximum of an additively decomposed function
containing a contribution for each local joint type $\jtypeG{\age}$. This can
be expressed in a factor graph with one set of nodes for all the contributions
and another for all the individual types of all the agents. An individual type $\typeA i$ 
of an agent~$i$ is connected to a contribution $\CoJTGE{\jtypeG{\age}} {\ed}$ 
only if $i$ participates in $\utI \ed$ and ${\jtypeG{\age}}=\langle \typeA j \rangle_{j\in \agSC{\utI\ed}}$ specifies  $\typeA i$ for agent~$i$, 
as illustrated in \fig{CGBGfactorgraph}.  We refer to this graph as the 
\emph{agent and type independence (ATI) factor graph}.\footnote{In previous work, we
referred to this as the `type-action' factor graph, since its variables correspond to 
actions selected for individual types \citep{Oliehoek:10:PhD}.}
Contributions are separated, not only by the joint type
to which they apply, but also by the local payoff function to which they
contribute. Consequently, both agent and type independence are naturally
expressed.  In the next two subsections, we discuss the application of NDP and
$\<MP>$ to this factor graph formulation in order to efficiently solve CGBGs. 

\subsubsection{Non-Serial Dynamic Programming for CGBGs}
\label{sec:ndp}

Non-serial dynamic programming (NDP) \citep{Bertele:72:NDPbook}
can be used
to find the maximum configuration of a factor graph. In the \emph{forward pass}, the variables in the factor graph are eliminated one by one according to some prespecified order.  
Eliminating the $k$th variable $v$ involves collecting all the factors in which it participates 
and replacing them with a new factor $f^k$ that represents the sum of the 
removed factors, given that $v$ selects a best response. Once all variables are eliminated, the \emph{backwards pass} begins, iterating through the variables in reverse order of elimination.  Each variable selects a best response to the variables already visited, eventually yielding an optimal joint policy.

The maximum number of agents participating in a
factor encountered during NDP  is known as the induced width $w$ of the ordering. 
The following result is well-known (see for instance \citep{Dechter:99:AI}):
\begin{theorem} 
\label{thm:NDP_plain_complexity}
NDP requires exponential time and space in the induced width $w$.
\end{theorem}
Even though NDP is still exponential, for sparse problems the induced with is much smaller than the total number
of variables $V$, i.e., $w \ll V$,  leading to an exponential speed up
over naive enumeration over joint variables. 

\begin{figure}
\begin{center}
{
\tiny
\psfrag{a1t1}[cc][Bc]{ $ \typeAI{1}{1} $ }
\psfrag{a1t2}[cc][Bc]{ $ \typeAI{1}{2} $ }
\psfrag{a2t1}[cc][Bc]{ $ \typeAI{2}{1} $ }
\psfrag{a2t2}[cc][Bc]{ $ \typeAI{2}{2} $ }
\psfrag{a3t1}[cc][Bc]{ $ \typeAI{3}{1} $ }
\psfrag{a3t2}[cc][Bc]{ $ \typeAI{3}{2} $ }
\psfrag{f2t1t1}[[cc][cc]{ $ \CoJTGE{\langle 1,1 \rangle} {1} $ }
\psfrag{f2t1t2}[[cc][cc]{ $ \CoJTGE{\langle 1,2 \rangle} {1} $ }
\psfrag{f2t2t1}[[cc][cc]{ $ \CoJTGE{\langle 2,1 \rangle} {1} $ }
\psfrag{f2t2t2}[[cc][cc]{ $ \CoJTGE{\langle 2,2 \rangle} {1} $ }
\psfrag{f3t1t1}[[cc][cc]{ $ \CoJTGE{\langle 1,1 \rangle} {2} $ }
\psfrag{f3t1t2}[[cc][cc]{ $ \CoJTGE{\langle 1,2 \rangle} {2} $ }
\psfrag{f3t2t1}[[cc][cc]{ $ \CoJTGE{\langle 2,1 \rangle} {2} $ }
\psfrag{f3t2t2}[[cc][cc]{ $ \CoJTGE{\langle 2,2 \rangle} {2} $ }
\psfrag{f1}[[cc][cc]{ $ f^1 $ }
\psfrag{f2}[[cc][cc]{ $ f^2 $ }
\psfrag{f3}[[cc][cc]{ $ f^3 $ }
\includegraphics[width=.8\columnwidth]{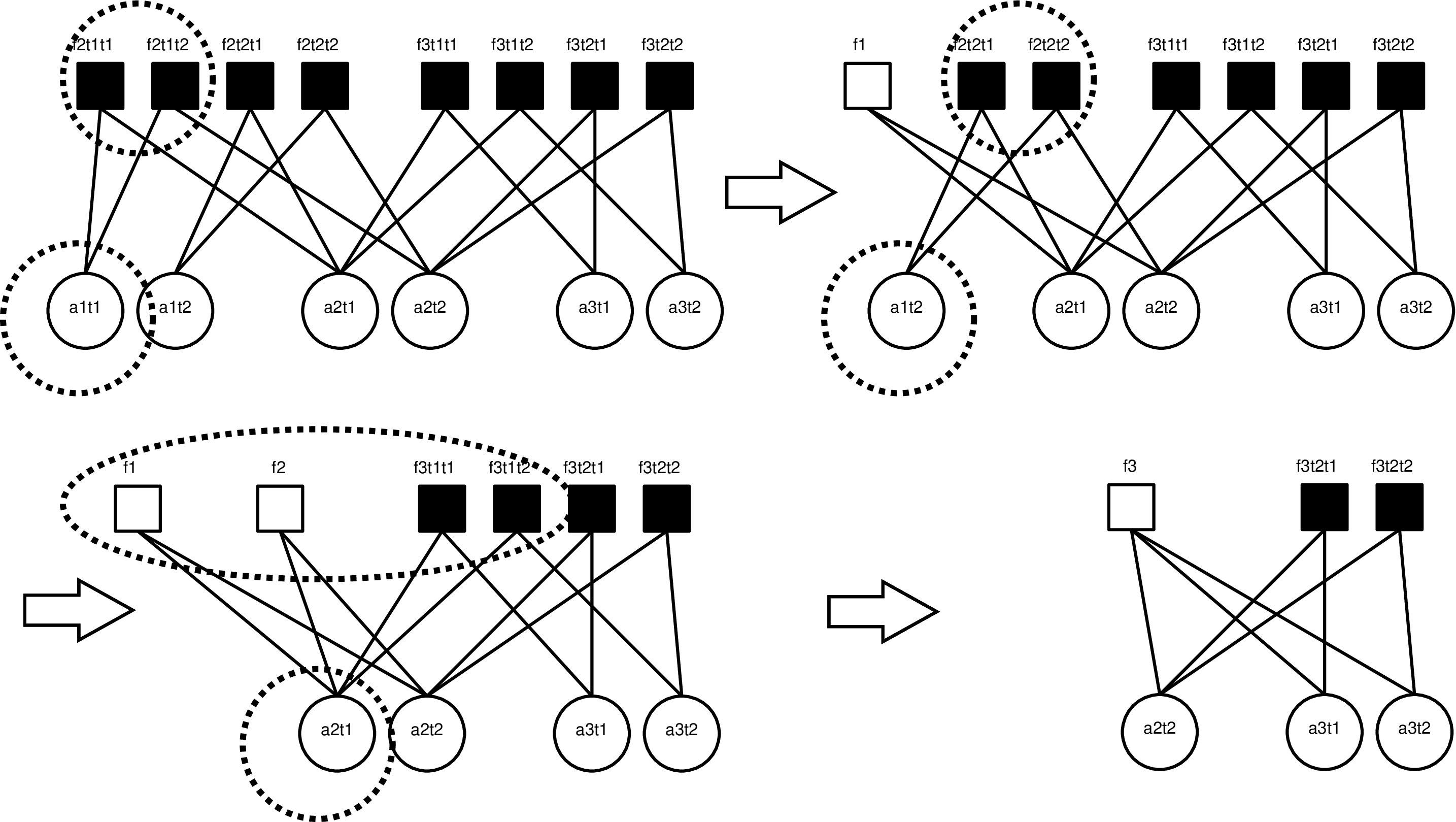}
}
\end{center}
\caption{
A few steps of NDP run on the factor graph of \fig{CGBGfactorgraph}. 
Variables are eliminated from left to right. Dotted ellipses indicate the part of the factor graph 
to be eliminated.
}
\label{fig:ATIFGNDP}
\end{figure}

In previous work \citep{Oliehoek:08:AAMAS}, we used NDP to optimally solve
CGBGs.  However, NDP was applied to the \emph{agent independence (AI)} factor graph (e.g., as  in \fig{CGSGFG}) that results from converting the CGBG to a CGSG.  Consequently, only agent independence was exploited.  In principle, we should be able to improve performance by applying NDP to the ATI factor graph introduced above, thereby exploiting both agent and type independence.
\fig{ATIFGNDP} illustrates a few steps of the resulting algorithm.

However, there are two important limitations of the NDP approach. 
First, the computational complexity is exponential in the induced width,
which in turn depends on the order in which the variables are eliminated.
Determining the optimal order 
(which \cite{Bertele:73:jCombTh} call the \emph{secondary optimization problem}) 
is NP-complete \citep{arnborg1987complexity}.  
While there are heuristics for determining the order, NDP scales poorly in practice on densely connected graphs \citep{Kok:06:JMLR}. Second, because of the particular shape that type independence
induces on the factor graph, we can establish the following:
\begin{theorem}
    \label{thm:NDP_ATI_w}
    The induced with of an ATI factor graph is lower bounded by the number of individual types: 
    $w \geq |\typeAS *| $, where $\typeAS *$ denotes the largest individual type set.
    \begin{proof}        
        Let us consider the first elimination step of NDP for an arbitrary 
        $\typeAI i m$ (i.e., an arbitrary variable).
        Now, each edge $\ed\in\edS$ (i.e., each payoff component) in which it participates induces 
        $O(| \typeAS * |^{|\ed|-1})$ contributions to which it is connected: 
        one contribution for each profile $\jtypeG{\ed \setminus  i}$ of types of the other 
        agents in $\ed$. 
        As a result, the new factor $f^1$ is connected to \emph{all} types of the
        neighbors in the interaction hyper-graph. The number of such types 
        is at least $| \typeAS * |$.         
    \end{proof}
\end{theorem}
Note that $|\typeAS *|$ is not the only term that determines $w$; the number of edges $\ed\in\edS$ in which agents participate as well as the elimination order still matter.
In particular, let $k$ denote the maximum degree of a contribution factor, i.e., the largest local scope 
$k = \max_{\ed\in\edS} | \agSC{\utI{\ed}} |$. Clearly, since there is a factor that has degree $k$, we have 
that $w\geq k$.

\begin{corollary}
    \label{cor:NDP_ATI_complexity}
    The computational complexity of NDP applied to an ATI factor graph is exponential in the number of individual types.
\begin{proof}
 This follows directly from theorems \ref{thm:NDP_plain_complexity} and \ref{thm:NDP_ATI_w}.
\end{proof}
\end{corollary}

Therefore, even given the ATI factor graph formulation, it seems unlikely that NDP will prove useful in exploiting type independence.  In particular, we hypothesize that NDP applied to the ATI factor graph will not perform significantly better than NDP on the AI factor graph. In fact, it is possible that the former performs worse than the latter. This is illustrated
in the last elimination step shown in \fig{ATIFGNDP}. A factor $f^3$ is introduced with degree $w=3$ and size $|\aAS *|^w$. In contrast, performing NDP on the AI factor graph using the `same'  left-to-right ordering has induced width $w=1$ and the size of the factors constructed is $ ( |\aAS *|^2 )^w  $ (where $2={|\typeAS *|}$). 

\subsubsection{Max-Plus for CGBGs}
\label{sec:max-plus}

In order to more effectively exploit type independence, we consider a second approach in which 
the factor graph is solved using the \<MP> message passing algorithm
\citep{Pearl:88,Wainwright:04:SC,Kok:05:robocup,Vlassis:07:MASbook}.
\<MP> was originally proposed by \cite{Pearl:88} under the name \emph{belief revision} to compute the maximum \emph{a posteriori} probability configurations in Bayesian networks. The algorithm is also known as max-product or min-sum \citep{Wainwright:04:SC}
and is a special case of the \emph{sum-product} algorithm \citep{Kschischang:01:IEEE_IT}---also referred to as
\emph{belief propagation} in probabilistic domains.
\<MP> can be implemented in either a centralized or decentralized way \citep{Kok:06:JMLR}. However, since we
assume planning takes place in a centralized off-line phase, we consider only the former.

$\<MP>$ algorithm is an appealing choice for several reasons.  First, on structured problems it has been shown to achieve excellent performance in practice \citep{Kschischang:01:IEEE_IT,Kok:06:JMLR,Kuyer:08:ECML}.  Second, unlike NDP, it is an \emph{anytime} algorithm that can provide results after each iteration
of the algorithm, not only at the end \citep{Kok:06:JMLR}.  Third, as we show below, its computational complexity is exponential only in the size of
the largest local payoff function's scope, which is fixed for a given CGBG.

At an intuitive level, $\<MP>$ works by iteratively sending messages
between the factors, corresponding to contributions, 
and variables, corresponding to (choices of actions for) types. 
These messages encode how much payoff the sender expects to be able to contribute to the
total payoff.
In particular, a message sent from a type~$i$ to a contribution $j$ encodes, for
each possible action, the payoff it expects to contribute. This is
computed as the sum of the incoming messages from other contributions $k \neq j$.
Similarly, a message sent from a contribution to a type~$i$ encodes the payoff it can contribute
conditioned on each available action to the agent with type $i$.\footnote{
For a detailed description of how the messages are computed, see
\cite[Sec.\ 5.5.3]{Oliehoek:10:PhD}.
}

$\<MP>$ iteratively passes these messages over the edges of the factor graph. Within each
iteration, the messages are sent either in parallel or sequentially with a
fixed or random ordering.  When run on an acyclic factor graph (i.e., a tree), it is
guaranteed to converge to an optimal fixed point
\citep{Pearl:88,Wainwright:04:SC}.  In cyclic factor graphs, such as those defined in
\sec{sol-met}, there are no guarantees that $\<MP>$ will
converge.\footnote{However, recent variants of the message passing approach
have slight modifications that yield convergence
guarantees~\citep{Globerson:08:NIPS07}.  Since we found that regular $\<MP>$
performs well in our experimental setting, we do not consider such variants in
this article.} However, experimental results have demonstrated that it works
well in practice even when cycles are present
\citep{Kschischang:01:IEEE_IT,Kok:06:JMLR,Kuyer:08:ECML}. This requires
normalizing the messages to prevent them from growing ever larger, 
e.g. by taking a weighted sum of the new and old messages (damping).

As mentioned above, the computational complexity of \<MP> on a CGBG is
exponential only in the size of the largest local payoff function's scope. More
precisely, we show here that this claim holds for one iteration of \<MP>.
In general, it is not possible to bound the number of iterations, since \<MP>
is not guaranteed to converge. However, by applying renormalization and/or
damping, \<MP> converges quickly in practice. Also, since \<MP> is an anytime algorithm, 
it is possible to limit the number of iterations to a constant number.

\begin{theorem} 
\label{thm:CGBG_MPcomplexity}
One iteration of $\<MP>$ run on the factor graph constructed for a
CGBG is tractable for small local neighborhoods, i.e., the only exponential dependence is in
the size of the largest local scope.
\begin{proof}
\lem{MPcomplexity} in the appendix characterizes the complexity of one iteration of \<MP>{} as
\begin{equation}
O\left( 
m^k \mult k^2 \mult l \mult F 
\right), 
\label{eq:MPcomplexity}
\end{equation} 
where, for a factor graph for a CGBG, the interpretation of the symbols is as follows:
\begin{itemize}
\item 
$m$ is the maximum number of values a type variable can take. It is given by
$m = |\aAS *|$, 
the size of the largest action set.

\item 
$k$ is the maximum degree of a contribution (factor), 
given by the largest local scope 
$k = 
\max_{\ed\in\edS} |\agSC{\utI{\ed}} |$. 

\item 
$l$
is the maximum degree of a type (variable). 
Again, each local payoff function $\ed\in\edS$ in which it participates induces $O(| \typeAS *
|^{k-1})$ contributions to which it is connected.
Let $\nrRA*$ denote the maximum number of edges in which an agent participates.
Then $l=O(\nrRA* \mult | \typeAS * |^{k-1} ) $.

\item
    $F = O(\nrR \mult |\typeAS*|^{k} )$ is the number of contributions, one for each \emph{local} joint type.
\end{itemize}
By substituting these numbers and reordering terms we get that one iteration of $\<MP>$ for a CGBG has cost:
\begin{equation}
O\left( 
|\aAS *| ^k \mult
k^2  \mult 
\nrR \nrRA* |\typeAS*|^{2k-1}  
\right), 
\label{eq:CGBG_MPcomplexity}
\end{equation} 
Thus, in the worst case, one iteration of $\<MP>$ scales 
polynomially with respect to the number of local  payoff functions $\nrR$ and the  largest sets of actions $ |\aAS *|$ and types $|\typeAS*|$. It scales exponentially in $k $.
\end{proof}
\end{theorem}

Given this result, we expect that $\<MP>$ will prove more effective than NDP at exploiting type independence.  In particular, we hypothesize that $\<MP>$ will perform better when applied to the ATI factor graph instead of the AI factor graph and that it will perform better than NDP applied to the ATI factor graph.  In the following section, we present experiments evaluating these hypotheses.

\subsection{Random CGBG Experiments}
\label{sec:experiments}

To assess the relative performance of NDP and $\<MP>$, we conduct a set of empirical evaluations on randomly generated CGBGs.  We use randomly generated games because they
allow for testing on a range of different problem parameters. 
In particular, we are
interested in the effect of scaling the number of agents $\nrA$, the
number of types $|\typeAS i|$ that each agent has, the number of
actions for each agent $|\aAS i|$, as well as the number of agents
involved in each payoff function, $|\agSC{\ed}|$.  We assume each
payoff function has an equal number of agents and refer to
this property as $k = \max_{\ed\in\edS} | \agSC{\ed} |
$, as in \thm{CGBG_MPcomplexity}.  

Furthermore, we empirically evaluate the influence of exploiting both agent and type 
independence versus exploiting only one or the other.  We do so by running both NDP and
\<MP>{} on the agent-independence (AI) factor graph (\fig{CGSGFG}), the
type-independence (TI) factor graph (\fig{CBGstructure}), and the agent and type
independence (ATI) factor graph (\fig{CGBGfactorgraph}).

These experiments serve three main purposes.  First, they empirically validate Theorems \ref{thm:NDP_ATI_w} and \ref{thm:CGBG_MPcomplexity}, confirming the difference in computational complexity between NDP and $\<MP>$.  Second, they quantify the magnitude of the difference in runtime performance between these two methods.  Third, they shed light on the quality of the solutions found by $\<MP>$, which is guaranteed to be optimal only on tree-structured graphs.

\subsubsection{Experimental Setup}
\label{sec:experimental-setup}

For simplicity, when generating CGBGs,
we assume that 1) the scopes of the local payoff functions have the same size $k$,
2) the individual action sets have the same size 
$|\aAS i|$, and 3) the individual type sets have the same size $|\typeAS i|$.
For each set of parameters,
we generate $1,000$ CGBGs on which to test.

Each game is generated following a procedure similar to that used by \citet{Kok:06:JMLR} for generating CGSGs.\footnote{The main difference is in the termination
condition: we stop adding edges when the interaction graph is fully connected
instead of adding a pre-defined number of edges.}
We start with a set of $\nrA$ agents with no local payoff functions
defined, i.e., they form an interaction hypergraph with no edges. 
As long as the interaction hypergraph is not yet connected, i.e., there does not exist a path between every pair of agents, we add a local payoff function involving $k$ agents.

\begin{figure}
    \centering
    \subfloat[$\nrA = 5$, $k=2$.]{

       \includegraphics[width=0.12\textwidth]{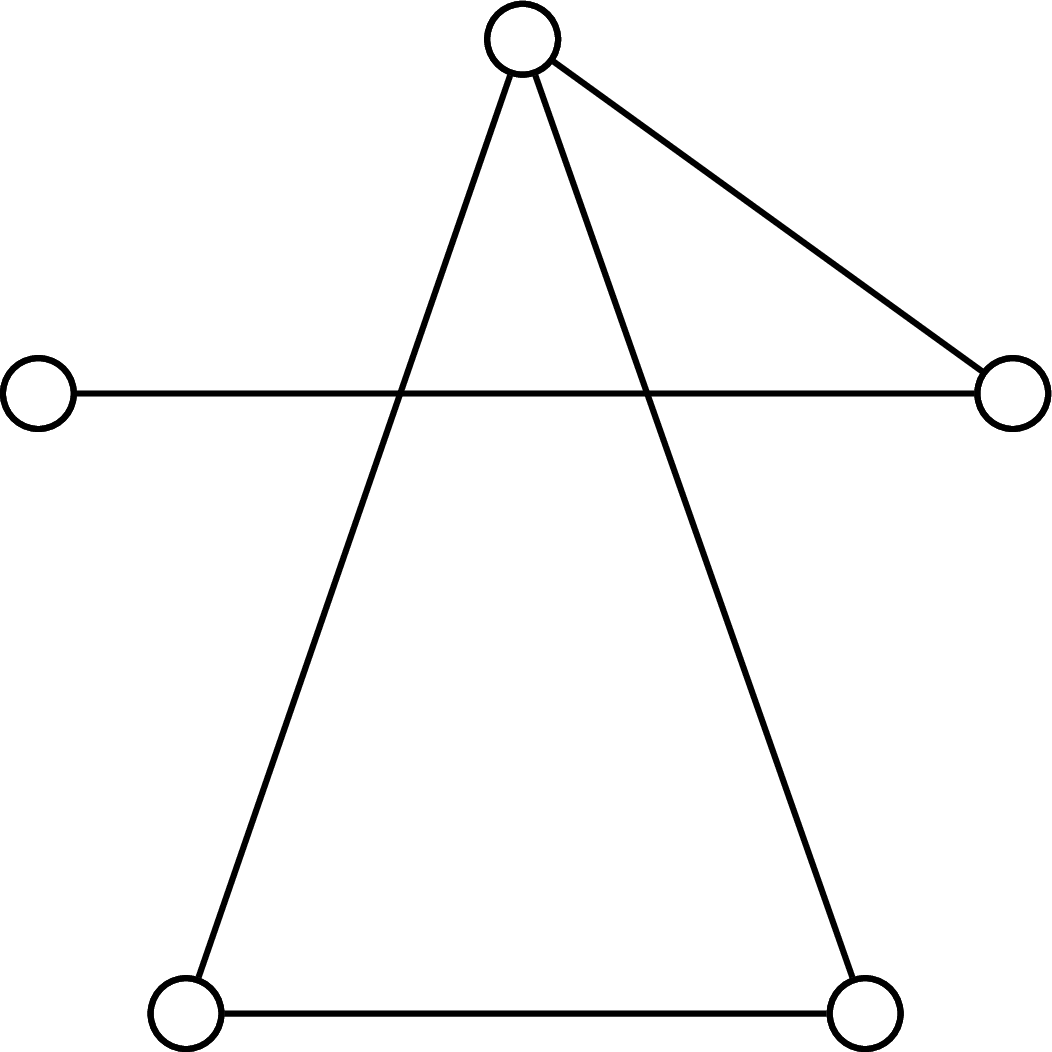}\hspace{0.01\textwidth}
       \includegraphics[width=0.12\textwidth]{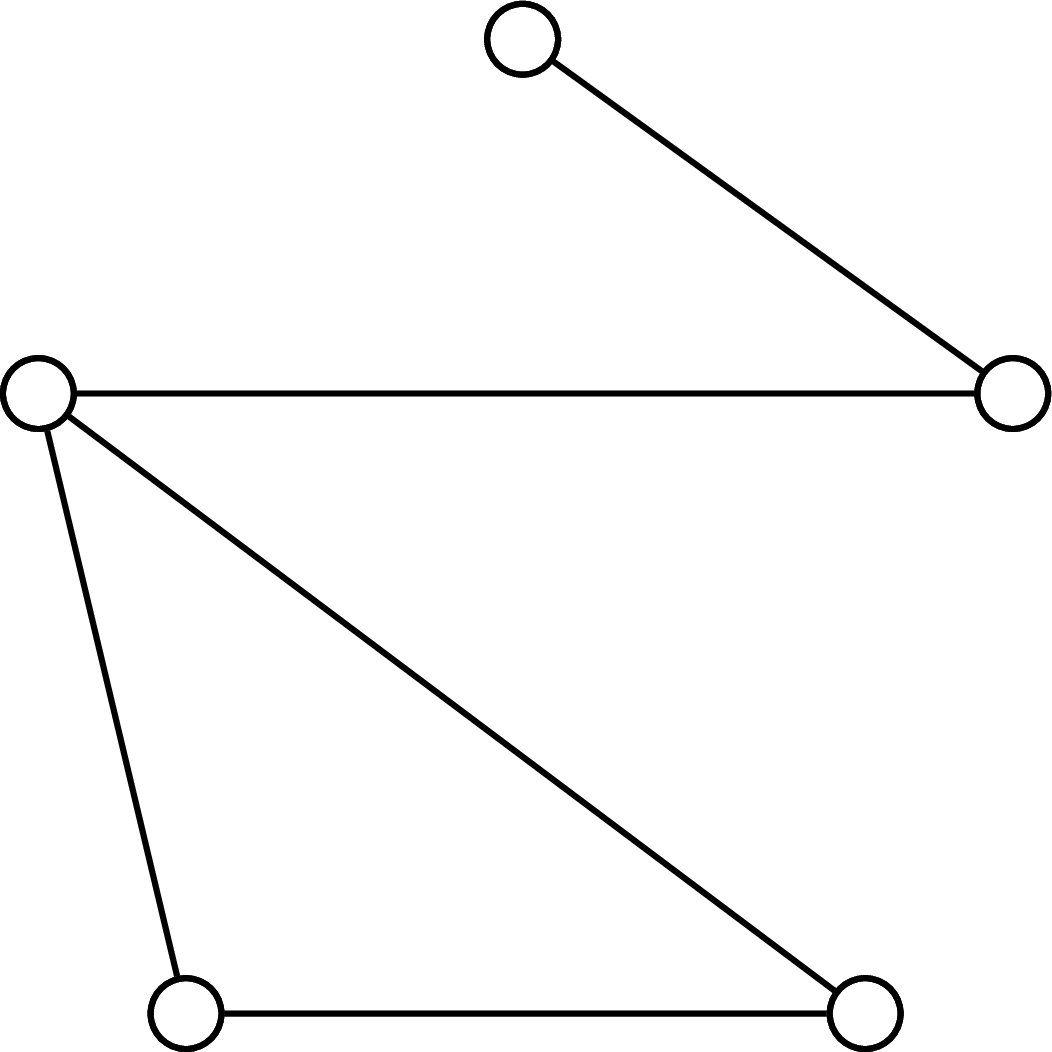}\hspace{0.01\textwidth}
       \includegraphics[width=0.12\textwidth]{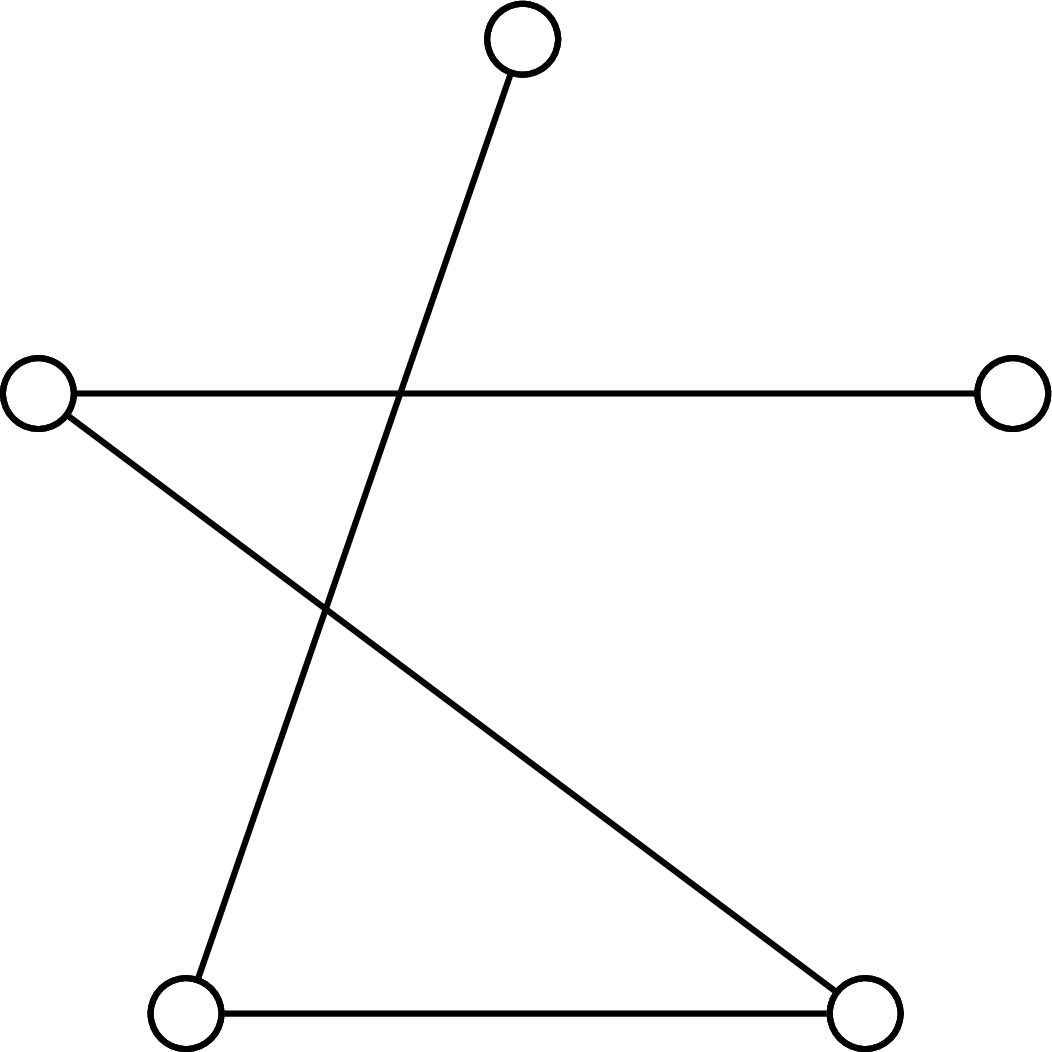}
    }\hfill
    \subfloat[$k=3$, left:$\nrA=5$, right:$\nrA=8$.]{
        \hspace{0.01\textwidth}

        \includegraphics[width=0.12\textwidth]{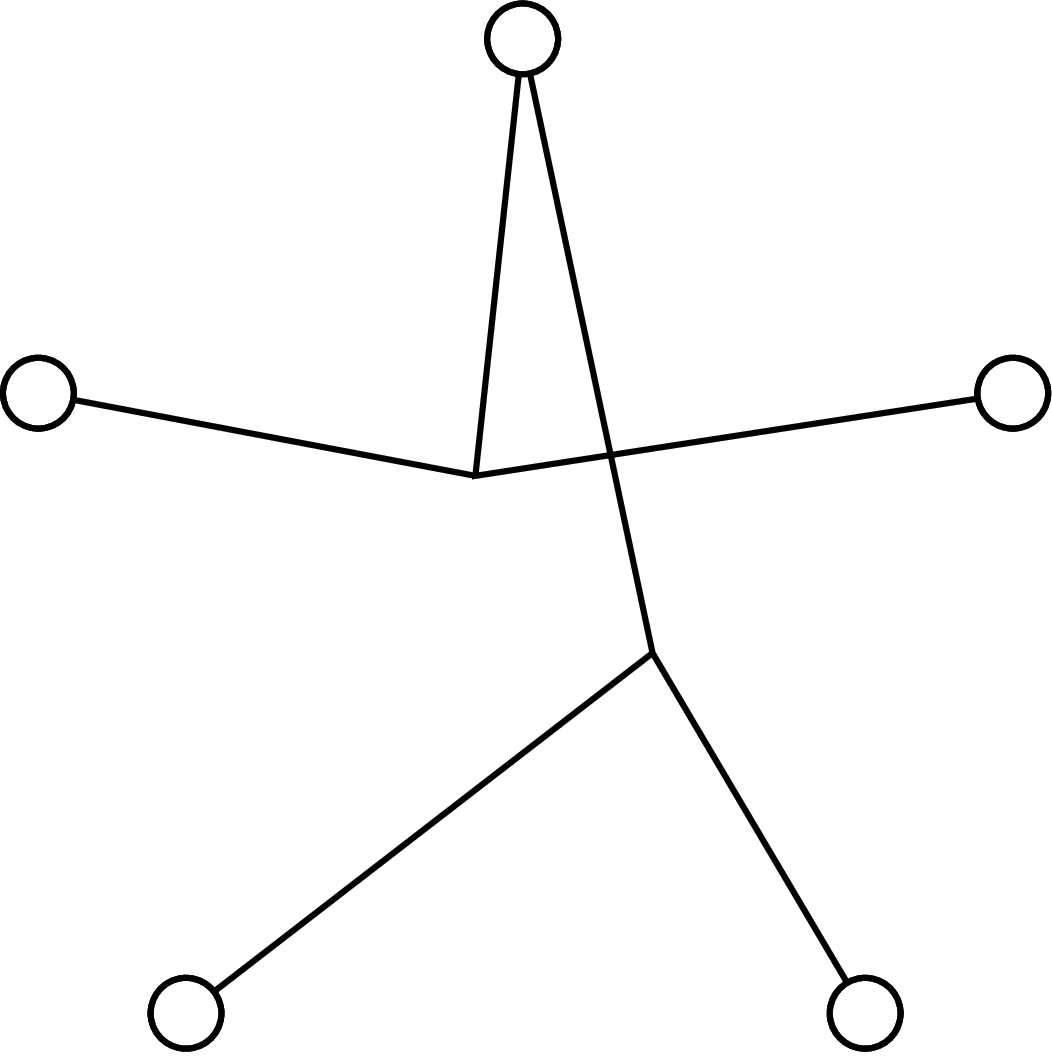}\hspace{0.01\textwidth}
        \includegraphics[width=0.12\textwidth]{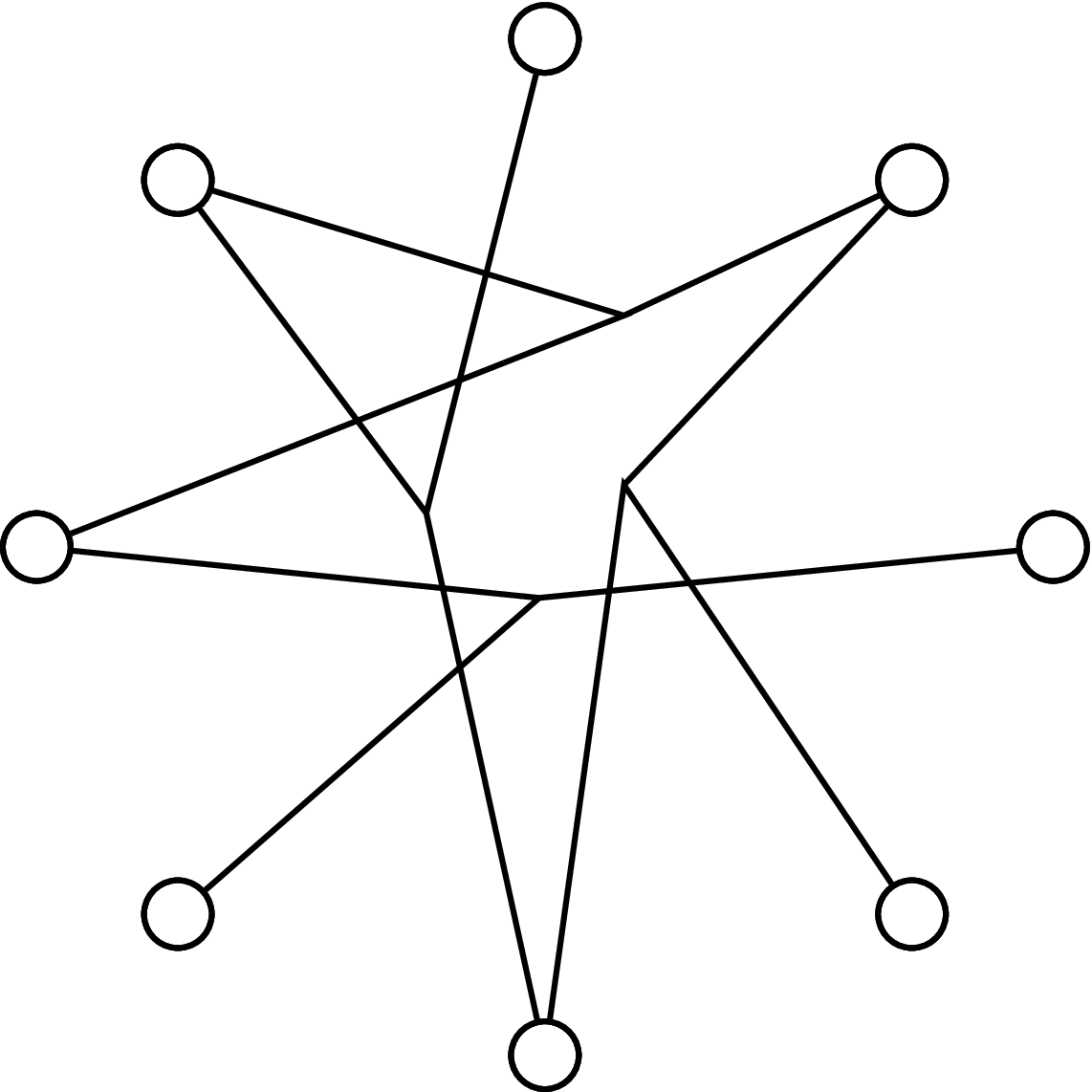}\label{fig:exampleGraphs:3agPerEdge}
        \hspace{0.01\textwidth}
    }\hfill
    \subfloat[$\nrA = 8$, $k=2$.]{

        \includegraphics[width=0.12\textwidth]{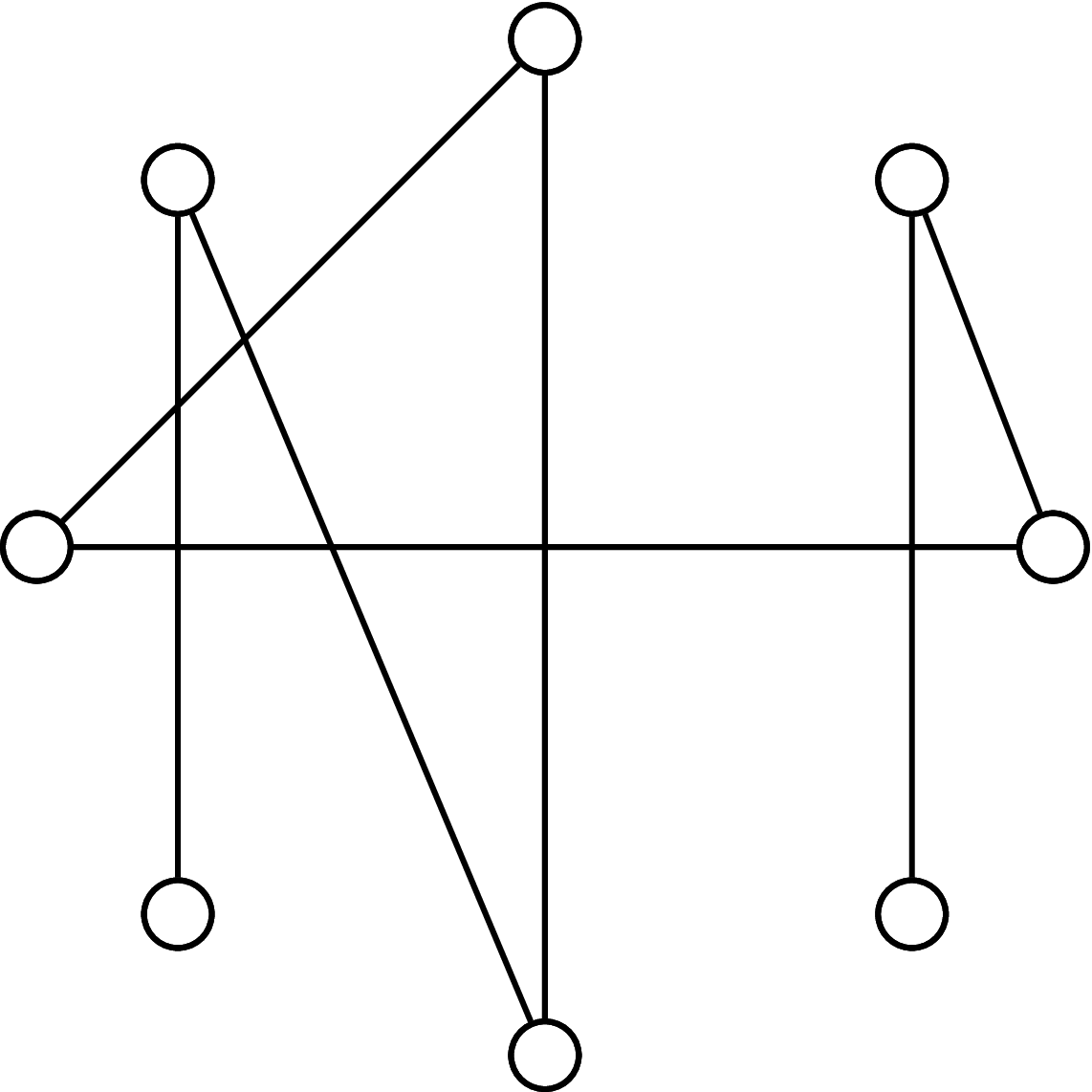}\hspace{0.01\textwidth}
        \includegraphics[width=0.12\textwidth]{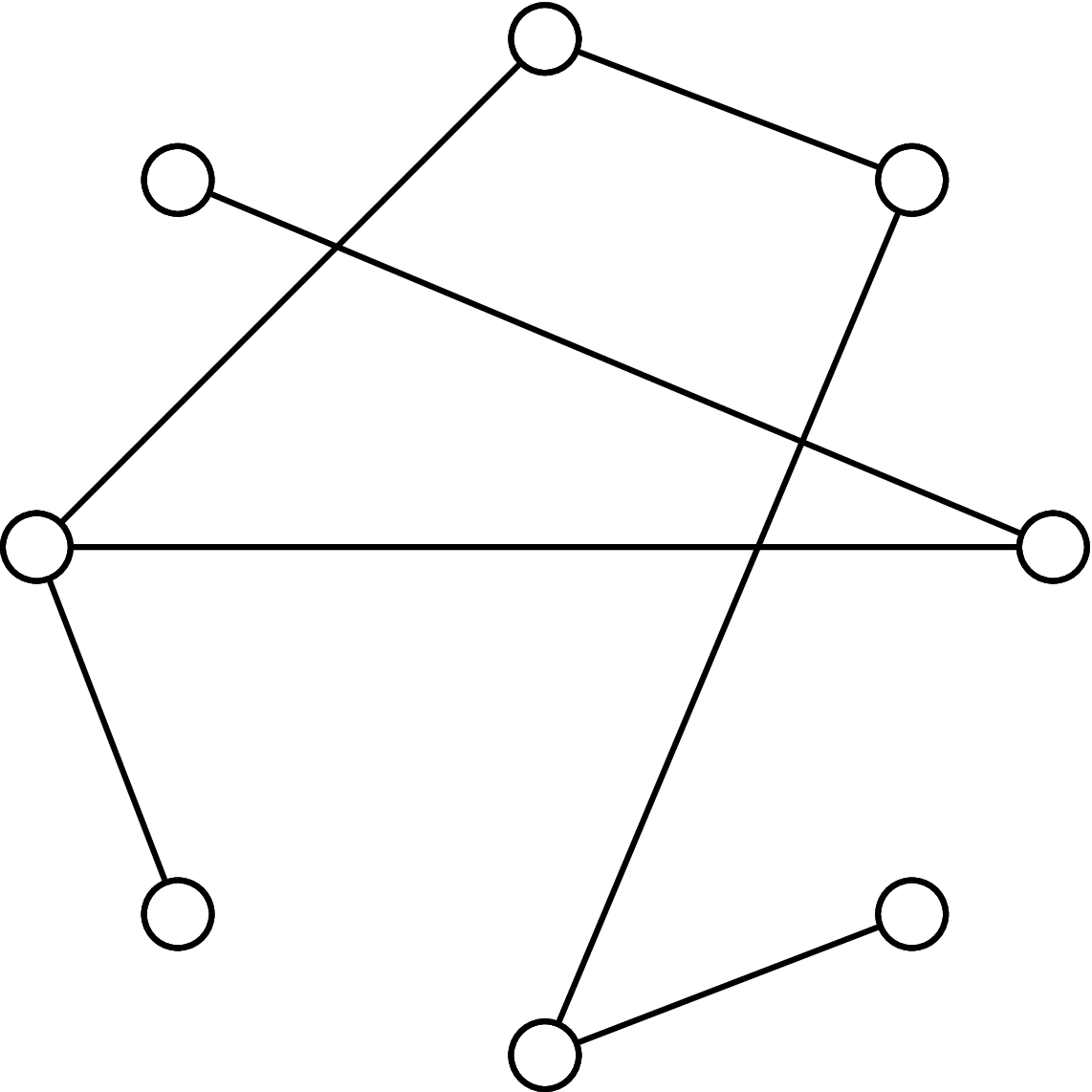}
    }\\
  \caption{Example interaction hypergraphs of randomly generated CGBGs for different parameter settings.
    }
  \label{fig:exampleGraphs}
\end{figure}

As a result, the number of edges in different CGBGs generated for the same
parameter setting may differ significantly.
The $k$ agents that participate in a new edge are selected uniformly at random from the
subset of agents involved in the fewest number of edges.  Payoffs
$\utI\ed(\jtypeG{\age},\jaG\age)$  
are
drawn from a normal distribution $\mathcal{N}(0,1)$, and the local
joint type probabilities $\Pr(\jtypeG{\age})$ are drawn from a uniform
distribution and then normalized. This algorithm results in fully
connected interaction hypergraphs that are balanced in terms of the number of payoff
functions in which each agent participates.
\fig{exampleGraphs} shows some examples of the interaction hypergraphs generated.

We test the following methods:
\begin{description}
\item[\bf NDP] Non-serial dynamic programming (see \sec{ndp}),
  run on the agent-independence factor graph (NDP-AI), the
  type-independence factor graph (NDP-TI), and the proposed agent-type
  independence factor graph (NDP-ATI).
\item[\bf MP] $\<MP>$ with the following parameters: 10 restarts, 25
  maximum iterations, sequential random message
  passing scheme, damping factor
  $0.2$. Analogous to NDP, there are three variations $\<MP>$-AI, $\<MP>$-TI,
  and $\<MP>$-ATI.
\item[\bf \BAGABAB] Bayesian game branch and bound ($\BAGABAB$) is a
  fast method for optimally solving CBGs \citep{Oliehoek:10:AAMAS}.  It performs
  heuristic search over partially specified policies.\footnote{In the
  experiments we used the 
  `MaxContributionDifference' joint type ordering and use the `consistent
  complete information' heuristic.}
\item[\bf $\AltMax$] Alternating Maximization with 10 restarts (see
  \sec{SolvingCBGs}).
\item[\bf CE] Cross Entropy optimization \citep{Boer:05:AnnalsOR:CEtutor} is a
    randomized optimization method that maintains a distribution over joint
    policies. It works by iterating the following steps: 
    1) sampling a set of joint policies from the distribution 
    2) using the best fraction of samples to update the distribution.
    We used two parameter settings: one that gave good results according to
    \citep{Oliehoek:08:Informatica} ($\CEnormal$), and one that is
    faster ($\CEfast$).\footnote{Both variants perform 10 restarts, use a
    learning rate of $0.2$ and perform what
    \citep{Oliehoek:08:Informatica} refer to as `approximate
    evaluation' of joint policies. $\CEnormal$ performs 300 and
    $\CEfast$ 100 simulations per joint policy. $\CEnormal$ performs
    $I=50$ iterations, in each of which $N=100$ joint policies are
    sampled of which $N_{b}=5$ policies are used to update the
    maintained distribution. $\CEfast$ uses $I=15,$ $N=40,$
    $N_{b}=2$.}
\end{description}

All methods were implemented using the MADP Toolbox
\citep{Spaan:08:MSDM}; the NDP and $\<MP>$ implementations also use
\problemName{libDAI} \citep{Mooij:08:libDAI}.  Experiments in this
section are run on an Intel Core i5 CPU ($2.67$GHz) using Linux, and the
timing results are CPU times. Each process is limited
to 1GB of memory use and the computation time for solving a
single CGBG is limited to $5s$. 

For each method, we report
both the average payoff and the
average CPU time needed to compute the solution.
In the plots in this section, each data point represents an average
over $N_g = 1,000$ games.  
The reported payoffs are
normalized with respect to those of $\<MP>$-ATI.
As such, the payoff of $\<MP>$-ATI is
always $1$. Error bars indicate the standard deviation of the sampled
mean $\sigma_{\rm mean} =
{\sigma}/{\sqrt{N_g}}$. 

\subsubsection{Comparing Methods}
\label{sec:comp-other}

\begin{figure}
    \centering
    \subfloat[Scaling $\nrA$: Payoff s ($k =2, |\typeAS i| = 3, |\aAS i| = 3$).]{

        \includegraphics[width=0.42\textwidth]{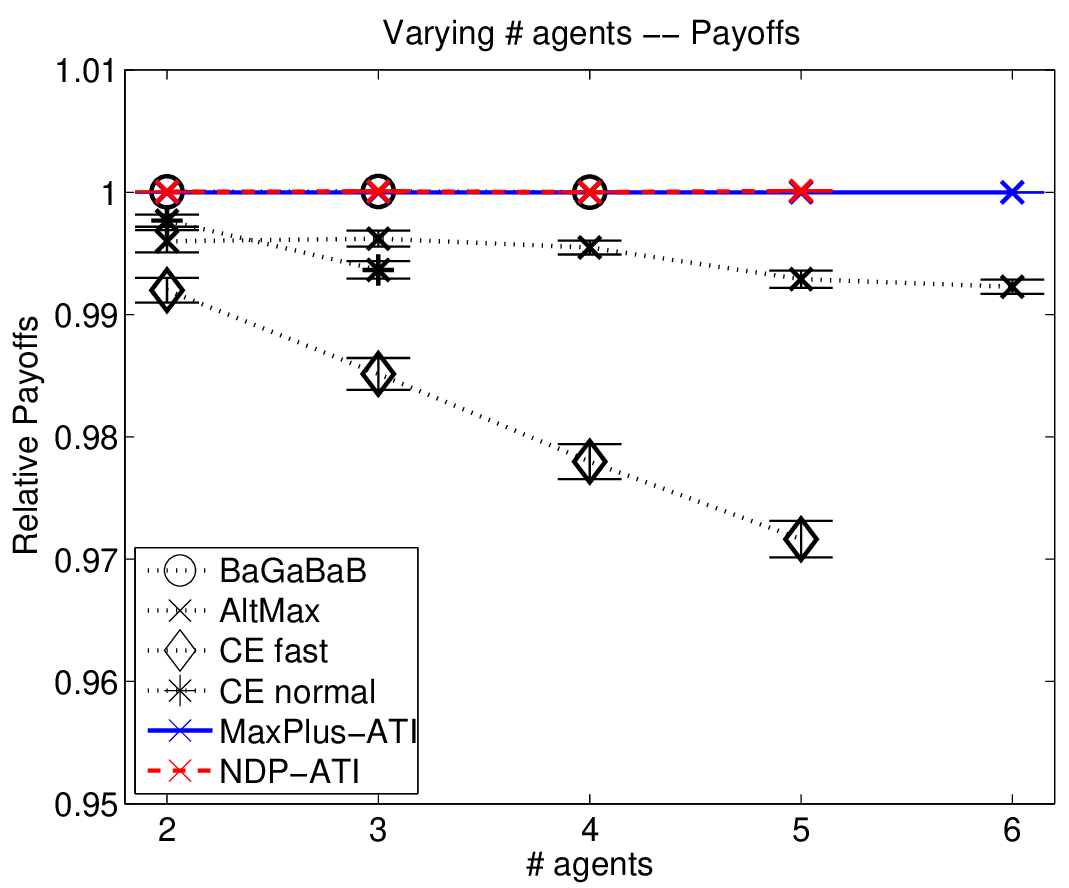}
        \label{fig:bgcgRandomVaryingAgents:nonGraphPayoff}
    }\hspace{0.05\textwidth}
    \subfloat[Scaling $\nrA$: Computation times ($k =2, |\typeAS i| = 3, |\aAS i| = 3$).]{

        \includegraphics[width=0.42\textwidth]{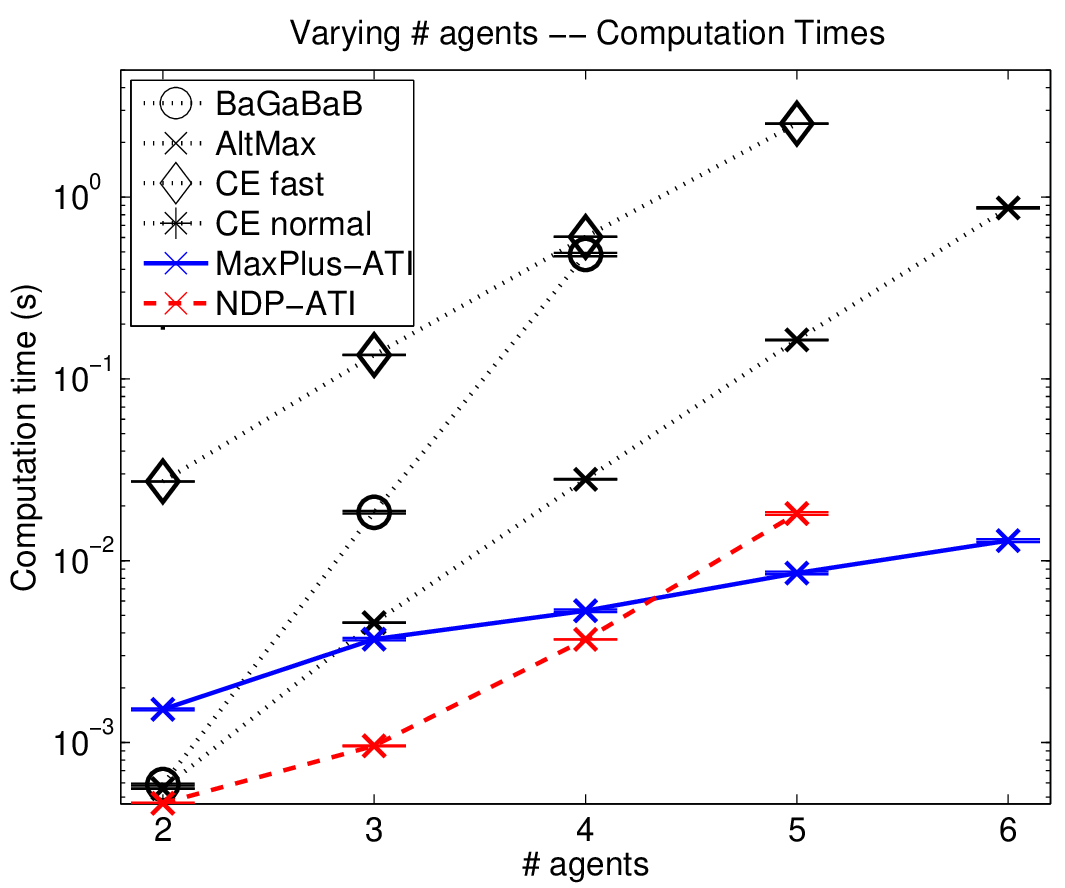}
        \label{fig:bgcgRandomVaryingAgents:nonGraphTiming}
    }\\
    \subfloat[Scaling $|\aAS i|$: Payoff ($k =2, |\typeAS i| = 3, \nrA = 5$).]{

        \includegraphics[width=0.42\textwidth]{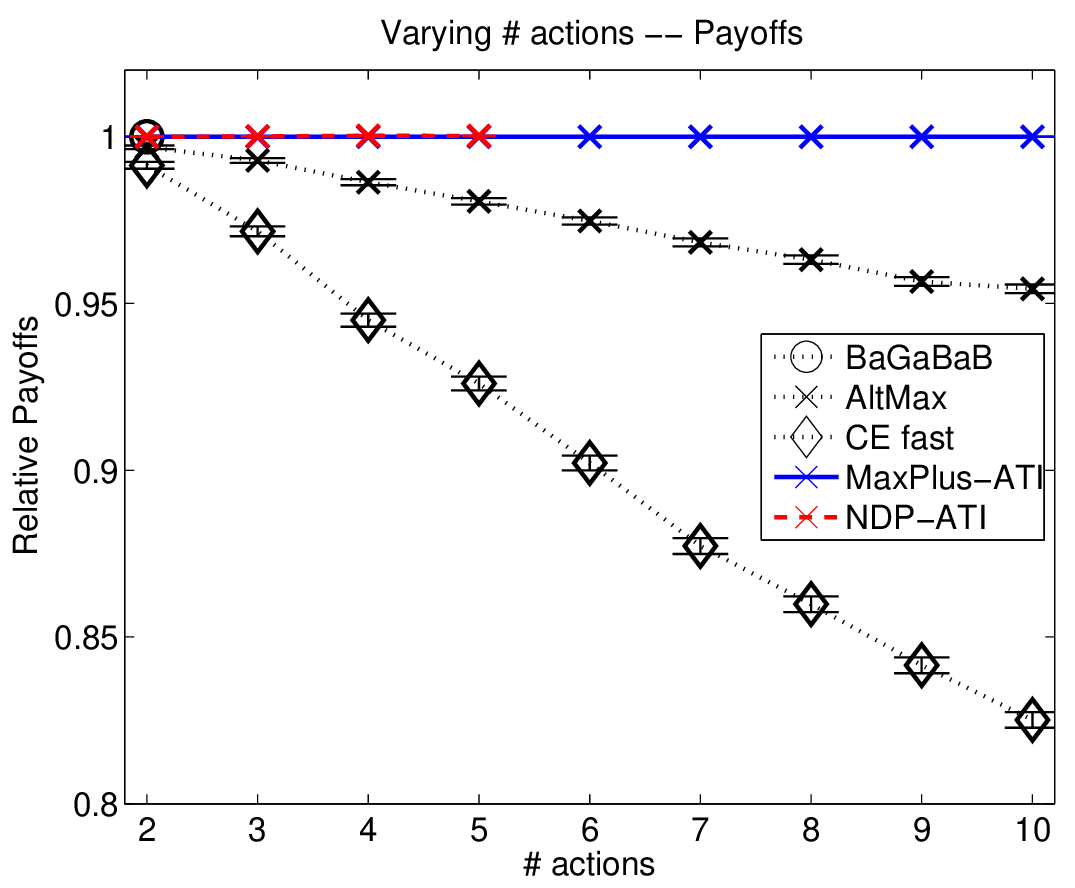}
        \label{fig:bgcgRandomVaryingActions:nonGraphPayoff}
    }\hspace{0.05\textwidth}
    \subfloat[Scaling $|\aAS i|$: Computation times ($k =2, |\typeAS i| = 3, \nrA = 5$).]{

        \includegraphics[width=0.42\textwidth]{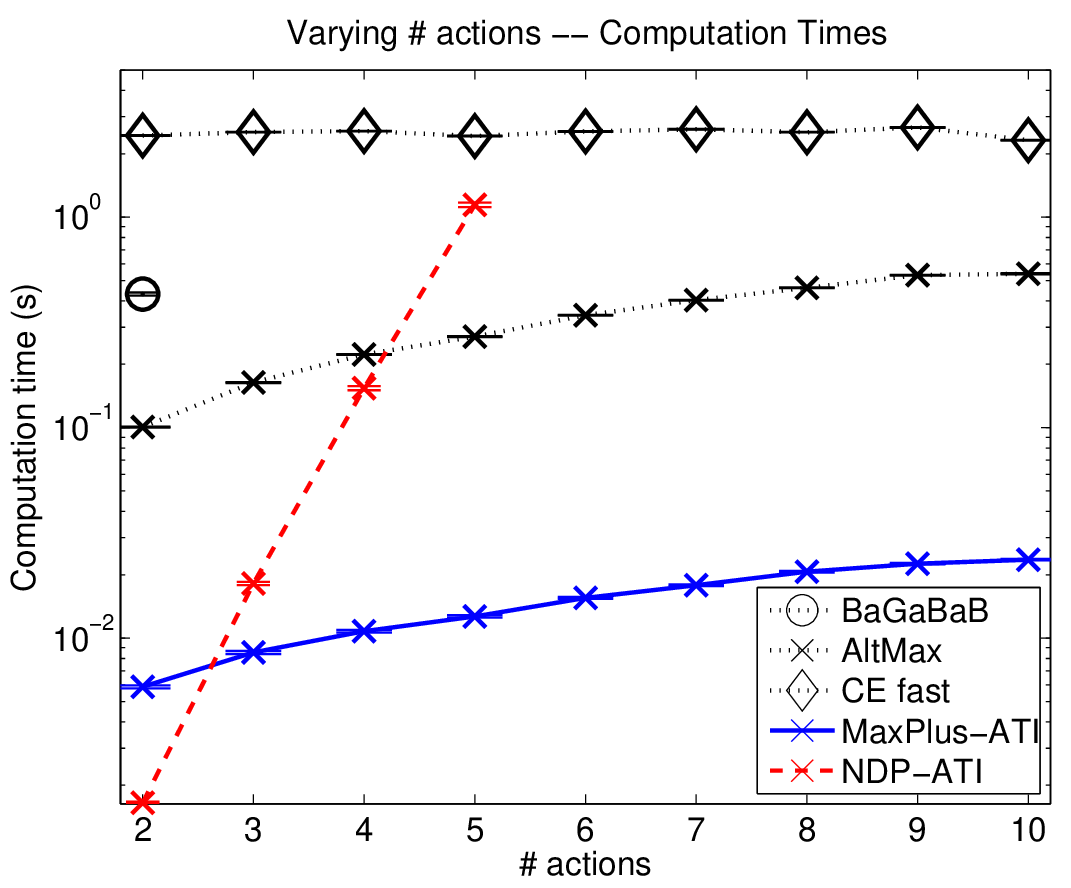}
        \label{fig:bgcgRandomVaryingActions:nonGraphTiming}
    }\\
    \subfloat[Scaling $|\typeAS i|$: Payoff ($k =2, |\aAS i| = 3, \nrA = 5$).]{

        \includegraphics[width=0.42\textwidth]{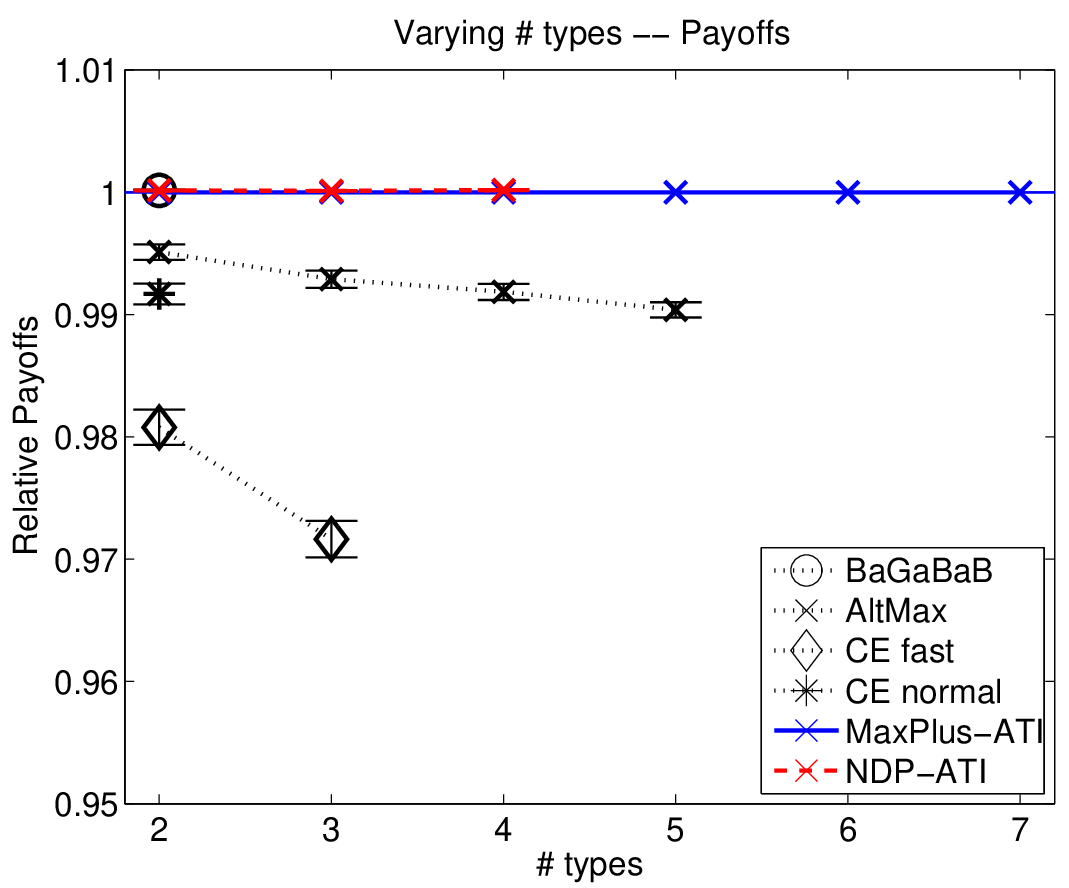}
        \label{fig:bgcgRandomVaryingTypes:nonGraphPayoff}
    }\hspace{0.05\textwidth}
    \subfloat[Scaling $|\typeAS i|$: Computation times ($k =2, |\aAS i| = 3, \nrA = 5$).]{

        \includegraphics[width=0.42\textwidth]{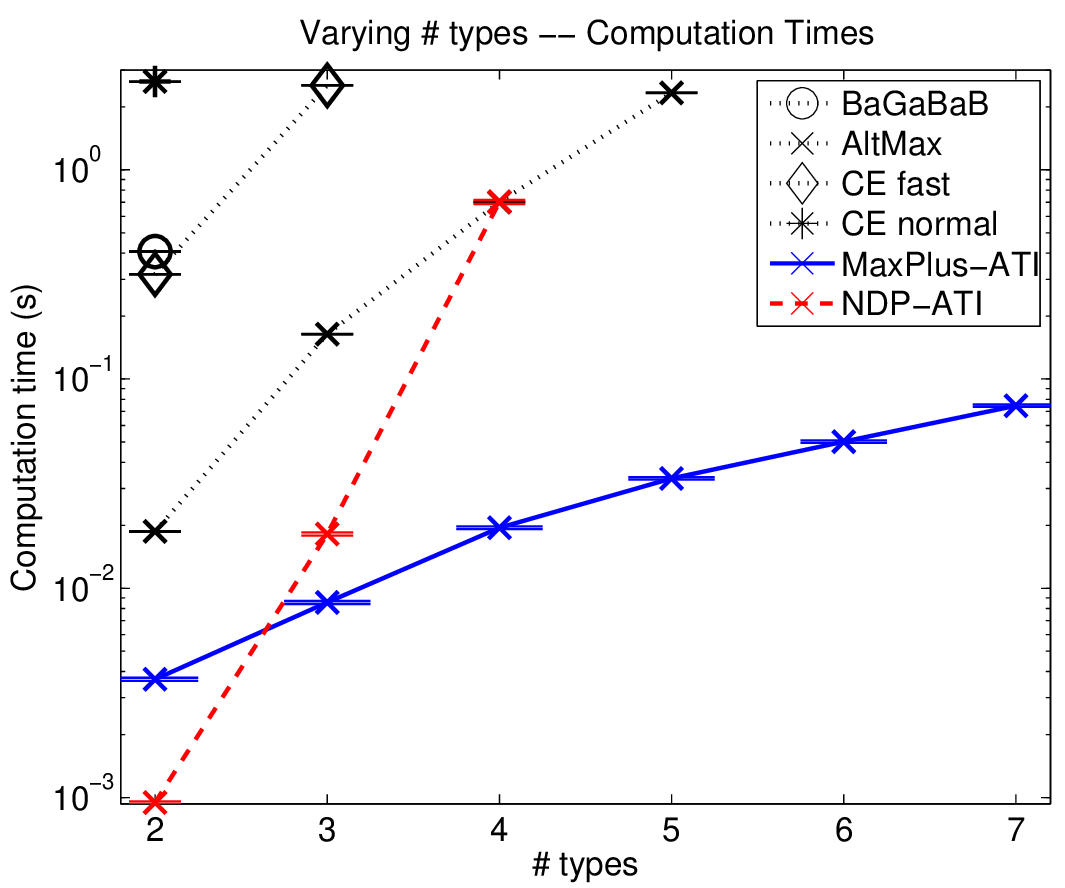}
        \label{fig:bgcgRandomVaryingTypes:nonGraphTiming}
    }\\
    \caption{
      Comparison of $\<MP>$-ATI and NDP-ATI with other methods,
      scaling the number of agents
      (\subref{fig:bgcgRandomVaryingAgents:nonGraphPayoff} and
      \subref{fig:bgcgRandomVaryingAgents:nonGraphTiming}), the number
      of actions (\subref{fig:bgcgRandomVaryingActions:nonGraphPayoff}
      and \subref{fig:bgcgRandomVaryingActions:nonGraphTiming})), and
      the number of types
      (\subref{fig:bgcgRandomVaryingTypes:nonGraphPayoff} and
      \subref{fig:bgcgRandomVaryingTypes:nonGraphTiming}).
  }
  \label{fig:bgcgRandomComparisonOther}
\end{figure}

First, we compare NDP-ATI and $\<MP>$-ATI with other methods that do not
exploit a factor graph representation explicitly, the results of which
are shown in \fig{bgcgRandomComparisonOther}.
These results demonstrate that, as the number of agents increases, the average
payoff of the approximate non-graphical methods goes down (\fig{bgcgRandomVaryingAgents:nonGraphPayoff}) while computation time goes up (\fig{bgcgRandomVaryingAgents:nonGraphTiming}), given that the other parameters are fixed at
at $k =2, |\typeAS i| = 3, |\aAS i| = 3$. Note that a data point
is not presented if the method exceeded the
pre-defined resource limits on one or more test runs.
For instance, $\BAGABAB$ can  compute solutions only up to 4 agents. Also,
\fig{bgcgRandomVaryingAgents:nonGraphTiming} suggests that NDP-ATI would on average complete within the 5s deadline for 6 agents.  However, because there is at least one run that
does not, the data point is not included.

Next, we fix the number of agents to 5, and vary the number of actions
$|\aAS i|$. While $\CEnormal$ never meets the time limit, the
computation time that $\CEfast$ requires is relatively independent of
the number of actions (\fig{bgcgRandomVaryingActions:nonGraphTiming}).
Payoff, however, drops sharply when the number of actions increases
(\fig{bgcgRandomVaryingActions:nonGraphPayoff}). The $\CE$ solvers
maintain a fixed-size pool of possible solutions, which explains both
phenomena: the same number of samples need to cover a larger search
space, but the cost of evaluating each sample is relatively
insensitive to $|\aAS i|$.

Finally, we consider the behavior of the different methods when
increasing the number of individual types $|\typeAS i|$ (\fig{bgcgRandomVaryingTypes:nonGraphPayoff}
and~\ref{fig:bgcgRandomVaryingTypes:nonGraphTiming}). Since the number of
policies for an agent is exponential in the number of types, the existing methods scale poorly. As established by Corollary~\ref{cor:NDP_ATI_complexity}, NDP's computational costs also grow exponentially with the number of types.
$\<MP>$, in contrast, scales much better in the number of types. Looking at the quality of the found policies,
we see that \<MP> achieves the optimal value in these experiments, while the other approximate methods achieve lower values.

\subsubsection{Comparing Factor-Graph Formulations}
\label{sec:comp-prop-meth}

We now turn to a more in-depth analysis of the NDP and $\<MP>$ methods, in order to establish the effect of
exploiting different types of independence. In particular, we test both methods on three different types of factor
      graphs: those with agent-independence (AI), type-independence (TI), and agent-type independence (ATI). \tab{FGprops} summarizes the types of factor graphs and the symbols used to describe their various characteristics.

\begin{table}
    \begin{centering}

    {\noindent
\renewcommand{\arraystretch}{1.5}
\setlength{\tabcolsep}{4pt}
\begin{tabular}{r||c|c|c|c|c|c}
FG type                         
& num. factors ($F$)                
& fact. size                                    
& fact. deg. ($k$)      
& num. vars                     
& var. size ($m$)                                    
& var. deg. ($l$)         
\tabularnewline
\hline
\hline
                    AI          
& $\nrR$                            
& $\left|\aAS*\right|^{\left|\typeAS*\right|k}$ 
& $|\ed_*|$             
& $\nrA$                        
& $\left|\aAS*\right|^{\left|\typeAS*\right|}$  
& $\nrRA{*}$                
\tabularnewline
                   TI           
& $\left|\typeAS*\right|^{\nrA}$    
& $\left|\aAS*\right|^{\nrA}$                   
& $\nrA$                
& $\nrA\left|\typeAS*\right|$   
& $\left|\aAS*\right|$                          
& $|\typeAS{*}|^{\nrA-1}$     
\tabularnewline
                         ATI    
& $\nrR\left|\typeAS*\right|^{k}$   
& $\left|\aAS*\right|^{k}$                      
& $|\ed_*|$             
& $\nrA\left|\typeAS*\right|$   
& $\left|\aAS*\right|$                          
& $|\typeAS{*}|^{k-1}$        
\tabularnewline
\end{tabular}
}
    \end{centering}
    \caption{A characterization of the different types of factor
      graphs: agent-independence (AI), type-independence (TI),
      agent-type independence (ATI). The symbols relate to \eq{MPcomplexity}. }
    \label{tab:FGprops}
\end{table}

\begin{figure}
  \centering
  \subfloat[Payoff, $k =2, |\typeAS i| = 3, |\aAS
    i| = 3$.]{

        \includegraphics[width=0.42\textwidth]{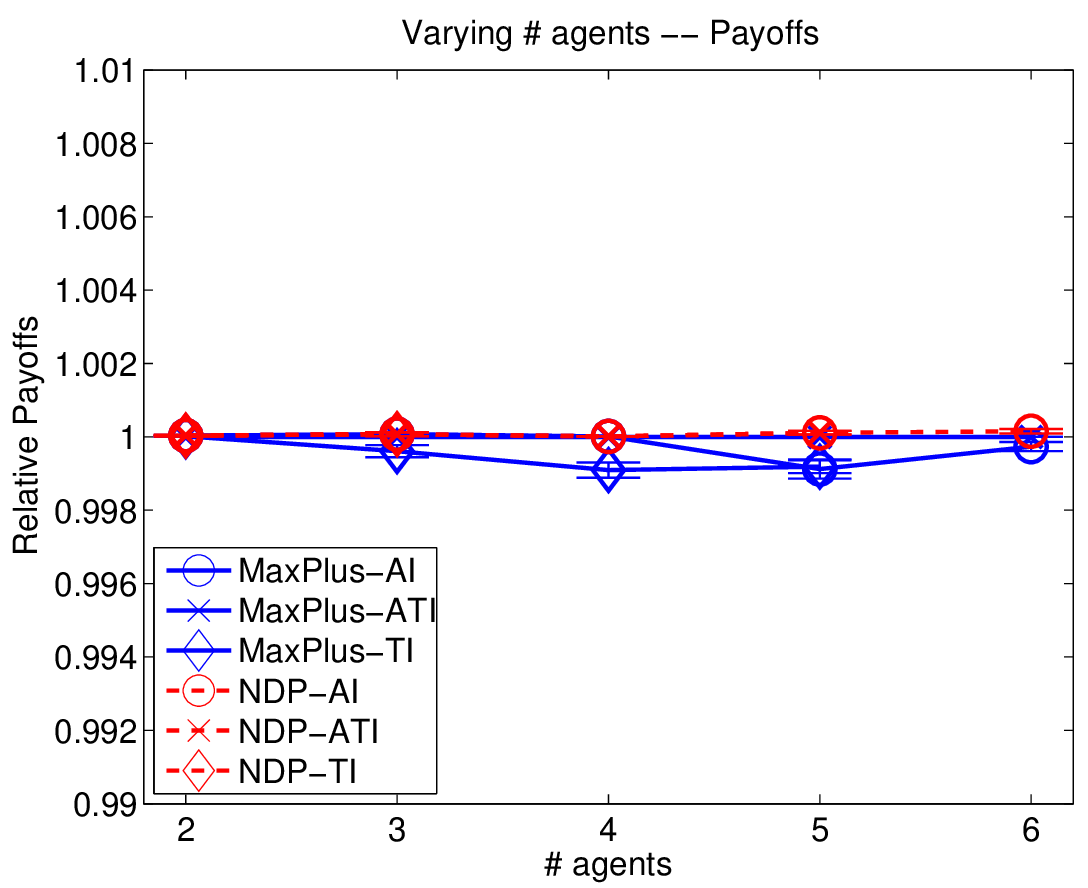}
        \label{fig:bgcgRandomVaryingAgents:graphPayoff}}\hspace{0.05\textwidth}
  \subfloat[Computation times, $k =2, |\typeAS i| = 3, |\aAS
    i| = 3$.]{

        \includegraphics[width=0.42\textwidth]{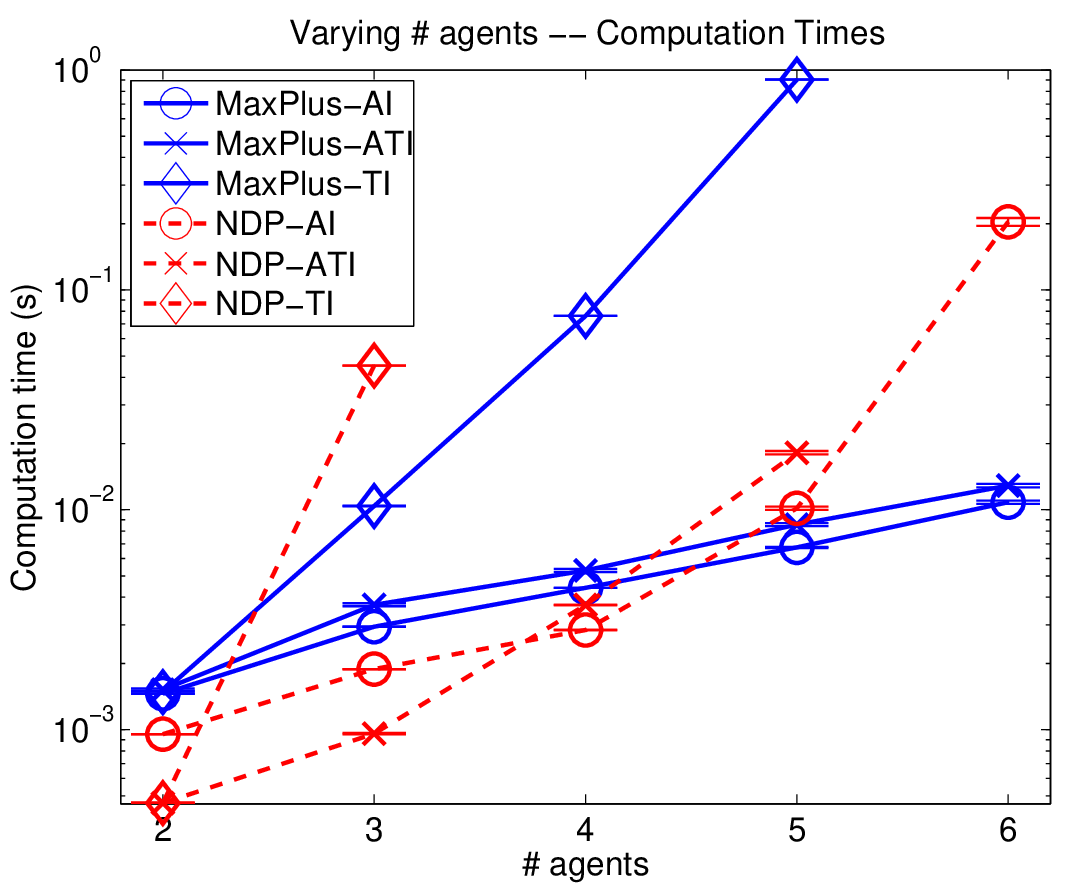}
        \label{fig:bgcgRandomVaryingAgents:graphTiming}}\\
  \subfloat[Payoff, $k =3, |\typeAS i| = 3, |\aAS
    i| = 3$.]{

        \includegraphics[width=0.42\textwidth]{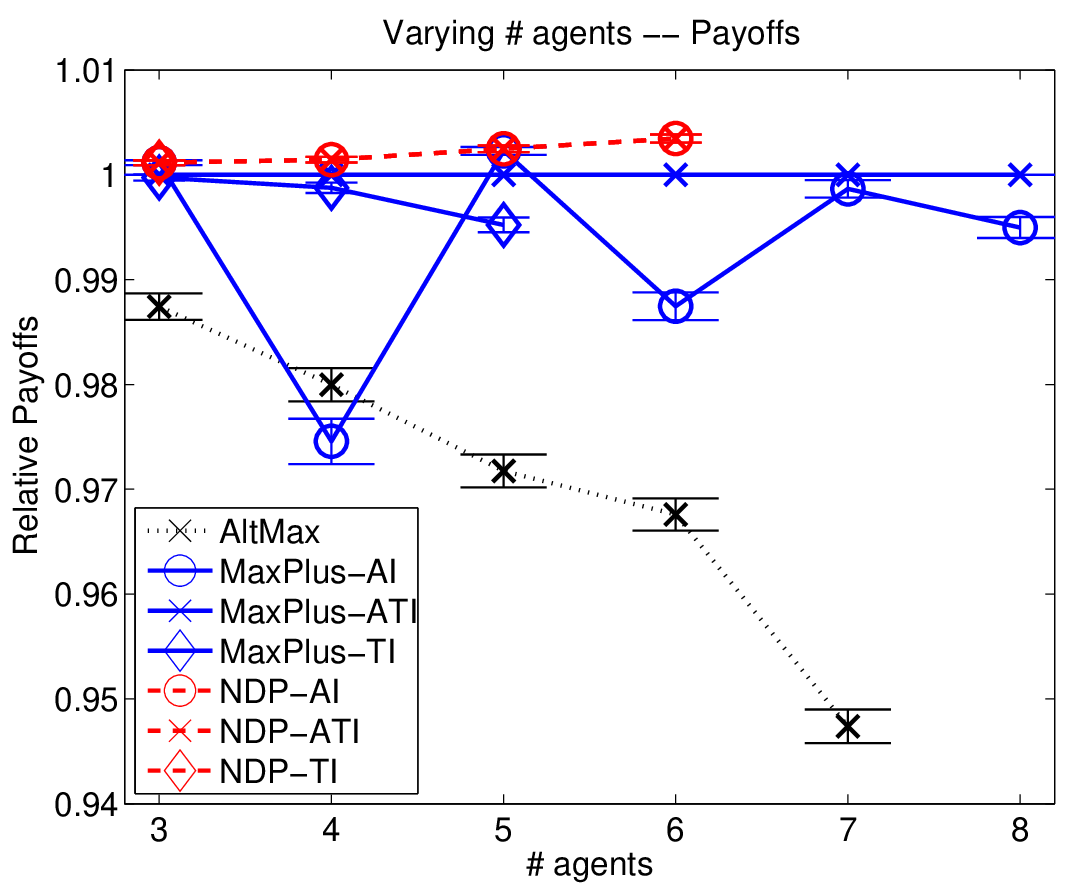}
        \label{fig:bgcgRandomVaryingAgents_nrAgPerEd3:payoff}}\hspace{0.05\textwidth}
  \subfloat[Computation times, $k =3, |\typeAS i| = 3, |\aAS
    i| = 3$.]{

        \includegraphics[width=0.42\textwidth]{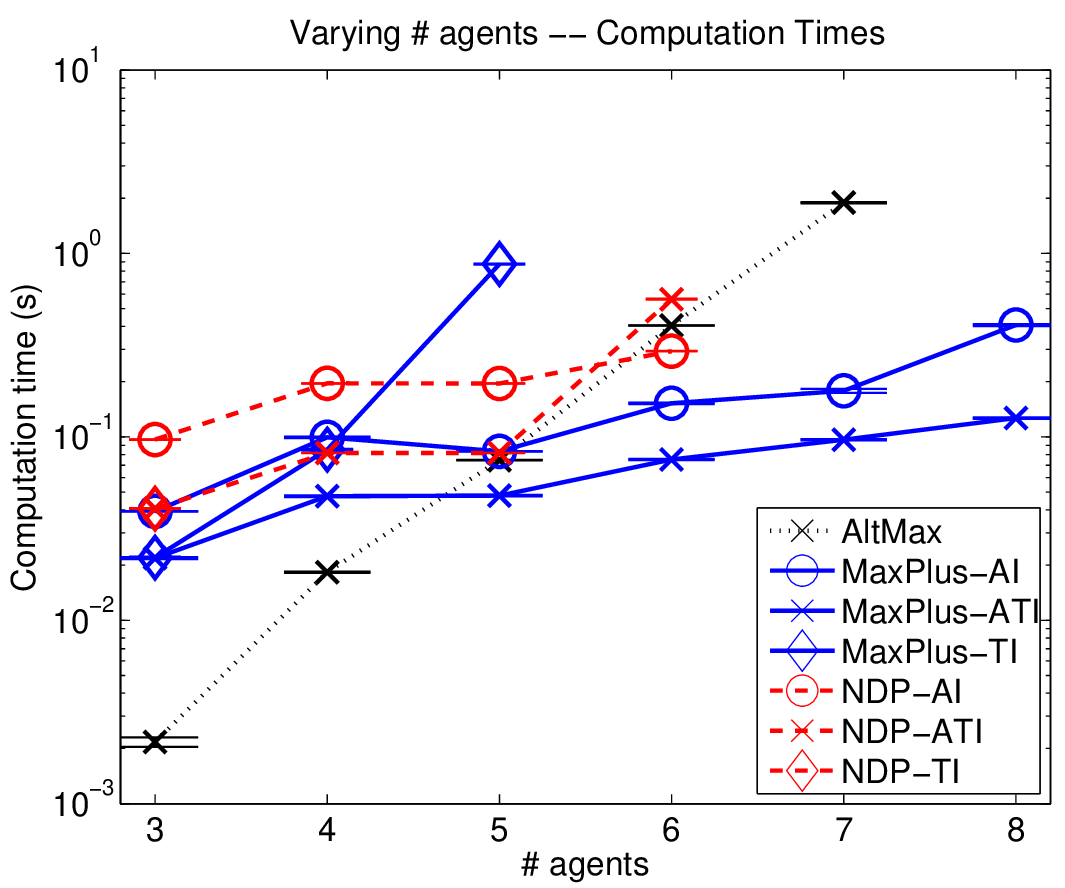}
        \label{fig:bgcgRandomVaryingAgents_nrAgPerEd3:timing}}\\
    \subfloat[Payoff, $k =2, |\typeAS i| = 4, |\aAS i| = 4$, and the
    average number of edges per game (note the two different
    $y$-axes indicated with two colors).]{

        \includegraphics[width=0.42\textwidth]{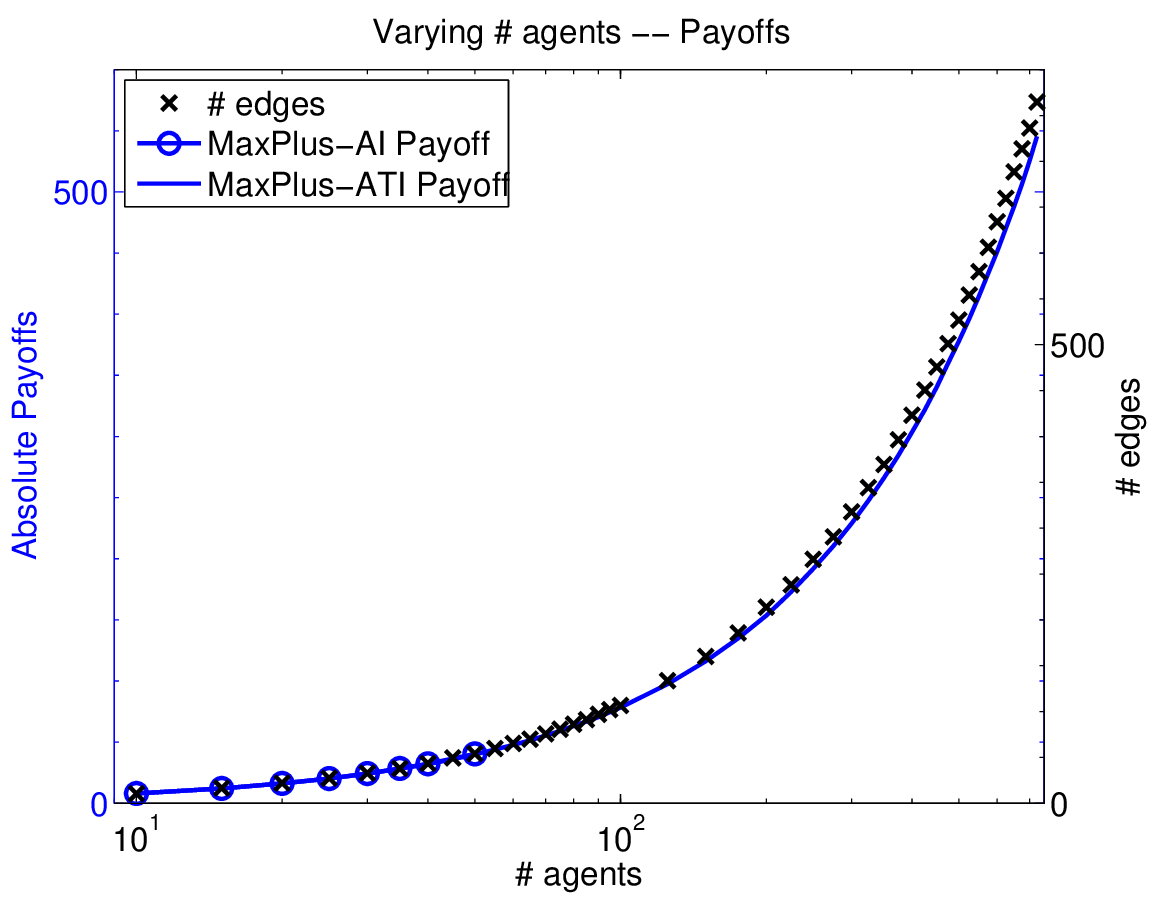}
        \label{fig:bgcgRandomVaryingManyAgents:payoff}}\hspace{0.05\textwidth}
    \subfloat[Computation times, $k =2, |\typeAS i| = 4, |\aAS i| =
    4$.]{

        \includegraphics[width=0.42\textwidth]{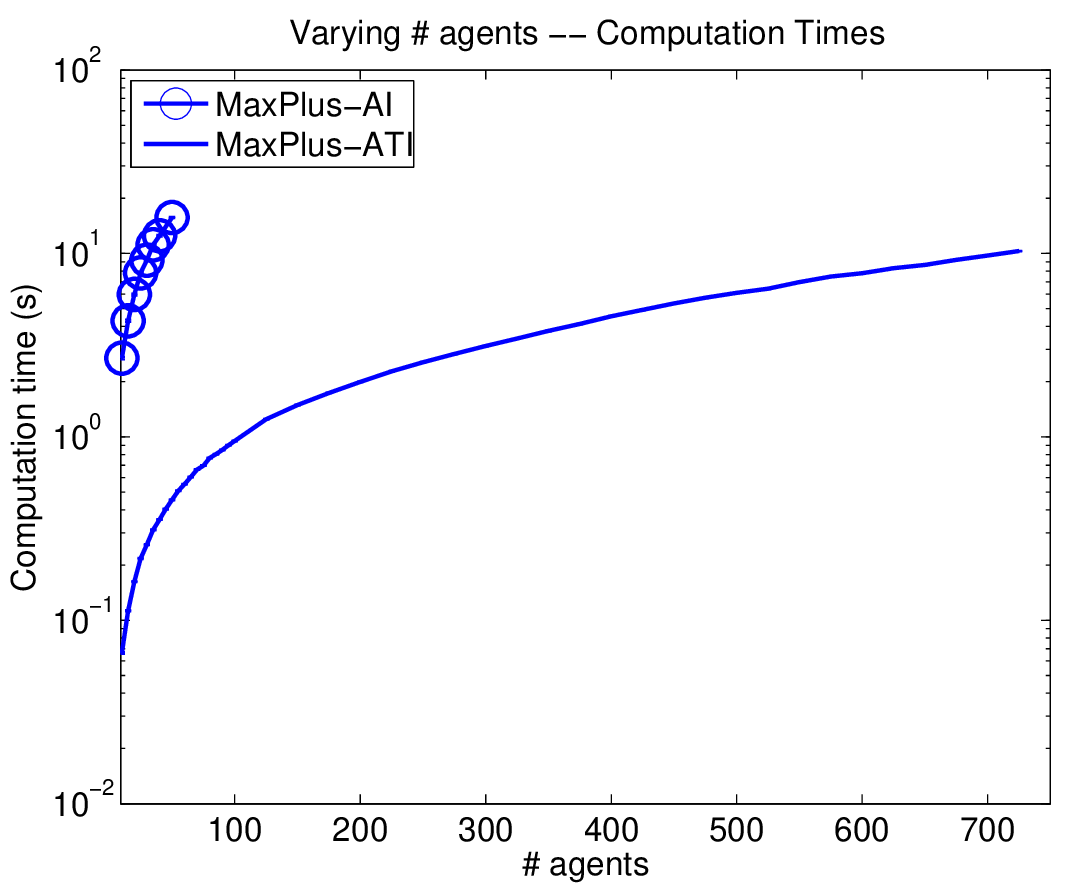}
        \label{fig:bgcgRandomVaryingManyAgents:timing}}
    \caption{Comparison of the proposed factor-graph methods when
      scaling the number of agents~$\nrA$. Plots
      \subref{fig:bgcgRandomVaryingAgents:graphPayoff} and
      \subref{fig:bgcgRandomVaryingAgents:graphTiming} consider $k=2$
      (analogous to \fig{bgcgRandomVaryingAgents:nonGraphPayoff} and
      \ref{fig:bgcgRandomVaryingAgents:nonGraphTiming}), while
      \subref{fig:bgcgRandomVaryingAgents_nrAgPerEd3:payoff} and
      \subref{fig:bgcgRandomVaryingAgents_nrAgPerEd3:timing} show
      results for hyper-edges with $k=3$.  Plots
      \subref{fig:bgcgRandomVaryingManyAgents:payoff} and
      \subref{fig:bgcgRandomVaryingManyAgents:timing} display the
      scaling behavior for many agents.}

    \label{fig:bgcgRandomVaryingAgents}
\end{figure}

First, we consider scaling the number of agents, using the same
parameters as in \fig{bgcgRandomVaryingAgents:nonGraphPayoff}
and~\ref{fig:bgcgRandomVaryingAgents:nonGraphTiming}.  
\fig{bgcgRandomVaryingAgents:graphPayoff} shows the payoff of the
different methods. The difference between them is not significant.
However, \fig{bgcgRandomVaryingAgents:graphTiming} shows the computation times of the same methods. As expected,
the methods that use only type independence scale poorly because the number
of factors in their factor graph is exponential in the number of agents (\tab{FGprops}).

\fig{bgcgRandomVaryingAgents_nrAgPerEd3:payoff}
and~\ref{fig:bgcgRandomVaryingAgents_nrAgPerEd3:timing} show similar comparisons for payoff functions involving three agents, i.e., $k =3$ (example interaction hypergraphs are shown in
\fig{exampleGraphs:3agPerEdge}). The difference in payoff between NDP-ATI and $\<MP>$-ATI 
is not significant (minimum p-value is 0.61907 for 6 agents). 
Differences with AM and the outlying points of $\<MP>$-AI are significant (p-value $< 0.05$). The NDP-AI and NDP-ATI methods scale
to 6 agents, while $\<MP>$-AI and $\<MP>$-ATI scale beyond. The payoff
of $\<MP>$-AI is worse and more erratic than the payoff of $\<MP>$-ATI. In
this case, due to increased problem complexity, the $\<MP>$ methods typically do not attain the true optimum.

These experiments clearly demonstrate that only $\<MP>$-AI and $\<MP>$-ATI
scale to larger numbers of agents. The poor scalability of the non-factor-graph methods is due to their failure to exploit the independence in CGBGs.  The methods using TI factor graphs scale poorly because they ignore independence between agents.
As hypothesized, NDP is not able to effectively exploit 
type independence and consequently NDP-ATI does not outperform NDP-AI. In fact, the experiments show that, in some cases, NDP-AI slightly outperforms NDP-ATI.

\fig{bgcgRandomVaryingManyAgents:payoff}
and~\ref{fig:bgcgRandomVaryingManyAgents:timing} show the performance
of $\<MP>$-AI and $\<MP>$-ATI for games with $k =2, |\typeAS i| = 4, |\aA
i| = 4$ and larger numbers of agents, from $n = 10$ up to $n=
725$ (limited by the allocated memory space). For this experiment, the
methods were allowed 30$s$ per CGBG instance.
\fig{bgcgRandomVaryingManyAgents:payoff} shows the absolute payoff
obtained by both methods and the growth in the number of payoff
functions (edges).  The results demonstrate that the $\<MP>$-ATI payoffs do not
deteriorate when increasing the number of payoff functions.  Instead,
they increase steadily at a rate similar to the
number of payoff functions.  This is as expected, since more payoff functions means there is
more reward to be collected.  $\<MP>$-AI scales only to 50 agents,
and its payoffs are close to those obtained by $\<MP>$-ATI.
\fig{bgcgRandomVaryingManyAgents:timing} provides clear experimental
corroboration of \thm{CGBG_MPcomplexity} (which states that there is no
exponential dependence on the number of agents), by showing scalability to 725 agents.

\begin{figure}[tb]
  \centering
  \subfloat[Scaling $|\aAS i|$: Payoff ($k =2,
    |\typeAS i| = 3, \nrA = 5$).]{

        \includegraphics[width=0.42\textwidth]{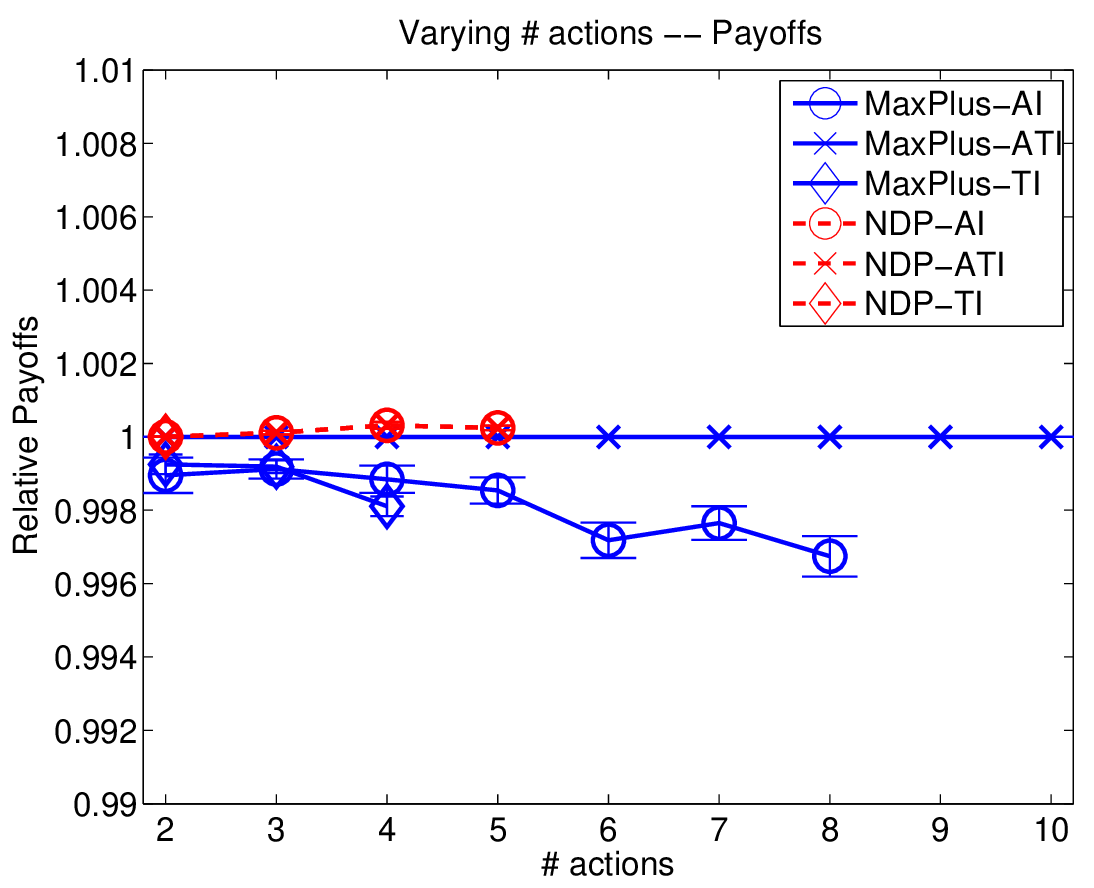}
        \label{fig:bgcgRandomVaryingActions:graphPayoff}}\hspace{0.05\textwidth}
  \subfloat[Scaling $|\aAS i|$: Computation times ($k =2,
    |\typeAS i| = 3, \nrA = 5$).]{

        \includegraphics[width=0.42\textwidth]{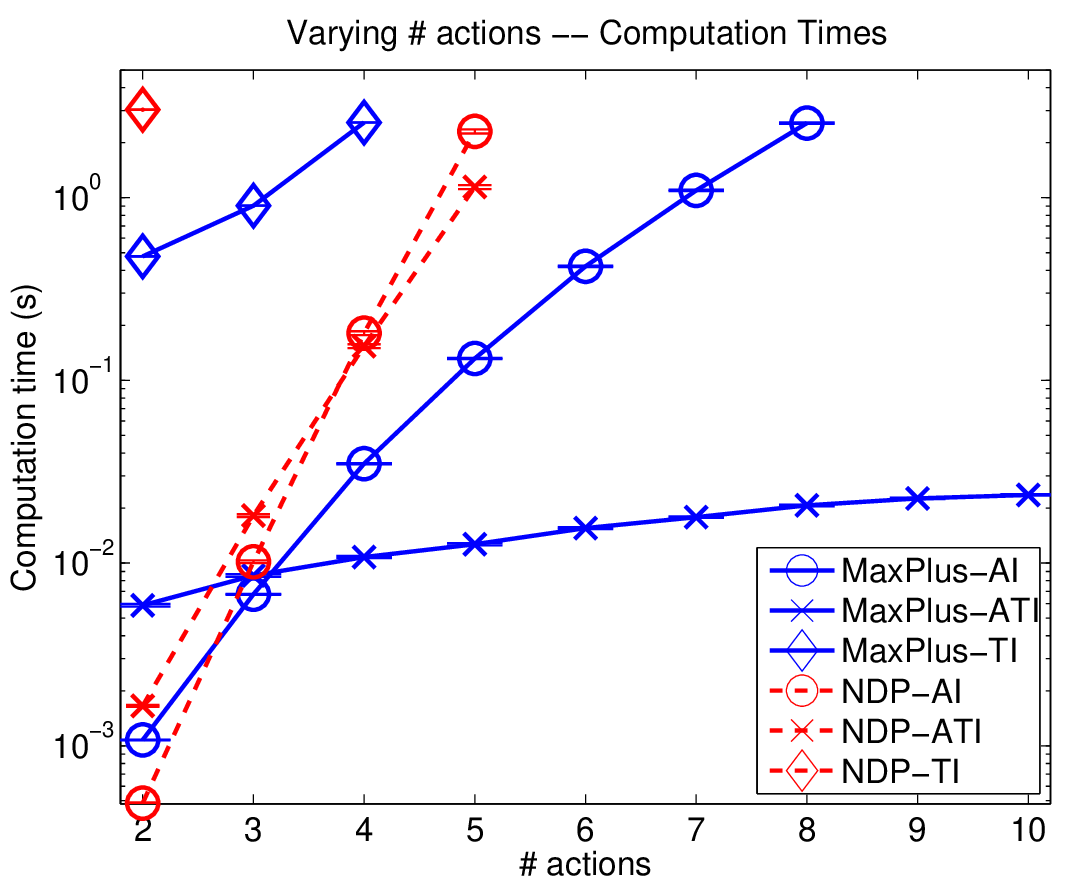}
        \label{fig:bgcgRandomVaryingActions:graphTiming}}\\
  \subfloat[Scaling $|\typeAS i|$: Payoff ($k =2, |\aAS i|
    = 3, \nrA = 5$).]{

        \includegraphics[width=0.42\textwidth]{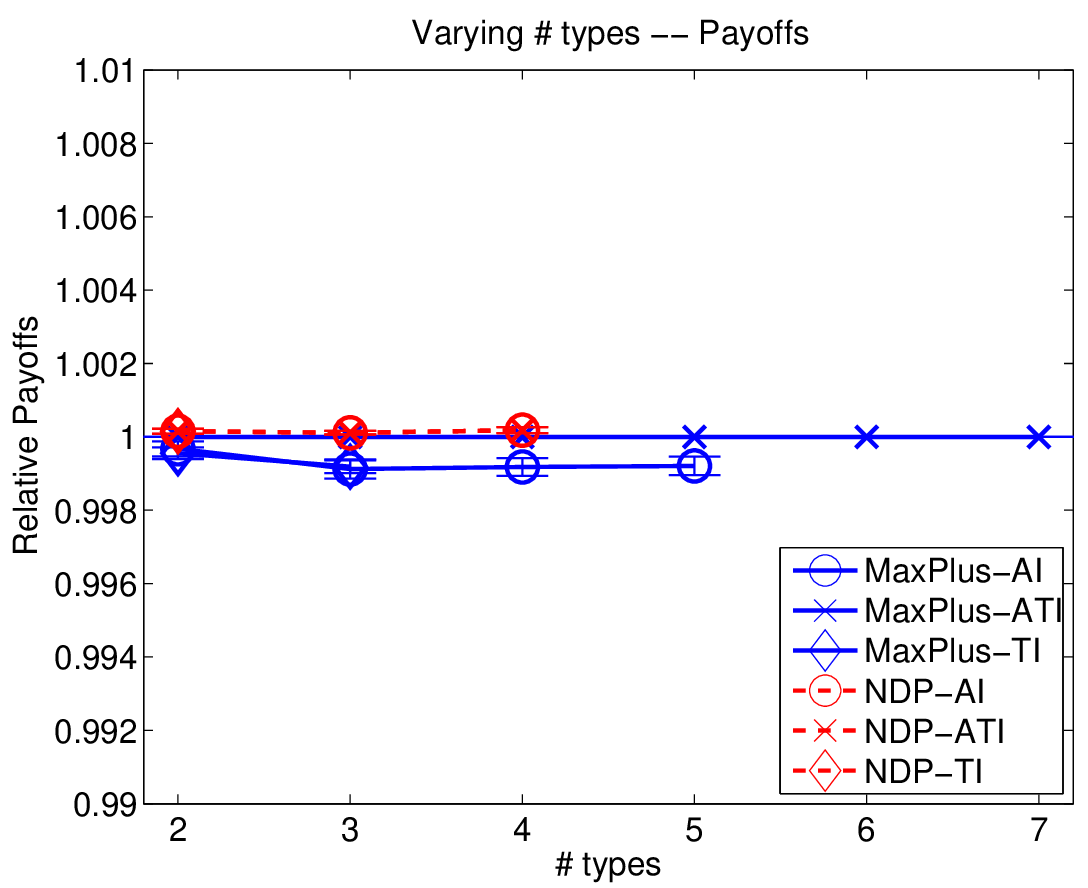}
        \label{fig:bgcgRandomVaryingTypes:payoff}}\hspace{0.05\textwidth}
  \subfloat[Scaling $|\typeAS i|$: Computation times ($k =2, |\aAS i|
    = 3, \nrA = 5$).]{

        \includegraphics[width=0.42\textwidth]{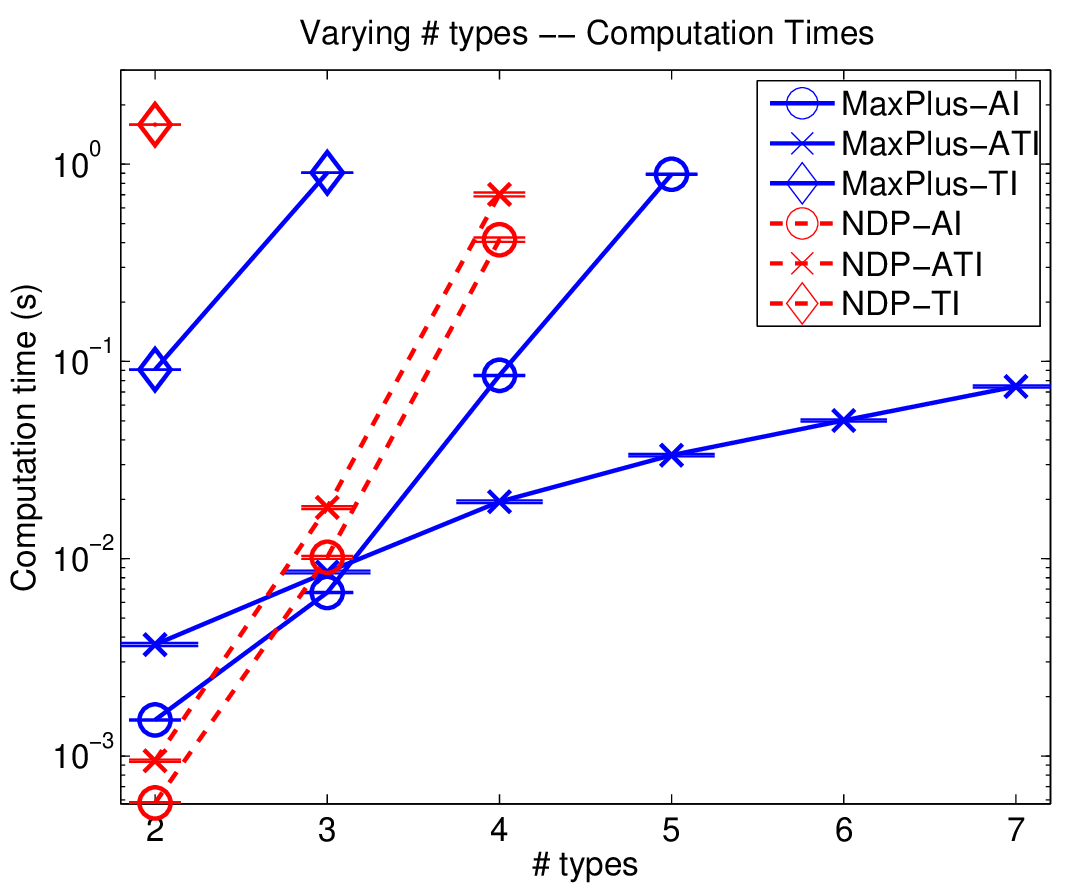}
        \label{fig:bgcgRandomVaryingTypes:timing}}
    \caption{Comparison of the proposed factor-graph methods when
      scaling the number of actions and the number of types. Plots
      \subref{fig:bgcgRandomVaryingActions:graphPayoff} and
      \subref{fig:bgcgRandomVaryingActions:graphTiming} consider
      scaling $|\aAS i|$ (analogous to
      \fig{bgcgRandomVaryingActions:nonGraphPayoff} and
      \ref{fig:bgcgRandomVaryingActions:nonGraphTiming}), while
      \subref{fig:bgcgRandomVaryingTypes:payoff} and
      \subref{fig:bgcgRandomVaryingTypes:timing} show results
      increasing $|\typeAS i|$ (cf.\
      \fig{bgcgRandomVaryingTypes:nonGraphPayoff} and
      \ref{fig:bgcgRandomVaryingTypes:nonGraphTiming}).}
      \label{fig:bgcgRandomVaryingActionTypes}  
\end{figure}

Analogously to \sec{comp-other}, we compare the different factor-graph
methods when increasing the number of actions and
types. \fig{bgcgRandomVaryingActions:graphTiming} shows that $\<MP>$-ATI scales better in the number of actions while obtaining 
payoffs close to optimal (when available) and better than
other $\<MP>$ variations (\fig{bgcgRandomVaryingActions:graphPayoff})
(differences are not significant). In fact, it is the only method whose computation time increases only slightly
when increasing the number of actions: the size of each factor is only
$\left|\aAS*\right|^{k}$ compared to
$\left|\aAS*\right|^{\left|\typeAS*\right|k}$ for AI and
$\left|\aAS*\right|^{\nrA}$ for TI (\tab{FGprops}). In this case $k=2$
and $n=5$, and in general $k \ll n$ in the domains we consider.

When scaling the number of types (\fig{bgcgRandomVaryingTypes:payoff}
and~\ref{fig:bgcgRandomVaryingTypes:timing}), again there are no significant differences in payoffs.  However, as expected given the lack of exponential dependence on $|\typeAS i|$ (\tab{FGprops}),  $\<MP>$-ATI  performs much better in terms of computation times.

Overall, the results presented in this section demonstrate that the proposed methods substantially outperform existing solution methods for CBGs.  In addition, the experiments confirm the hypothesis that NDP is not able to effectively exploit type independence,
resulting in exponential scaling with respect to the number of types. $\<MP>$ on ATI factor graphs is able to effectively exploit both agent and type independence, resulting in much better scaling behavior with respect to all model parameters.
Finally, the experiments showed that the value of the found solutions was not significantly lower than the optimal value and, in many case, significantly better than that found by other approximate solution methods. 

\subsection{Generalized Fire Fighting Experiments}
\label{sec:ff-exp}

The results presented above demonstrate that $\<MP>$-ATI can improve
performance on a wide range of CGBGs.  However, all of the CGBGs used
in those experiments were randomly generated.  In this section, we aim
to demonstrate that the advantages of $\<MP>$-ATI extend to a more
realistic problem.  To this end, we apply it to a 2-dimensional
implementation of the \<GFF> problem described in \sec{nffExample}.
Each method was limited to 2Gb of memory and allowed 30s computation
time.

In this implementation, the $\nrHouses$ houses are uniformly spread across a 2-dimensional plane, 
i.e., the $x$ and $y$ coordinates for each house are drawn from a uniform distribution
over the interval $[0,1]$.
Similarly, each of the $\nrA$ agents is assigned a random location and can choose to fight fire at any of $\ffnrAcs$ nearest houses, subject to the constraint
that at most $k$ agents can fight fire at a particular house. We enforce this constraint by
making a house unavailable to additional agents once it is in the action sets of $k$ agents.\footnote{While this can lead to a sub-optimal assignment,
we do not need the best assignment of action sets in order to compare methods.}
In addition, each agent is assigned, in a similar fashion, the $\ffnrObs$ nearest houses that it can observe. As mentioned in \sec{nffExample}, a type $\typeA i$
is defined by the $\ffnrObs$ observations the agent receives
from the surrounding houses: $\typeA{i} \in \{\ffOfl_i, \ffOnf_i\}^\ffnrObs $.
We assume that $\ffnrObs \leq \ffnrAcs$ to ensure that no local payoff function depends on an agent simply due to that agent's type, i.e., an agent never observes a house at which it cannot fight fire.

To ensure there are always enough houses, the number of houses is made 
proportional to both the number of agents
and actions: $\nrHouses=\text{ceil}(\GFFdensity \mult \ffnrAcs \mult
\nrA )$, where $\GFFdensity$ is set to $1.2$ unless noted otherwise. 
Each house has a fire level that is drawn uniformly from $\{0,\dots,\nrFLs-1\}$. The probability that an
agent receives the observation $\ffOfl$ for a house $\house$ depends on its fire level, as shown
in \tab{GFF_Oprobs}. Observations of different houses by a single agent~$i$  are assumed 
to be independent, but observations of different agents that can observe the same house are coupled 
through the hidden state. 
\begin{table}
\begin{center}
\begin{tabular}{c|c}
$\fl\house$ & $\Pr(\ffOfl|\fl\house)$\tabularnewline
\hline
\hline 
$0$ & $0.2$\tabularnewline
\hline 
$1$ & $0.5$\tabularnewline
\hline 
$>1$ & $0.8$\tabularnewline
\end{tabular}
\end{center}
\caption{The observation probabilities of a house $\house$.}
\label{tab:GFF_Oprobs}
\end{table}

The local reward induced by each house is depends on the fire level and the number of agents that 
chose to fight fire at that house. It is specified by
\[
R(\fl\house,\nrA_{\text{present}})=-\fl\house\mult0.7^{\nrA_{\text{present}}}. 
\]
As in \sec{2ffExample}, this reward can be transformed to a (in this case local) utility function
by taking the expectation with respect to the hidden state:
\[
 \utI{\house}(\jtypeG{\house},\jaG{\house} ) = \sum_{\fl\house = 1}^{\nrFLs} 
 \Pr( \fl\house | \jtypeG{\house} ) R(\fl\house, \text{CountAgentsAtHouse}(\fl\house,\jaG\house) ),
\]
where CountAgentsAtHouse() counts the number of agents for which $\jaG\house$ specifies to fight fire at 
house $\house$.

This formulation of $\GFF$, while still abstract, captures the essential coordination challenges inherent in many realistic problems.  For instance, this formulation may directly map to the problem of fire fighting in the Robocup Rescue Simulation league \citep{Kitano:99:ICSMC}: fire fighting agents are distributed in a city and must decide at which houses to fight fire. While limited communication may be possible, it is infeasible for each agent to broadcast all its observations to all the other agents. If instead it is feasible to compute a joint BG-policy based on the agents' positions, then they can effectively coordinate without broadcasting their observations.
When interpreting the houses as queues, it also directly corresponds to problems in queueing
networks~\citep{Cogill:04:approxDP}.

\begin{figure}[p]
\newcommand{\GFFfigSize}{0.42\columnwidth}
\noindent
~\hfill%
\subfloat[Runtime for varying number of actions when each agent observes 1 house (2 types per agent).]
{%
\includegraphics[width=\GFFfigSize]
{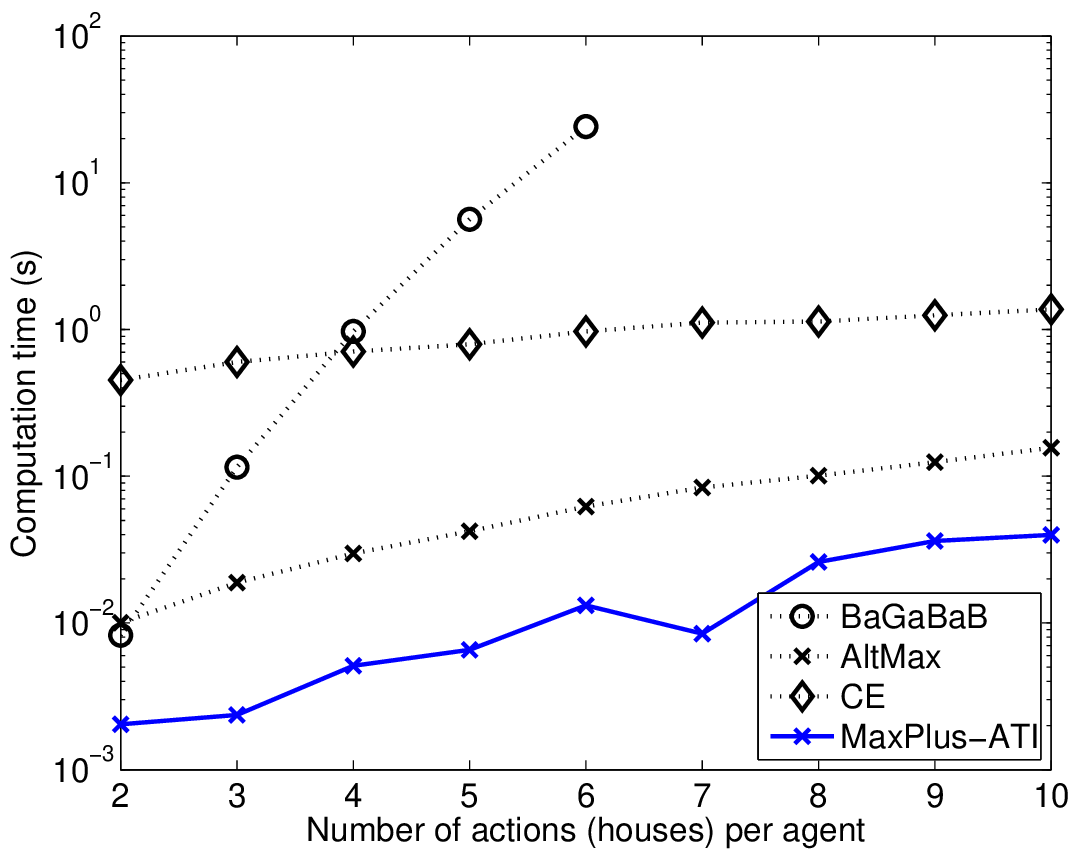}
\label{fig:GFF_varAcs_O1}
}%
\hfill%
\subfloat[Runtime for varying number of actions when each agent observes 2 houses (4 types per agent).]
{
\includegraphics[width=\GFFfigSize]
{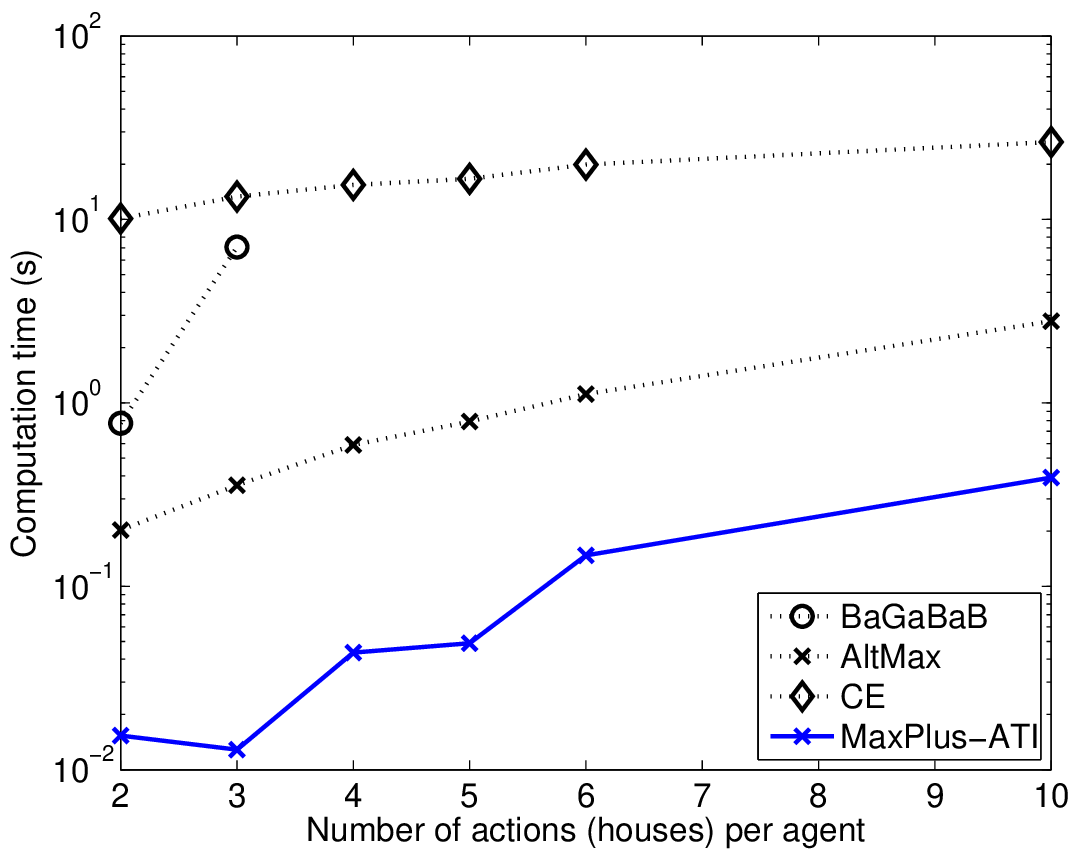}
\label{fig:GFF_varAcs_O2}
}%
\hfill~

~ \hfill%
\subfloat[The values for \fig{GFF_varAcs_O1}.
The value decreases since the number of houses
is proportional to the number of actions.]
{%
\includegraphics[width=\GFFfigSize]
{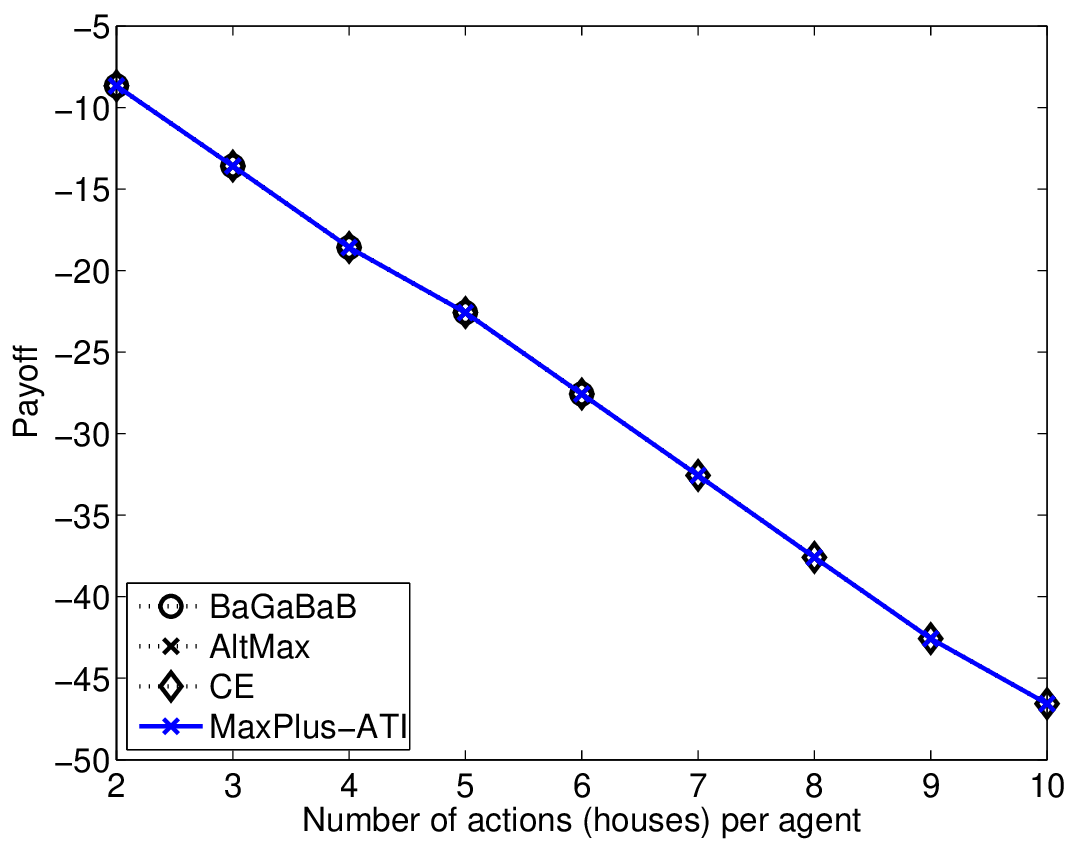}
\label{fig:GFF_varAcs_O1_Value}
}%
\hfill%
\subfloat[Varying number of agents. Each agent observes 2 houses (4 types per agent).]
{
\includegraphics[width=\GFFfigSize]
{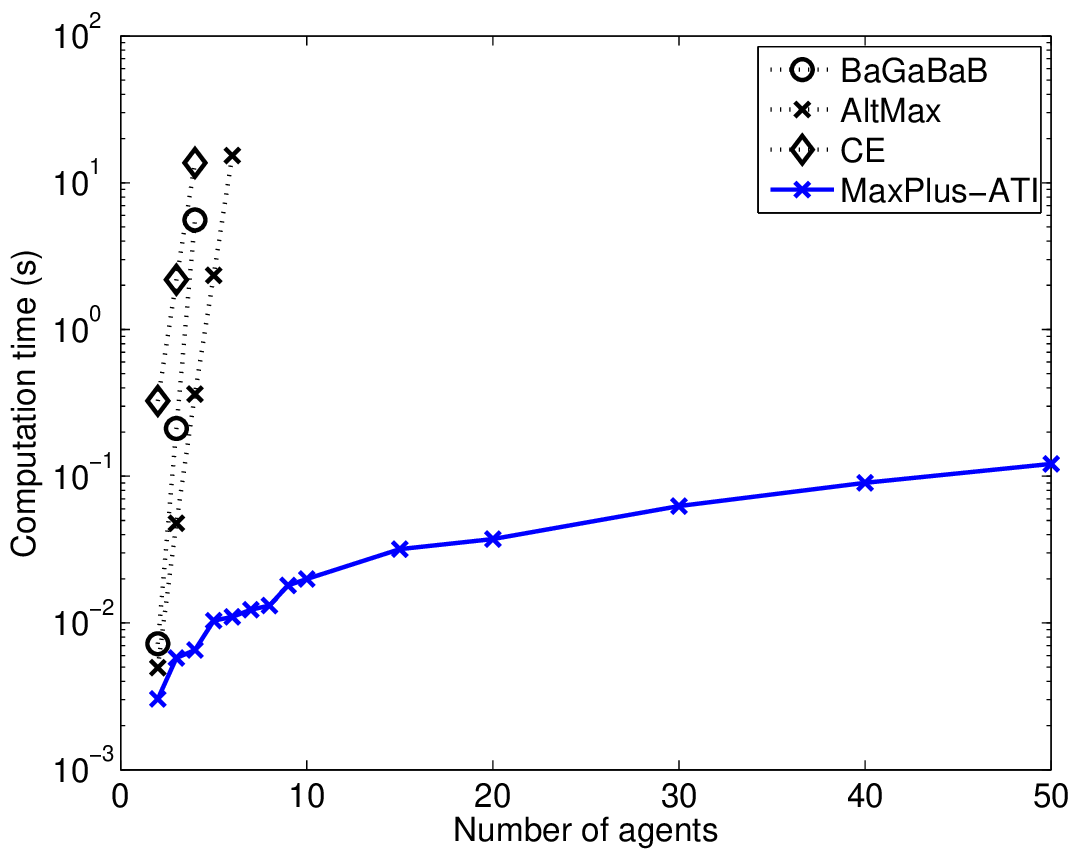}
\label{fig:GFF_varAgents}
}%
\hfill~%

~\hfill%
\subfloat[Runtime for varying $k$, the maximum number of agents that participate in a payoff component.
Results are for 4 agents with 3 actions and 2 observed houses (4
types), for different values of $\GFFdensity$.
]
{%
\includegraphics[width=\GFFfigSize]
{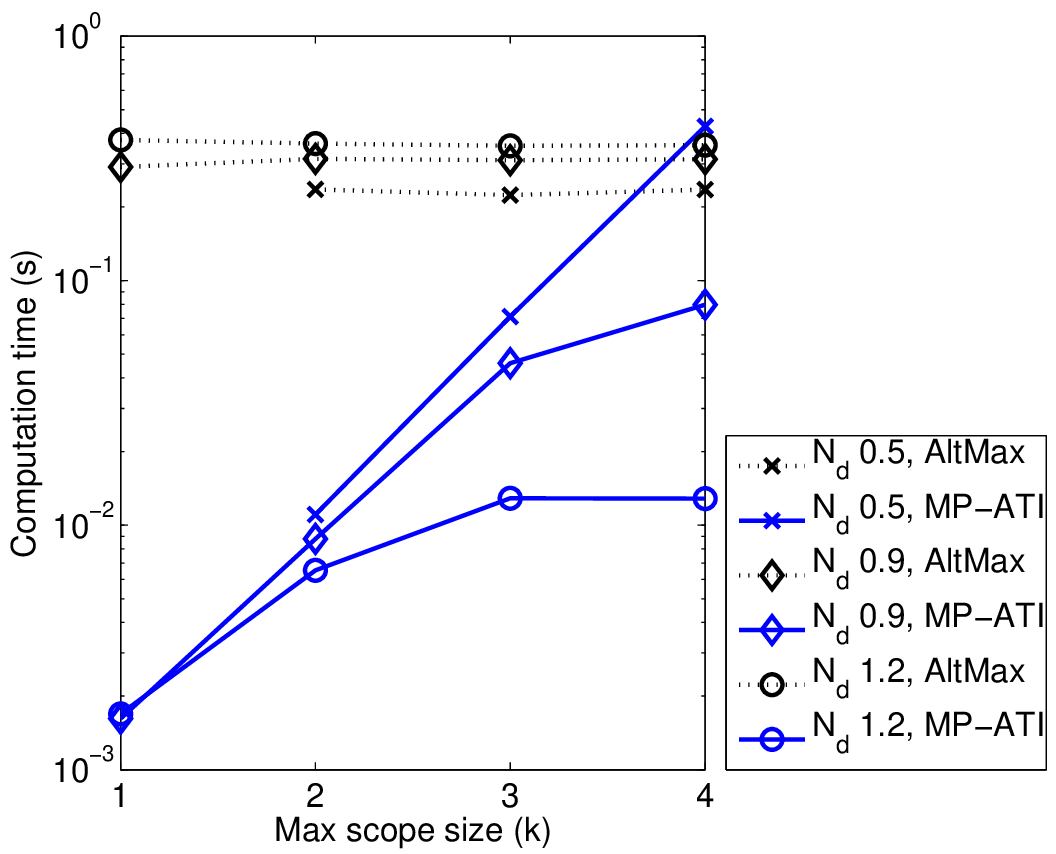}
\label{fig:GFF_varK}
}%
\hfill%
\subfloat[Runtime for varying number of houses that are observed per agent.
Note that the number of types is exponential in this number.
Results are for 4 agents with 5 actions.]
{
\includegraphics[width=\GFFfigSize]
{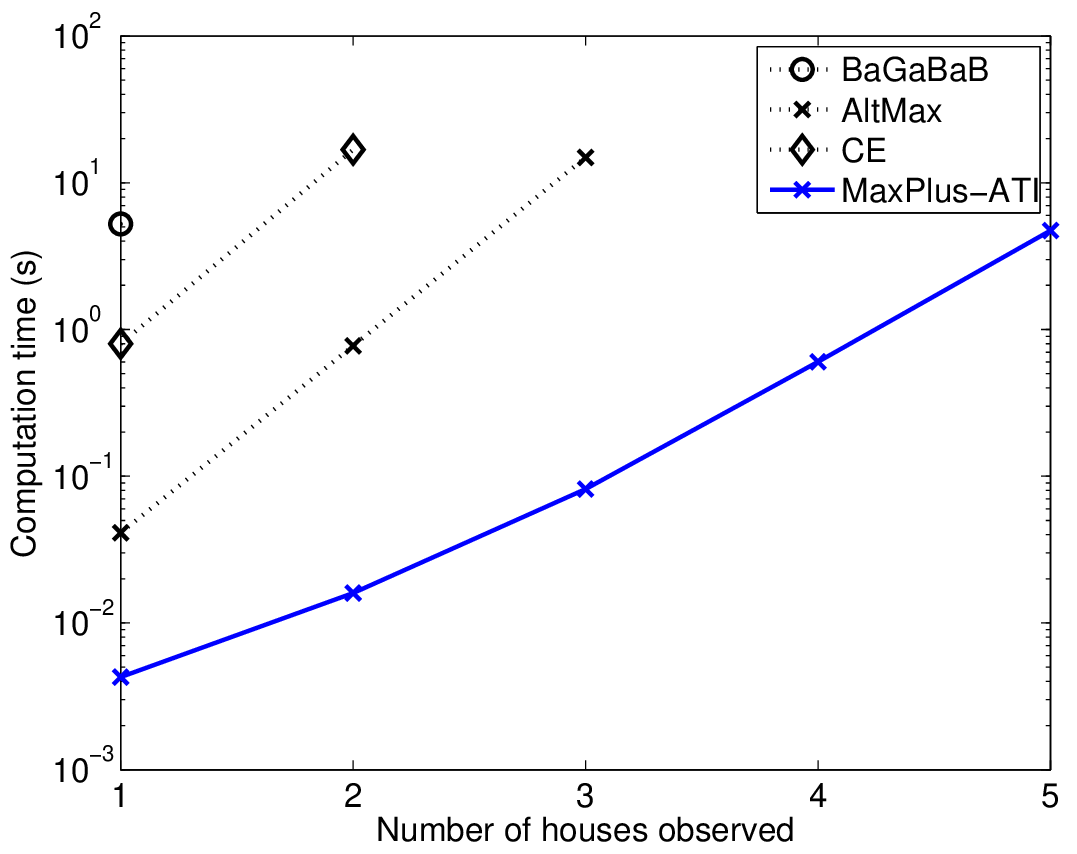}
\label{fig:GFF_varObs}
}
\hfill~
\caption{Results for the $\GFF$ problem.}
\label{fig:GFFresults}
\end{figure}

The results of our experiments in this domain, shown in \fig{GFFresults}, demonstrate that \<MP> has the
most desirable scaling behavior in a variety of different parameters.
In particular, \fig{GFF_varAcs_O1} and \ref{fig:GFF_varAcs_O2} show that all approximate
methods scale well with respect to the number of actions per agent, but \<MP>  
performs best. \fig{GFF_varAcs_O1_Value} shows that this scalability does not come at the expense of
solution quality. For all settings, all the methods computed
solutions with the same value (other plots of value are thus omitted for brevity).
The advantage of exploiting agent independence is illustrated in \fig{GFF_varAgents},
which demonstrates that \<MP> scales well with the number of agents, in contrast to the other methods.
In \fig{GFF_varK} we varied the $\GFFdensity$
parameter, which determines how many houses are present in the
domain. It shows that \<MP> is sensitive to $k$, the maximum number
of agents that participate in a house.  However, it also demonstrates that when the interaction is sparse (i.e.,
when there are many houses per agent and therefore on average fewer than $k$ agents per house) the increase
in runtime is much better than the worst-case exponential growth.%
\footnote{The dense setting $\GFFdensity=0.5$
does not have data points at $k=1$ because in this case there are not enough houses to be assigned to the agents.}
\fig{GFF_varObs} shows how runtime scales with $\ffnrObs$, the number of houses observed
by each agent. Since the number of types is exponential in $\ffnrObs$,  \<MP>'s runtime
is also exponential in $\ffnrObs$.  Nonetheless, it substantially outperforms the other
approximate methods.

\providecommand{\utET}[2]{ \utF_{#1}^{#2} }
\section{Exploiting Independence in Dec-POMDPs}
\label{sec:DecPOMDPs}

While CBGs model an important class of collaborative decision-making problems, they apply only to \emph{one-shot} settings, i.e., where each agent needs to select only one action.  However, the methods for exploiting agent and type independence that we proposed in \sec{CGBGs} can also provide substantial leverage in \emph{sequential} tasks, in which agents take a series of actions over time as the state of the environment evolves. In particular, many sequential collaborative decision-making tasks can be formalized as \emph{decentralized partially observable Markov decision processes (Dec-POMDPs)}~\citep{Bernstein:02:Complexity}.
In this section we demonstrate how CGBGs can be used in a planning method for the subclass of \emph{factored} Dec-POMDPs with additively factored rewards \citep{Oliehoek:08:AAMAS}. The resulting method, called \FSPCF, can find approximate solutions for classes of problems that cannot be addressed at all by any other planning methods.

Our aim is not to present $\FSPCF$ as a main contribution of this article.  Instead, we merely use it as a vehicle for illustrating the utility of the CGBG framework. Therefore, for the sake of conciseness, we do not describe this method in full technical detail.  Instead, we merely sketch the solution approach and supply references to other work containing more detail.  

\subsection{Factored Dec-POMDPs}
\label{sec:DecPOMDPmodel}

In a Dec-POMDP, multiple agents must collaborate to maximize the sum of the common rewards they receive over multiple timesteps.  Their actions affect not only their immediate rewards but also the state to which they transition.  While the current state is not known to the agents, at each timestep each agent receives a private observation correlated with that state.  In a factored Dec-POMDP, the state consists of a vector of state variables and the reward function is the sum of a set of local reward functions.

\begin{definition} \label{def:Dec-POMDP}A factored Dec-POMDP
is a tuple $\left\langle \agentS,\mathcal{S},\mathcal{A},T,\mathcal{R},\mathcal{O},O,\bO,h\right\rangle $,
where
\begin{itemize}
\item $\agentS=\{\agentI1,\dots,\agentI\nrA\}$ is the set of agents.
\item $\sS=\sfacIS1\times\ldots\times\sfacIS\nrSF$ is the factored state space.
That is, $\sS$ is spanned by $\sfacS=\left\{ \sfacIS1,\dots,\sfacIS\nrSF\right\}$,
a set of state variables, or \emph{factors.} A state corresponds to
an assignment of values for all factors $s=\left\langle \sfacI1,\dots,\sfacI\nrSF\right\rangle $.
\item $\jaS=\times_{i}\aAS i$ is the set of \emph{joint actions}, where
$\aAS i$ is the set of actions available to agent~$i$.
\item $T$ is a transition function specifying the state transition probabilities $\Pr(s'|s,\ja)$.
\item $\mathcal{R}= \{R^1,\dots,R^\rho\}$ is the set of $\nrR$ local reward functions.
Again, these correspond to a interaction graph with (hyper) edges $\edS$
such that the total immediate reward $R(s,\ja)= \sum_{\ed\in\edS}R^e({\bf x}_e,\jaG{\ed}).$
\item $\joS=\times_{i}\oAS i$ is the set of joint observations
$\jo=\left\langle \oA1,...,\oA\nrA\right\rangle $.
\item $O$ is the observation function, which specifies observation probabilities $\Pr(\jo|\ja,s')$.
\item $\bO$
 is the initial state distribution at time
$\ts=0$.
\item $h$ is the horizon, i.e., the number of stages. We consider the case where
    $h$ is finite.
\end{itemize}
\end{definition}

At each stage $\ts=0\dots h-1$, each agent takes an individual action
and receives an individual observation. Their goal is to maximize the
expected \emph{cumulative reward} or \emph{return}. The planning task
entails finding a \emph{joint} policy $\jpol=\left\langle \polA1,\dots,\polA\nrA\right\rangle $,
that specifies an individual policy $\polA i$ for each agent $i$.
Such an individual policy in general specifies an individual action
for each action-observation history $\oaHistAT i\ts=(\aAT i0,\oAT i1,\dots,\aAT i{\ts-1},\oAT i\ts)$,
e.g., $\polA i(\oaHistAT i\ts)=\aAT i\ts$. 
However, when only allowing \emph{deterministic}
or \emph{pure} policies, $\polA i$ maps each observation history
$(\oAT i1,\dots,\oAT i\ts)=\oHistAT i\ts\in\oHistATS i\ts$ to an
action, e.g., $\polA i(\oHistAT i\ts)=\aAT i\ts$. 
In a factored Dec-POMDP, the transition and observation model can
be compactly represented in a \emph{dynamic Bayesian network (DBN)} \citep{Boutilier:99:JAIR}.

\subsubsection{Sequential Fire Fighting}
\label{sec:sff}

As a running example we consider the $\FFG$ problem, which is a sequential variation of $\GFF$ from \sec{CGBGmodel}, originally introduced by \cite{Oliehoek:08:AAMAS}.\footnote{In \citep{Oliehoek:08:AAMAS}, this problem is referred to as \problemName{FireFightingGraph}.  We use a different name here to distinguish it from $\GFF$, which is also graphical.} In this version, each house, instead of simply being on fire or not, has an integer \emph{fire level} that can change over time. 
Thus, the state of the environment is factored, using one state variable for the fire level in each house. 
Each agent receives an observation about the house at which it fought fire in the last stage.
It observes flames ($\ffOfl$) at this house with probability $0.2$ if $\fl{\house}=0$, with probability $0.5$
if $\fl{\house}=1$, and with probability $0.8$ otherwise. 

At each stage, each agent~$i$ chooses at which of its assigned houses to fight fire.  
These actions affect (probabilistically) to what state the environment transitions: 
the fire level of each house $\house$ is influenced by its previous value, by the actions of the
agents that can go  to $\house$ and the fire level of the neighboring houses.
The transitions in turn determine what reward is generated.
Specifically, each house generates a negative reward equal to its expected fire level at the next stage $\fl1'$.  
Thus, the reward function can be described as the sum of local reward functions, one for each house.  
We consider the case of $\nrHouses=4$ as shown in \fig{nffExample}. 
For house $H=1$ the reward is specified by\begin{equation}
\REd1(\flG{\left\{ 1,2\right\} },\aA1)=\sum_{\fl1'}-\fl1'\Pr(\fl1'|\flG{\left\{ 1,2\right\} },\aA1),
\label{eq:Red-back-projected}
\end{equation}
where $\flG{\left\{ 1,2\right\} }$ denotes $\left\langle \fl1,\fl2\right\rangle $.
This formulation is possible because
$\fl1,\fl2$ and $\aA1$ are the only variables that influence the
probability of $\fl1'$. Similarly, the other local reward
functions are given by $\REd2(\flG{\left\{ 1,2,3\right\} },\jaG{\left\{ 1,2\right\} })$,
$\REd3(\flG{\left\{ 2,3,4\right\} },\jaG{\left\{ 2,3\right\} })$
and $\REd4(\flG{\left\{ 3,4\right\} },\aA3)$. For more details about the formulation of $\FFG$ as a factored Dec-POMDP, see \citep{Oliehoek:10:PhD}.

\subsubsection{Solving Dec-POMDPs}
Solving a Dec-POMDP entails finding an optimal joint policy.
Unfortunately, optimally solving Dec-POMDPs is NEXP-complete \citep{Bernstein:02:Complexity},
as is finding an $\eps$-ap\-pro\-xi\-mate solution
\citep{Rabinovich:03:AAMAS}. 

Given these difficulties, most research efforts have focused
on special cases that are more tractable. 
In particular, assumptions of \emph{transition and observation independence
(TOI)} \citep{Becker:04:JAIR} have been investigated to exploit independence
between agents, e.g., as in ND-POMDPs \citep{Nair:05:AAAI,Varakantham:07:AAMAS}. However, given the TOI assumption,
many interesting tasks, such as two
robots carrying a chair, cannot be modeled.  Recently, \cite{Witwicki:10:ICAPS} 
proposed \emph{transition-decoupled} Dec-POMDPs, in which there is limited interaction between the agents. 
While this approach speeds up both optimal and approximate solutions of this subclass, scalability remains limited 
(the approach has not been tested with more than two agents)
and the sub-class still is quite restrictive (e.g., it does not admit the chair carrying scenario).

Other work has considered approximate methods for the general class of Dec-POMDPs based on 
representing a Dec-POMDP using CBGs \citep{Emery-Montemerlo:04:AAMAS} or on 
approximate dynamic programming \citep{Seuken:07:IJCAI:MBDP}.
In the remainder of this section, we show CGBGs can help the former category of methods achieve unprecedented scalability with respect to the number of agents.

\subsection{Factored Dec-POMDPs as Series of CGBGs}

A stage $t$ of a Dec-POMDP can be represented as a CBG, given a \emph{past joint policy} $\pJPolT{t}$.
Such a  $\pJPolT{t}=( \jdrT{0},\dots,\jdrT{t-1} )$ specifies the joint \emph{decision rules} for the first $t$ stages.
An individual decision rule $\drAT{i}{t}$ of agent~$i$ corresponds to the part of its policy that specifies actions for stage~$t$.
That is, $\drAT{i}{t}$ maps from observation histories $\oHistAT{i}{t}$ to actions $\aA{i}$.
Given a $\pJPolT{t}$, the corresponding CGB is constructed as follows:
\begin{itemize}
\item The action sets are the same as in the Dec-POMDP,
\item Each $i$'s action-observation history~$\oaHistAT i\ts$ corresponds to its type: $\typeA i\defas\oaHistAT i\ts$,
\item The probability of joint types is specified by $\Pr(\oaHistT\ts|\bO,\pJPolT{\ts})=\sum_{\s^{\ts}}\Pr(\s^{\ts},\oaHistT\ts|\bO,\pJPolT{\ts})$.
\item The payoff function $\utF(\jtype,\ja) = Q(\oaHistT\ts,\ja)$, the expected payoff for the remaining stages.
\end{itemize}

Similarly, a factored Dec-POMDP can be represented by a series of CGBGs. This is possible because, in general, the Q-function is factored.  
In other words, it is the sum of a set of \emph{local Q-functions} of the following form: $\QEd{\ed}(\sfacGT{\sfe}\ts,\oaHistGT{\age}\ts,\jaG{\age})$.
Each local Q-function depends on a subset of state factors (the state factor
scope) and the action-observation histories and actions of a subset
of agents (the agent scope)~\citep{Oliehoek:10:PhD}. 

Constructing a CGBG for a stage~$\ts$ is similar to constructing a CBG.  
Given a past joint policy $\pJPolT\ts$, we can construct the local payoff functions for the CGBG: 
\begin{equation}
\utI\ed(\jtypeG{\age},\jaG{\age})\defas\QEdP{\ed}{\pJPolT{\ts}}(\oaHistGT{\age}{\ts},\jaG{\age})=\sum_{\sfacGT{\age}\ts}\Pr(\sfacGT{\sfe}\ts|\oaHistGT{\age}\ts,\bO,\pJPolT{\ts})\QEd{\ed}(\sfacGT{\sfe}\ts,\oaHistGT{\age}\ts,\jaG{\age}).
\label{eq:QEd(oaH,a)-definition}
\end{equation}
As such, the structure of the CGBG is induced by the structure of the Q-value function.

As an example, consider 3-agent $h=2$ $\FFG$.
The last stage $t=1$ can be represented as a CGBG given a past joint policy $\pJPolT{1}$.
Also, since it is the last stage, the factored immediate reward function, (e.g., as in Equation \ref{eq:Red-back-projected})
represents all the expected future reward. That is, it coincides with an optimal factored Q-value function \citep{Oliehoek:08:AAMAS}
and can be written as follows:
{\renewcommand{\ts}{1}
\begin{equation}
\QEdP{1}{\pJPolT{\ts}}(\oaHistGT{\age}{\ts},\jaG{\age})
=
\sum_{\sfacGT{\age}\ts}
\Pr(\sfacGT{\sfe}\ts | \oaHistGT{\age}\ts,\bO,\pJPolT{\ts})
\REd{1}(\sfacGT{\sfe}\ts,\jaG{\age}).
\label{QEd(oaH,a)}
\end{equation}
}
This situation can be represented using a CGBG, by using the Q-value function as the payoff function:
$\utI\ed(\jtypeG{\age},\jaG{\age})\defas\QEdP{\ed}{\pJPolT{\ts}}(\oaHistGT{\age}{\ts},\jaG{\age})$
as shown in \fig{FFG_CGBG}. The figure shows the 4 components of the CGBG, each one corresponding to the payoff associated with one house.
It also indicates an arbitrary BG policy for agent~2.
Note that, since components 1 and 4 of the Q-value function have scopes that are `subscopes' of components 2 and 3 respectively,
the former can be absorbed into the latter, reducing the number of components without increasing the size of those that remain.

\begin{figure}[tb]
 
\noindent {\noindent
\small
\renewcommand{\multirowsetup}{\centering}
\setlength\arrayrulewidth{\thickertableline}\arrayrulecolor{black}
\setlength{\tabcolsep}{4pt}
\setlength{\firstcol}{0.5cm}

\noindent \begin{center}
\begin{tabular}{cc|c}
\multicolumn{3}{c}{}\tabularnewline
\multicolumn{3}{c}{}\tabularnewline
$\typeA1$ & & \tabularnewline
\hline
\multirow{2}{\firstcol}{$\left(\ffOfl\right)$} & $\ffA1$ & $-0.25$\tabularnewline
& $\ffA2$ & $-1.10$\tabularnewline
\hline
\multirow{2}{\firstcol}{$\left(\ffOnf\right)$} & $\ffA1$ & $-0.14$\tabularnewline
& $\ffA2$ & $-0.79$\tabularnewline
\multicolumn{3}{c}{}\tabularnewline
\multicolumn{3}{c}{$Q^{\ed=1},\ \agScopeSymb=\{1\}$}\tabularnewline
\end{tabular}\hspace{0.05\columnwidth}\begin{tabular}{cc|cc|cc}
\multicolumn{2}{c}{} & \multicolumn{4}{c}{$\typeA2$}\tabularnewline
& & \multicolumn{2}{c|}{$\left(\ffOfl\right)$} & \multicolumn{2}{c}{$\left(\ffOnf\right)$}\tabularnewline
$\typeA1$ & & $\ffA2$ & \cellcolor{tablemid} $\ffA3$ & \cellcolor{tablemid} $\ffA2$ & $\ffA3$\tabularnewline
\hline
\multirow{2}{\firstcol}{$\left(\ffOfl\right)$} & $\ffA1$ & $-0.55$ & \cellcolor{tablemid} $-1.60$ & \cellcolor{tablemid} $-0.50$ & $-1.5$0\tabularnewline
& $\ffA2$ & $0$ & \cellcolor{tablemid} $-0.55$ & \cellcolor{tablemid} $0$ & $-0.50$\tabularnewline
\hline
\multirow{2}{\firstcol}{$\left(\ffOnf\right)$} & $\ffA1$ & $-0.16$ & \cellcolor{tablemid} $-1.10$ & \cellcolor{tablemid} $-0.14$ & $-1.00$\tabularnewline
& $\ffA2$ & $0$ & \cellcolor{tablemid} $-0.16$ & \cellcolor{tablemid} $0$ & $-0.14$\tabularnewline
\multicolumn{6}{c}{}\tabularnewline
\multicolumn{6}{c}{$Q^{\ed=2},\ \agScopeSymb=\left\{ 1,2\right\} $}\tabularnewline
\end{tabular}
\par\end{center}

\noindent \begin{center}
\begin{tabular}{cc|c}
\multicolumn{3}{c}{}\tabularnewline
& \multicolumn{1}{c}{} & \multicolumn{1}{c}{}\tabularnewline
$\typeA3$ & & \tabularnewline
\hline
\multirow{2}{\firstcol}{$\left(\ffOfl\right)$} & $\ffA3$ & $-1.50$\tabularnewline
& $\ffA4$ & $-0.51$\tabularnewline
\hline
\multirow{2}{\firstcol}{$\left(\ffOnf\right)$} & $\ffA3$ & $-1.10$\tabularnewline
& $\ffA4$ & $-0.15$\tabularnewline
\multicolumn{3}{c}{}\tabularnewline
\multicolumn{3}{c}{$Q^{\ed=4},\ \agScopeSymb=\left\{ 3\right\} $}\tabularnewline
\end{tabular}\hspace{0.05\columnwidth}\begin{tabular}{cc|cc|cc}
& \multicolumn{1}{c}{} & \multicolumn{4}{c}{$\typeA2$}\tabularnewline
& & \multicolumn{2}{c|}{$\left(\ffOfl\right)$} & \multicolumn{2}{c}{$\left(\ffOnf\right)$}\tabularnewline
$\typeA3$ & & $\ffA2$ & \cellcolor{tablemid} $\ffA3$ & \cellcolor{tablemid} $\ffA2$ & $\ffA3$\tabularnewline
\hline
\multirow{2}{\firstcol}{$\left(\ffOfl\right)$} & $\ffA3$ & $-1.10$ & \cellcolor{tablemid} $0$ & \cellcolor{tablemid} $-0.71$ & $0$\tabularnewline
& $\ffA4$ & $-1.90$ & \cellcolor{tablemid} $-1.10$ & \cellcolor{tablemid} $-1.70$ & $-0.71$\tabularnewline
\hline
\multirow{2}{\firstcol}{$\left(\ffOnf\right)$} & $\ffA3$ & $-1.00$ & \cellcolor{tablemid} $0$ & \cellcolor{tablemid} $-0.58$ & $0$\tabularnewline
& $\ffA4$ & $-1.90$ & \cellcolor{tablemid} $-1.00$ & \cellcolor{tablemid} $-1.60$ & $-0.58$\tabularnewline
\multicolumn{6}{c}{}\tabularnewline
\multicolumn{6}{c}{$Q^{\ed=3},\ \agScopeSymb=\left\{ 2,3\right\} $}\tabularnewline
\end{tabular}
\par\end{center}
}
\caption{
A CGBG for $t=1$ of $\FFG$. Given a past joint policy $\pJPolT{1}$, each joint type $\jtype$ corresponds to 
a joint action-observation history $\oaHistT{1}$. The entries give the Q-values $\QEd\ed(\oaHistGT{\ed}{t},\jaG{\ed} )$. Highlighted is an arbitrary
policy for agent~2.
}
\label{fig:FFG_CGBG}
\end{figure}

The following theorem by \cite{Oliehoek:08:AAMAS}, shows that modeling a factored Dec-POMDP in this way is in principle exact.
\begin{theorem}
Modeling a factored Dec-POMDP with additive rewards using a series
of CGBGs is exact: it yields the optimal solution when using an optimal Q-value function.
\end{theorem}

\begin{figure}
\centering
\psfrag{t0}[tc][cc]{$h-3$}
\psfrag{t1}[tc][cc]{$h-2$}
\psfrag{t2}[tc][cc]{$h-1$}
\psfrag{FL1}[tc][cc]{$\fl1$}
\psfrag{FL1'}[tc][cc]{$\fl1'$}
\psfrag{a1}[tc][cc]{$\aA{1}$}
\psfrag{o1}[tc][cc]{$\oA{1}$}
\psfrag{R1}[tc][cc]{$\REd{1}$}
\psfrag{FL2}[tc][cc]{$\fl2$}
\psfrag{FL2'}[tc][cc]{$\fl2'$}
\psfrag{a2}[tc][cc]{$\aA{2}$}
\psfrag{o2}[tc][cc]{$\oA{2}$}
\psfrag{R2}[tc][cc]{$\REd{2}$}
\psfrag{FL3}[tc][cc]{$\fl3$}
\psfrag{FL3'}[tc][cc]{$\fl3'$}
\psfrag{a3}[tc][cc]{$\aA{3}$}
\psfrag{o3}[tc][cc]{$\oA{3}$}
\psfrag{R3}[tc][cc]{$\REd{3}$}
\psfrag{FL4}[tc][cc]{$\fl4$}
\psfrag{FL4'}[tc][cc]{$\fl4'$}
\psfrag{R4}[tc][cc]{$\REd{4}$}
\includegraphics[scale=0.19]{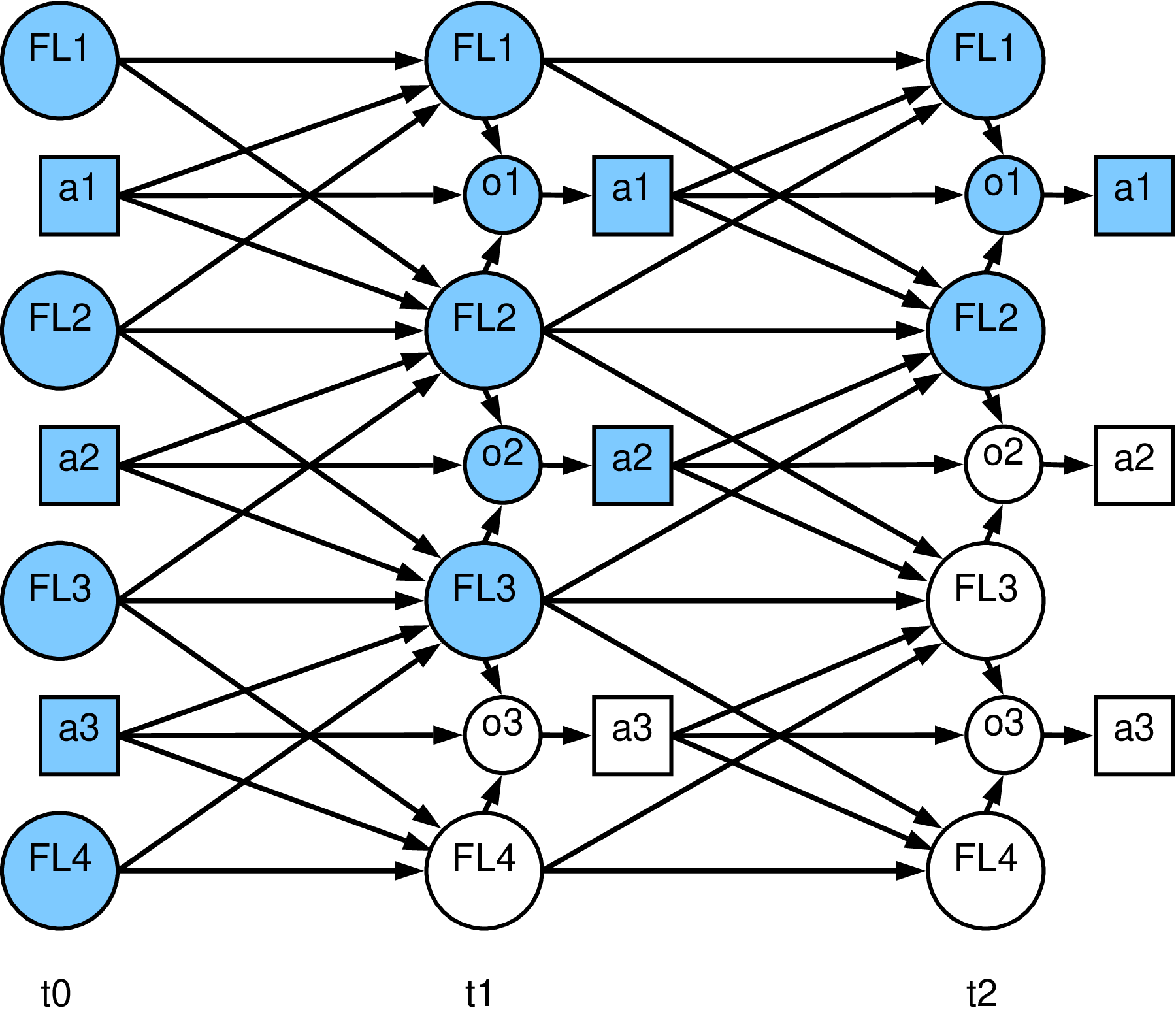}
\caption{The scope of $\QEd{1}$,
  illustrated by shading, increases when going back in time in $\FFG$.}
\label{fig:FFF-scope}
\end{figure}

While an optimal Q-value function is factored, 
the last stage contains the most independence:
when moving back in time towards $\ts=0$,
the scope of dependence grows, due to the transition and observation functions.
\fig{FFF-scope} illustrates this process in \<FFG>.  Thus, even though the value function is factored, the
scopes of its components may at earlier stages include all state factors and agents.

\subsection{Factored Forward Sweep Policy Computation}

Defining the payoff function for each CBG representing a stage of a Dec-POMDP requires computing 
$Q(\oaHistT\ts,\ja)$, an optimal payoff function.
Unfortunately, doing so is intractable.  
Therefore, Dec-POMDP methods based on CBGs typically use approximate Q-value functions instead. 
One option ($\QBG$) is to assume that each agent always has access to the joint observations for all previous stages, but can access only its individual observation for the current stage.
Another choice is based on the underlying POMDP ($\QPOMDP$), i.e., a POMDP with the same transition 
and observation function in which a single agent receives the joint observations and takes joint actions.  
A third option is based on the underlying MDP ($\QMDP$), in which this single agent can directly observe the state.

Given an approximate Q-function, an approximate solution for the Dec-POMDP can be computed via 
\emph{forward-sweep policy computation} ($\FSPC$) by simply solving the CBGs for stages $0,1,\dots,h-1$ 
consecutively. The solution to each CBG is a joint decision rule specified by $\jdrT\ts\defas\jpolBG^{\ts,*}$ 
that is used to augment the past policy $\pJPolT{\ts+1} = (\pJPolT{\ts}, \jdrT\ts )$. 
The CBGs are solved consecutively because both the probabilities and the payoffs at each stage depend on the past policy. 
It is also possible to compute an optimal policy via backtracking, as in \emph{multiagent} $A^*$ ($\MAA$).  
However, since doing so is much more computationally intensive, we focus on $\FSPC$ in this article.

In the remainder of this section, we describe a method we call $\FSPCF$ for approximately solving
a broader class of factored Dec-POMDPs in a way that scales well in the number
of agents.  The main idea is simply to replace the CBGs used in each stage of the Dec-POMDPs with CGBGs, which are then solved using the methods presented in \sec{CGBGs}.

Since finding even bounded approximate solutions for Dec-POMDPs is NEXP-complete, 
any computationally efficient method must necessarily make unbounded approximations. 
Below, we describe a set of such approximations designed to make $\FSPCF$ a practical algorithm.  
While we present only an overview here, complete details are available in \citep{Oliehoek:10:PhD}. 
Also, though the results we present in this article evaluate only the complete resulting method, 
\cite{Oliehoek:10:PhD} empirically evaluated each of the component approximations separately.

\subsubsection{Approximate Inference}

One source of intractability in $\FSPCF$ lies in the marginalization required to compute the probabilities $\Pr(\oaHistGT{\age}\ts|\bO,\pJPolT{\ts})$
and $\Pr(\sfacGT{\sfe}\ts|\oaHistGT{\age}\ts,\bO,\pJPolT{\ts})$.  In particular, constructing each CGBG requires generating each component $\ed$ separately. However, 
as \eq{QEd(oaH,a)-definition} shows, in general this requires the probabilities 
$\Pr(\sfacGT{\sfe}\ts|\oaHistGT{\age}\ts,\bO,\pJPolT{\ts})$.
Moreover, in any efficient solution algorithm for Dec-POMDPs, the probabilities
$\Pr(\oaHistGT{\age}\ts|\bO,\pJPolT{\ts})$ are necessary, as illustrated by 
\eq{CGBGsolution}.
Since maintaining and marginalizing over $\Pr(\s,\oaHistT\ts|\bO,\pJPolT\ts)$ is intractable, we resort to approximate inference, as is standard practice when computing probabilities over states with many factors. Such methods perform well in many cases
and the error they introduce can in some cases be theoretically bounded \citep{Boyen+Koller:98:UAI}.

In our case, we use the factored frontier (FF) algorithm \citep{Murphy:01:UAI} to perform approximate inference
on a DBN constructed for the past joint policy $\pJPolT\ts$ under
concern. This DBN models stages $0,\dots,\ts$ and has both state factors
and action-observation histories as its nodes. We use FF because it
is simple and allows computation of some useful intermediate representations
when a heuristic of the form $\QEd{\ed}(\sfacGT{\sfe}\ts,\jaGT{\age}{\ts})$
(e.g., factored $\QMDP$) is used. Other approximate inference algorithms (e.g., \citealt{Murphy:02:PHD,Mooij:08:PhD}), could also be used.

\subsubsection{Approximate Q-Value Functions}
\label{sec:FactQs}

Computing the optimal value functions to use as payoffs for the CGBG for each stage is intractable.  For small Dec-POMDPs, heuristic payoff functions such as $\QMDP$ and $\QPOMDP$ are typically used instead \citep{Oliehoek:08:JAIR}.  However, for Dec-POMDPs of the size we consider in this article, solving the underlying MDP or POMDP is also intractable.  

Furthermore, factored Dec-POMDPs pose an additional problem: the scopes of $Q^{*}$ increase when going backwards in time, such that they are typically fully coupled for earlier stages (see \fig{FFF-scope}).  This problem is exacerbated when $\QMDP$ and $\QPOMDP$ are used as heuristic payoff functions because they become fully coupled through just one backup (due to the maximization
over joint actions that is conditioned on the state or belief). 

Fortunately, many researchers have considered factored approximations for factored MDPs and factored POMDPs \citep{Schweitzer:85:JMAA,Koller:99:IJCAI,Koller:00:UAI,Schuurmans:01:NIPS,Guestrin:01:IJCAI,Guestrin:01:PuUII,Guestrin:03:JAIR,Farias:03:OR,Szita:08:AC}.  
We follow a similar approach for Dec-POMDPs by using value functions with predetermined approximate scopes. The idea is that in many cases the influence of a state factor quickly vanishes with the number of links in the DBN. 
For instance, in the case of transition and observation independence (TOI), the optimal scopes equal those of the factored immediate reward function. 
In many cases where there is no complete TOI, the amount of interaction is still limited, 
making it possible to determine a 
reduced set of scopes for each stage that affords a good approximate solution. 
For example, consider the optimal scopes shown in \fig{FFF-scope}.  
Though $x_4$ at $h-3$ can influence $x_2$ at $h-1$, the magnitude of this influence is likely to be small.  
Therefore, restricting the scope of $Q^1$ to exclude $x_4$ at $h-3$ is a reasonable approximation.

Following the literature on factored MDPs, we use manually specified scope structures (this is equivalent to specifying basis functions). 
In the experiments presented in this article, we simply use the immediate reward scopes at each stage, though many alternative strategies are possible.
While developing methods for finding such scope structures automatically is an important goal, it is beyond the scope of this article. 
A heuristic approach suffices to validate the utility of the CGBG framework because our methods require only 
a good approximate factored value function whose scopes preserve \emph{some} independence. 

To compute a heuristic given a specified scope structure,
we  use an approach we call \emph{transfer planning}.  Transfer planning is motivated by the observation that, for a factored Dec-POMDP, the value function is `more factored' than for a factored MDP. 
In the former, dependence propagates over time, while the latter becomes fully coupled through just one backup.
Therefore, it may be preferable to directly approximate the
factored Q-value function of the Dec-POMDP rather than the  $\QMDP$ function.  
To do so, we use the solution of smaller \emph{source} problems that involve fewer agents.
That is, transfer planning directly tries to find heuristic values 
$
\QEdP{\ed}{\pJPolT{\ts}}(\oaHistGT{\age}{\ts},\jaG{\age})\defas\Q^{\tpSP}(\oaHistT\ts,\ja)
$
by solving tasks that are similar (but smaller) and using their value functions $\Q^{\tpSP}$.
The  $\Q^{\tpSP}$ can result from the solutions of the smaller Dec-POMDPs, or of their underlying MDP or POMDP.
In order to map the values $\Q^{\tpSP}$ of the source tasks to the CGBG components $\QEdP{\ed}{\pJPolT{\ts}}$,
we specify a mapping from agents participating in a component $\ed$ to agents in the source problem.

Since no formal claims can be made about these approximate Q-values, we cannot guarantee that they constitute an admissible heuristic.  
However, since we rely on $\FSPC$, which does not backtrack, an admissible heuristic is not necessarily better.  
Performance depends on the accuracy, not the admissibility of the heuristic \citep{Oliehoek:08:JAIR}.  
The experiments we present below demonstrate that these approximate Q-values are accurate enough to enable high quality solutions.

\subsection{Experiments}

\label{sec:FactDPOMDPs:Experiments}

\providecommand{\DICEN}{{\DICE\text{-normal}}}
\providecommand{\DICEF}{\DICE\text{-fast}}

\begin{figure*}
\begin{centering}
\includegraphics[width=0.2\paperwidth]{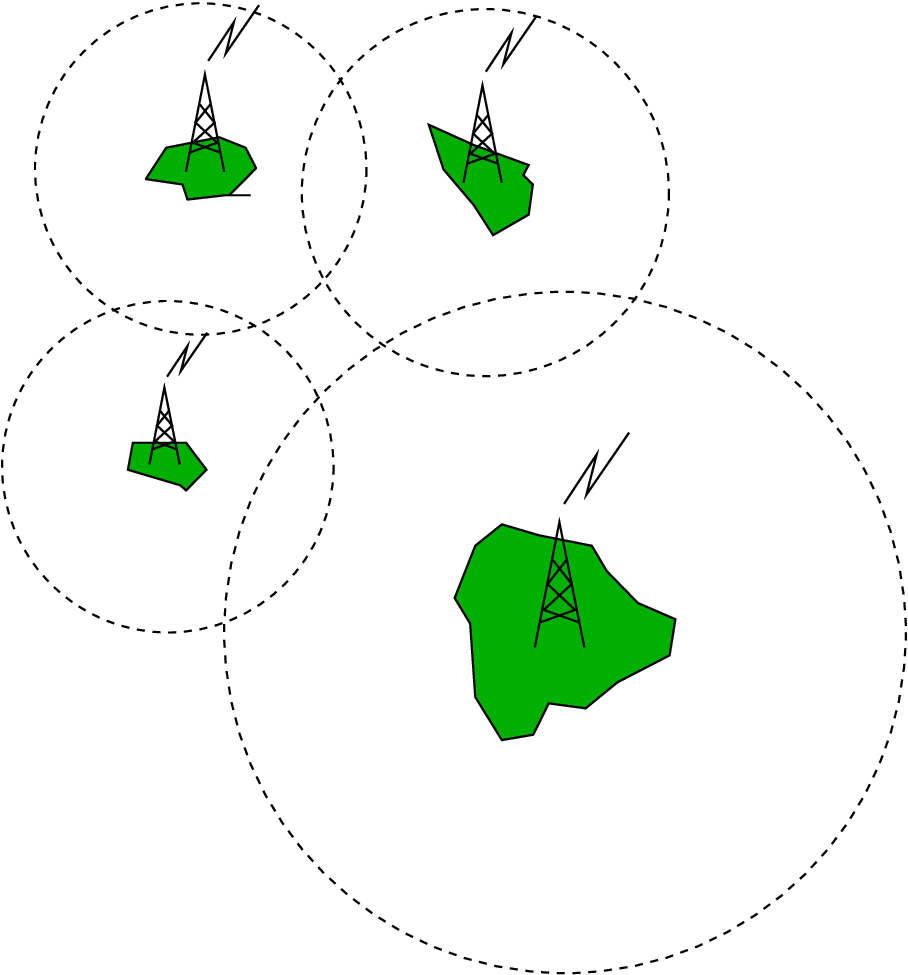}
\caption{The $\Aloha$ problem with four islands
arranged in a square. }
\label{fig:Aloha}
\end{centering}
\end{figure*}

We evaluate $\FSPCF$ on two problem domains: $\FFG$ and the
$\Aloha$ problem \citep{Oliehoek:10:PhD}.
The latter consists of a number of islands, each equipped with a
radio used to transmit messages to its local population.
Each island has
a queue of messages that it needs to send and at each time step
can decide whether or not to send a message. 
When two neighboring islands attempt to send a message in the same timestep, a
collision occurs.
Each island can noisily observe whether 
a successful transmission (by itself or the neighbors), no
transmission, or a collision occurred. 
At each timestep, each island receives a reward of $-1$ for each message in its queue.
\<Aloha> is considerably more complex than \<FFG>. First, it 
has 3 observations per agent, which means that the number of observation histories
grows much faster. Also,
the transition model of $\Aloha$ is more densely connected than
$\FFG$: the reward component for each island is affected by the island itself and
all its neighbors. As a result,
in all the $\Aloha$ problems we consider, there is at least one
 immediate reward function whose scope contains 3 agents, i.e., $k=3$.
\fig{Aloha} illustrates the case 
with four islands in a square configuration. 
The experiments below also consider variants in which islands are connected in a line.

\begin{figure*}
\begin{centering}		
\scriptsize
\psfrag{f1}[cc][cc]{ $\utI 1$ }
\psfrag{f2}[cc][cc]{ $\utI 2$ }
\psfrag{f3}[cc][cc]{ $\utI 3$ }
\psfrag{f4}[cc][cc]{ $\utI 4$ }
\psfrag{t1t1}[cc][cc]{$ \CoJT{\langle \ffOnf_1, \ffOnf_2 \rangle }$}
\newcommand{\NRHIST}{M}
\psfrag{f2t1t1t1}[cc][cc]{ $ \CoJT{\langle 1, 		1,       1        \rangle } $ }
\psfrag{f2t1t1tn}[cc][cc]{ $ \CoJT{\langle 1, 		1,       \NRHIST  \rangle } $ }
\psfrag{f2t1tnt1}[cc][cc]{ $ \CoJT{\langle 1, 		\NRHIST, 1        \rangle } $ }
\psfrag{f2t1tntn}[cc][cc]{ $ \CoJT{\langle 1, 		\NRHIST, \NRHIST  \rangle } $ }
\psfrag{f2tnt1t1}[cc][cc]{ $ \CoJT{\langle \NRHIST, 	1,       1        \rangle } $ }
\psfrag{f2tnt1tn}[cc][cc]{ $ \CoJT{\langle \NRHIST, 	1,       \NRHIST  \rangle } $ }
\psfrag{f2tntnt1}[cc][cc]{ $ \CoJT{\langle \NRHIST, 	\NRHIST, 1        \rangle } $ }
\psfrag{f2tntntn}[cc][cc]{ $ \CoJT{\langle \NRHIST, 	\NRHIST, \NRHIST  \rangle } $ }
\psfrag{a1t1}[cc][cc]{ $ \oaHistATI{1}{t}{1}  $ }
\psfrag{a1tn}[cc][cc]{ $ \oaHistATI{1}{t}{\NRHIST} $ }
\psfrag{a2t1}[cc][cc]{ $ \oaHistATI{2}{t}{1}  $ }
\psfrag{a2tn}[cc][cc]{ $ \oaHistATI{2}{t}{\NRHIST} $ }
\psfrag{a3t1}[cc][cc]{ $ \oaHistATI{3}{t}{1}  $ }
\psfrag{a3tn}[cc][cc]{ $ \oaHistATI{3}{t}{\NRHIST} $ }
\psfrag{a4t1}[cc][cc]{ $ \oaHistATI{4}{t}{1}  $ }
\psfrag{a4tn}[cc][cc]{ $ \oaHistATI{4}{t}{\NRHIST} $ }
\psfrag{b1}[cc][cc]{ $ \polBGA{1} $ }
\psfrag{b2}[cc][cc]{ $ \polBGA{2} $ }
\psfrag{b3}[cc][cc]{ $ \polBGA{3} $ }
\psfrag{b4}[cc][cc]{ $ \polBGA{4} $ }
\includegraphics[width=1.0\columnwidth]
{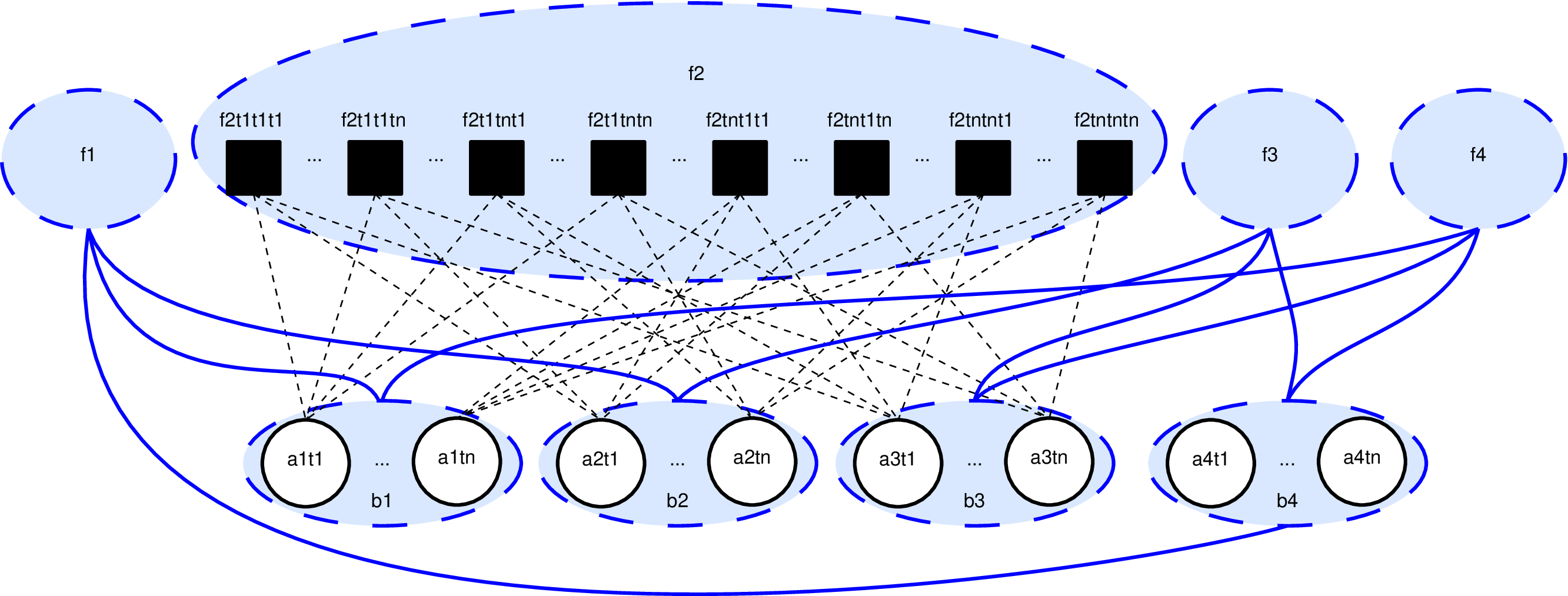}
\end{centering}
\caption{The ATI factor graph for immediate reward scopes in $\Aloha$. 
The connections of $\utI2$ are shown in detail by the black dashed lines; 
connections for other factors are summarized abstractly by the blue solid lines.}	
\label{fig:AlohaATI}
\end{figure*}

In all cases, we use immediate reward scopes that have been reduced 
(i.e., scopes that form a proper sub-scope of another scope are removed) 
before computing the factored Q-value functions. 
This means that for all stages, the factored Q-value function has the same
factorization and thus the ATI factor graphs have identical shapes 
(although the number of types differ).
For $\FFG$, the ATI factor graph for a stage $t$ is the same as that for $\GFF$ (see \fig{CGBGfactorgraph}), 
except that the types $\typeAI{i}{k}$ now correspond to action-observation 
histories~$\oaHistATI{i}{t}{k}$.
The ATI factor graph for a stage $t$ of the $\Aloha$ problem is shown in \fig{AlohaATI}.

To compute $\QTP$, the transfer-planning heuristic, for the $\FFG$ problem, we use 
 2-agent $\FFG$ as the source problem for all the edges 
and map the lower agent index in a scope 
to agent 1 and the higher index to agent 2. 
For the $\Aloha$ problem, we use the 3-island in-line variant as
the source problem and perform a similar mapping, i.e., the lowest agent index in a
scope is mapped to agent 1, the middle to agent~2
and the highest to agent~3. 
For both problems we use the $\QMDP$ and $\QBG$ 
heuristic for the source problems.

For problems small enough to solve optimally, we compare the solution quality of $\FSPCF$ to that of GMAA*-ICE,
the state-of-the-art method for optimally solving Dec-POMDPs \citep{Spaan:11:IJCAI}.
We also compare against several other approximate methods for
solving Dec-POMDPs, including non-factored $\FSPC$ and
direct cross-entropy policy search ($\DICE$) \citep{Oliehoek:08:Informatica}, one of the few 
methods demonstrated to work on Dec-POMDPs with more than three agents 
that are not transition and observation independent.
For non-factored $\FSPC$, we use alternating maximization with
10 restarts to solve the CBGs. For $\DICE$ we again use the two parameter
settings described in \sec{experiments} ($\DICEN$ and $\DICEF$).

As  baselines, we include a random joint policy and the best joint policy in which each agent selects the same fixed action for all possible histories (though the agents can select different actions from each other).  Naturally, these simple policies are suboptimal.  However, in the case of the fixed-action baseline, simplicity is a virtue.  The reason stems from a fundamental dilemma Dec-POMDP agents face about how much to exploit their private observations.  Doing so helps them accrue more local reward but makes their behavior less predictable to other agents, complicating coordination. Because it does not exploit private observations at all, the disadvantages of the fixed-action policy are partially compensated by the advantages of predictability, yielding a surprisingly strong baseline.%

We also considered including a
baseline in which the agents are allowed 
to select different actions at each timestep (but are still constrained to a fixed action 
for all histories of a given length).  
However, computing the best fixed policy of this form proved intractable.%
\footnote{In fact, the complexity of doing so is $O(|\aAS{*}^{nh})$, i.e., exponential 
in both the number of agents and the horizon. This is consistent with the complexity 
result for the non-observable problem (NP-complete) \citep{Pynadath:02:JAIR}. By searching 
for an open-loop plan, we effectively treat the problem as nonobservable.}
Note that it is not possible to use the solution to the underlying factored MDP as a baseline, for two reasons.  First, computing such solutions is not feasible for problems of the size we consider in this article.  Second, such solutions would not constitute meaningful baselines for comparison.  On the contrary, since such solutions cannot be executed in a decentralized fashion without communication, they provide only a loose upper bound on the performance possible with a Dec-POMDP, as quantitatively demonstrated by \cite{Oliehoek:08:JAIR}.

\begin{figure}[t]
\newcommand{\GMAAFvsOptfigSize}{0.485\columnwidth}
~\hfill%
\subfloat[\<FFG>.]{%
\includegraphics[width=\GMAAFvsOptfigSize]
{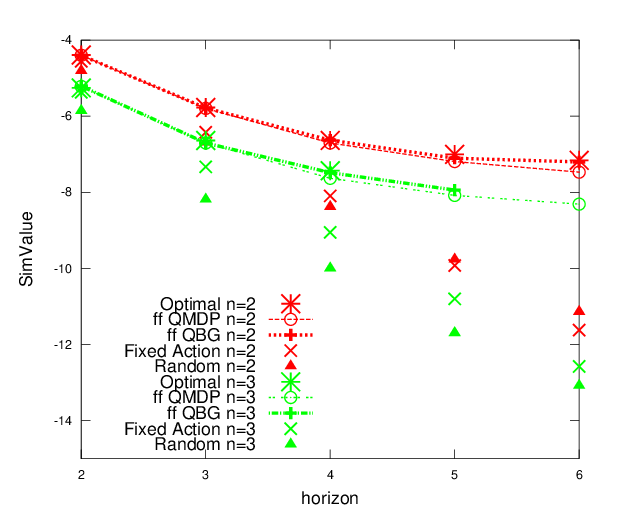}
\label{fig:GMAAvsOptimal_FFG}
}%
\hfill
\subfloat[\<Aloha>.]{%
\includegraphics[width=\GMAAFvsOptfigSize]
{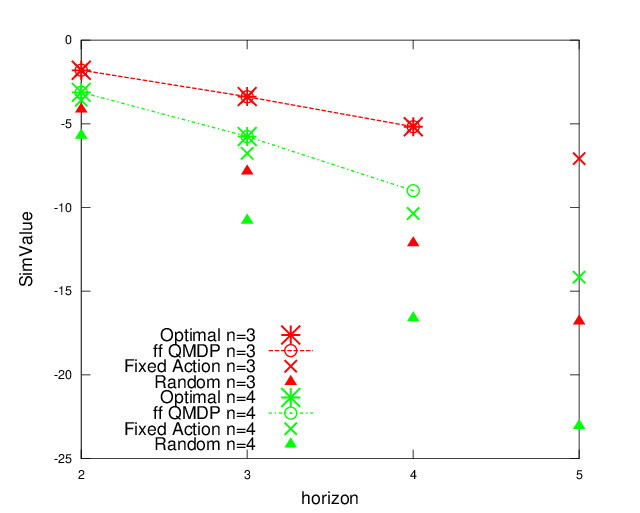}
\label{fig:GMAAvsOptimal_Aloha}
}
\hfill~
\caption{$\FSPCF$ (ff) solution quality compared to optimal and the baselines.}
\label{fig:GMAAFvsOptimal}
\end{figure}

The experimental setup is as presented in \sec{experimental-setup},
but with a memory limit of 2Gb and a maximum computation time of 1
hour. 
The reported statistics are means over 10 restarts
of each method. Once joint policies have been computed, we perform $10,000$ simulation runs to estimate their true values.

\fig{GMAAFvsOptimal} compares \<FSPCF>'s solutions to optimal solutions 
on both problems. \fig{GMAAvsOptimal_FFG} show the results for \<FFG> with two (red) and
three agents (green). Optimal solutions were computed up to horizon 6 in the former and horizon 4 in the latter problem.  \<FSPCF> with the $\QBG$ TP heuristic achieves the optimal value for all these instances. When using the $\QMDP$ TP heuristic, results are near optimal. For three agents, the optimal value is available only up to $h=4$. Nonetheless, the curve of \<FSPCF>'s values has the same shape of
as that of the optimal values for two agents, which suggests these points are near optimal as well. While the fixed action baselines performs relatively well for shorter horizons, it is worse than random for longer horizons because there always is a chance that the non-selected house will keep burning forever. 

\fig{GMAAvsOptimal_Aloha} shows results for \<Aloha>. The $\QBG$ TP heuristic is omitted since it performed the same as using $\QMDP$. For all settings at which we could compute the optimal value, \<FSPCF> matches this value.
Since the \<Aloha> problem is more complex,
\<FSPCF> has difficulty computing solutions for higher horizons. In addition, the fixed action baseline performs surprisingly well, performing optimally for 3 islands and near optimally for 4 islands.  As with \<FFG>, we expect that it would perform worse for longer horizons: if one agent sends messages for several steps in a row, its neighbor is more likely to have messages backed up in its queue.  However, we cannot test this assumption since there are no existing methods capable of solving \<Aloha> to such horizons against which to compare.

\begin{figure}[t]
\newcommand{\GMAAFvsResultsFFG}{0.45\columnwidth}
\newcommand{\GMAAFvsResultsFFGHspace}{\hspace{-7mm}}
~\hfill
\subfloat[Value for $h=5$.]{%
\includegraphics[width=\GMAAFvsResultsFFG]
{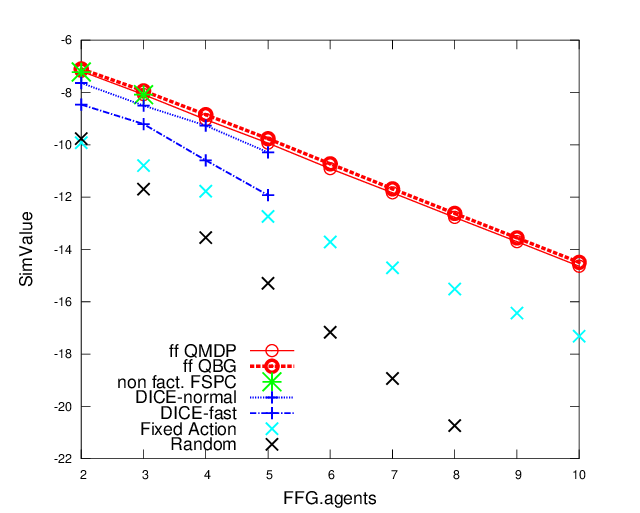}
\label{fig:GMAAvsOther_FFG:Value-h5}
}
\hfill~
\subfloat[Runtime for $h=5$.]{
\includegraphics[width=\GMAAFvsResultsFFG]
{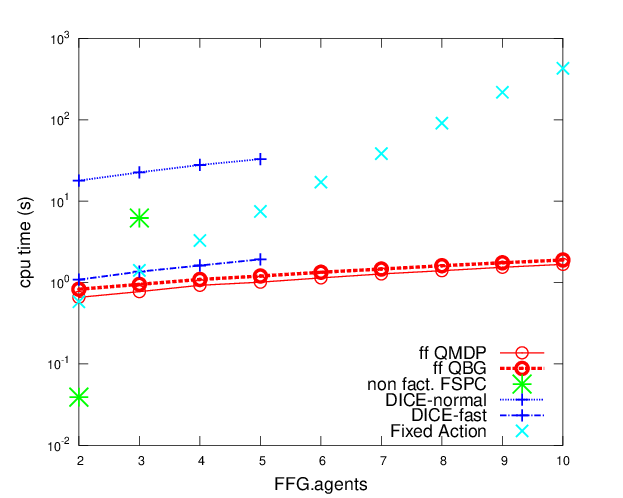}
\label{fig:GMAAvsOther_FFG:runtime-h5}
}
\hfill~

\caption{A comparison of $\FSPCF$ (ff) with different heuristics and other methods
on the $\FFG$ problem.}

\label{fig:GMAAvsOther_FFG}
\end{figure}

\fig{GMAAvsOther_FFG} compares \<FSPCF> to other approximate methods
on the \<FFG> domain with $h=5$. For all numbers of agents, \<FSPCF> finds solutions as good as or better than those of non-factored $\FSPC$, $\DICEN$, $\DICEF$, and the fixed-action and random baselines.  In addition, its running time scales much better than that of non-factored $\FSPC$ and the fixed-action baseline. Hence, this result highlights the complexity of the problem, as even a simple baseline scales poorly. \<FSPCF> also runs
substantially more quickly than $\DICEN$ and slightly more quickly than $\DICEF$, both of which run out of memory when there are more than five agents.

\begin{figure}
\newcommand{\GMAAFvsResultsAloha}{0.45\columnwidth}
\newcommand{\GMAAFvsResultsAlohaHspace}{\hspace{-7mm}}
~\hfill
\subfloat[Expected value for $h=3$.]{%
\includegraphics[width=\GMAAFvsResultsAloha]
{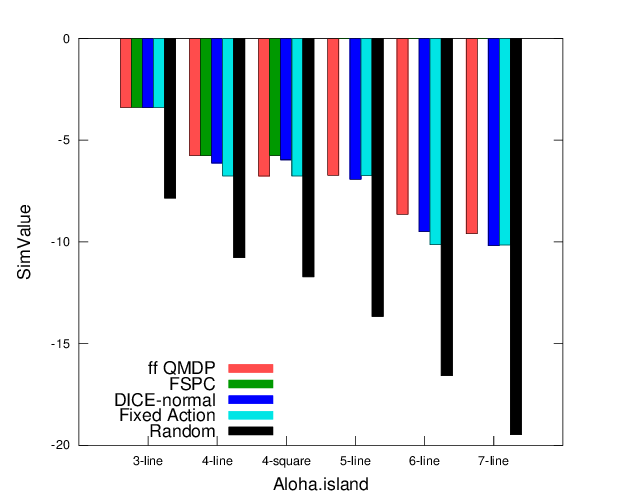}
\label{fig:GMAAvsOther_Aloha:Value-h3}
}%
\hfill
\subfloat[Runtime for $h=3$.]{
\includegraphics[width=\GMAAFvsResultsAloha]
{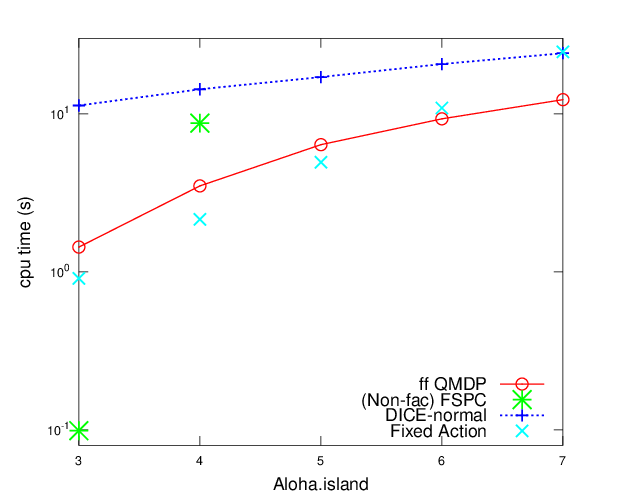}
\label{fig:GMAAvsOther_Aloha:runtime-h3}
}
\hfill~

\caption{A comparison of $\FSPCF$ with different heuristics and other methods
on the $\Aloha$ problem with $h=3$.} 

\label{fig:GMAAvsOther_Aloha}
\end{figure}

\fig{GMAAvsOther_Aloha} presents a similar comparison for the $\Aloha$ problem with $h=3$. 
$\DICEF$ is omitted from these plots because $\DICEN$ outperformed it. \fig{GMAAvsOther_Aloha:Value-h3} shows that the value achieved by $\FSPCF$ matches or nearly matches that of all the other methods on all island configurations.  \fig{GMAAvsOther_Aloha:runtime-h3} shows the runtime results for the inline 
configurations.\footnote{We omit the four agents in a square configuration in order
to more clearly illustrate how runtime in the inline configurations scales with respect to the number 
of agents.} While the runtime of $\FSPCF$ is consistently better than that of $\DICEN$, non-factored $\FSPC$ and the fixed-action baseline 
are faster for small numbers of agents. Non-factored \FSPC{} is faster 
for three agents because the problem is fully coupled: there are 3 local payoff functions involving 2, 3, and 2 agents, so k=3. Thus \<FSPCF> incurs the
overhead of dealing with multiple factors and constructing the FG but achieves no speedup in return.  However, the runtime of $\FSPCF$ scales much better as the number of agents increases.

Overall, these results demonstrate that \<FSPCF> is a substantial improvement over existing approximate Dec-POMDP methods in terms of scaling with respect to the number of agents.  However, the ability of \<FSPCF> to scale with respect to the horizon remains limited, since the number of types in the CGBGs still grows exponentially with the horizon. In future work we hope to address this problem by  clustering
types  \citep{Emery-Montemerlo:05:ICRA,Oliehoek:09:AAMAS,Wu:11:AI}. In particular, by clustering the individual types of an agent that induce similar payoff profiles  \citep{Emery-Montemerlo:05:ICRA} or probabilities over types of other agents \citep{Oliehoek:09:AAMAS} it is possible to scale to much longer horizons. When aggressively clustering to a constant number of types, runtime can be made linear in the horizon \citep{Wu:11:AI}.
However, since such an improvement  is orthogonal to the use of CGBGs, it is beyond the scope of the current article.
Moreover, we empirically evaluate the error introduced by a minimal number of approximations required to achieve scalability with respect to the number of agents. Introducing further approximations would confound these results. 

\begin{figure}
\centering
\subfloat[Expected value for $h=2,\dots,4$.]{%
\includegraphics[width=0.49\columnwidth]{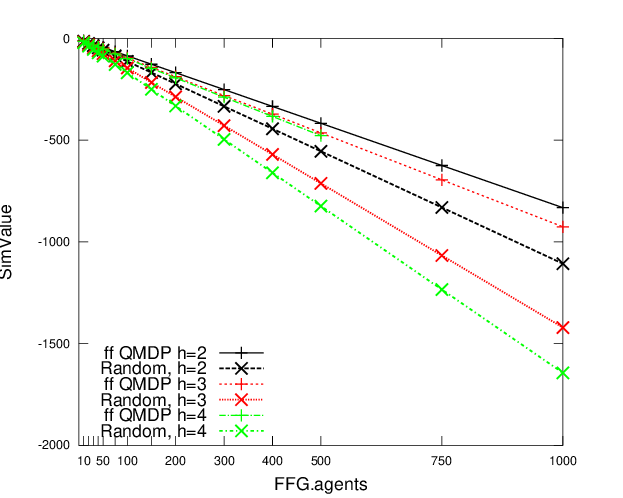}
\label{fig:ManyAgents:valueh2-4}}%
\subfloat[Expected value for $h=5,6$.]{
\includegraphics[width=0.49\columnwidth]{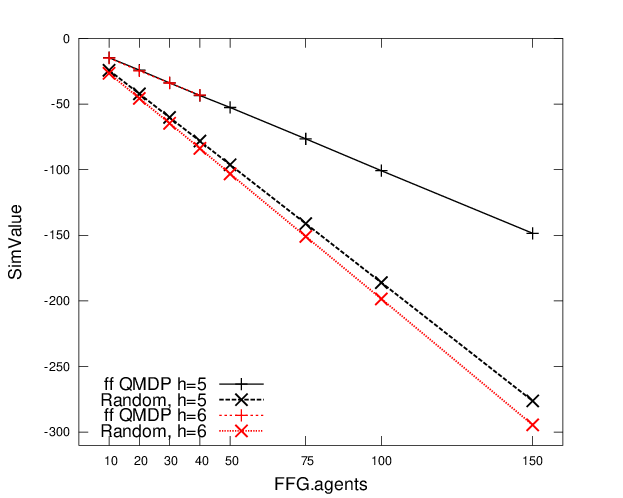}
\label{fig:ManyAgents:valueh5-6}}

\subfloat[Runtime for $h=2,\dots,6$.]{
\includegraphics[width=0.49\columnwidth]{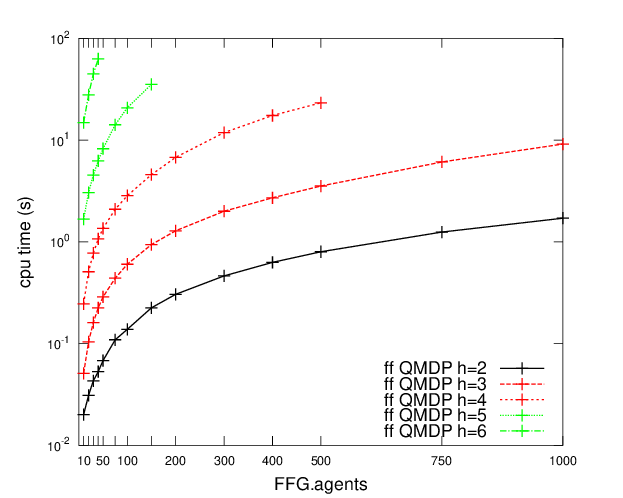}
\label{fig:ManyAgents:runtime}}
\caption{$\FSPCF$ results on $\FFG$ with many agents.}
\label{fig:ManyAgents}
\end{figure}

Nonetheless, even in its existing form, $\FSPCF$ shows great promise due to its ability to exploit both agent and type independence in the CGBG stage games.  To determine the limits of its scalability with respect to the number of agents, we conducted additional experiments applying $\FSPCF$ with the $\QMDP$ TP heuristic to \<FFG> with many more
agents.  The results, shown in \fig{ManyAgents}, do not include a fixed action
baseline because 1) performing simulations of all considered fixed action joint
policies becomes expensive for many agents, and 2) the number of such  joint
policies grows exponentially with the number of agents.

As shown in \fig{ManyAgents:valueh2-4}, $\FSPCF$ successfully computed solutions for up to 1000 agents for $h=2,3$ and 750 agents
for $h=4$. For $h=5$, it computed solutions for up to 300 agents; even
for $h=6$ it computed solutions for 100 agents,
as shown in \fig{ManyAgents:valueh5-6}. Note that, for the computed entries for $h=6$, the expected value is
roughly equal to $h=5$. This implies that the probability of any fire
remaining at stage $\ts=5$ is close to zero, a pattern we also observed for the optimal solution in \fig{GMAAFvsOptimal}.
As such, we expect that the found solutions for these settings with many agents are in fact close to optimal.
The runtime results, shown in \fig{ManyAgents:runtime},
increase linearly with respect to the number of agents. While the runtime increases with the number of agents, the bottleneck in our experiments preventing even further scalability was insufficient memory, not computation time.

These results are a large improvement in the state of the art with respect to scalability in
the number of agents. Previous approaches for general Dec-POMDPs scaled only to 3
\citep{Oliehoek:08:AAMAS} or 5 agents \citep{Oliehoek:08:Informatica}.\footnote{However, in
very recent work, \citet{Wu:10:UAI} present results for up to 20 agents on
general Dec-POMDPs.}
  Even
when making much stricter assumptions such as transition and observation
independence, previous approaches have not scaled beyond 15 agents
\citep{Varakantham:07:AAMAS,Marecki:08:AAMAS,Varakantham:09:ICAPS,Kumar:09:AAMAS}.

Though these experiments evaluate only the complete $\FSPCF$ method, in \citep{Oliehoek:10:PhD} we have empirically evaluated each of its component approximations separately.  For the sake of brevity, we do not present those experiments in this article.  However, the results confirm that each approximation is reasonable.  In particular, they show that 1) approximate inference has no significant influence on performance, 2) $\MP$ solves CGBGs as well as or better than alternating maximization, 3) the use of  1-step back-projected scopes (i.e., scopes grown by one projection back in the DBN) can sometimes slightly outperform the use of immediate reward scopes, 4) there is not a large performance difference when using optimal scopes, and 5) depending on the heuristic, allowing backtracking can improve performance.

\section{Related Work}
\label{sec:relatedWork}

The wide range of research related to the work presented in this article can be mined for various alternative strategies for solving CGBGs.  In this section, we briefly survey these alternatives, which fall into two categories: 1) converting CGBGs to other types of games, 2) converting them to constraint optimization problems. We also discuss the relation to the framework of action-graph games \citep{Jiang:08:TR}.

The first strategy is to convert the CGBG to another type of game.  In particular, CGBGs can be converted to CGSGs in the same way CBGs can be converted to SGs.  These CGSGs can be modeled using interaction hyper-graphs or factor graphs such as those shown in \fig{EdgeDecomposition} and solved by applying NDP or $\<MP>$.  However, since this approach does not exploit type independence, the size of local payoff functions scales exponentially in the number of types, making it impractical for large problems. In fact, this approach corresponds directly to the methods, tested in \sec{experiments}, that exploit only agent independence (i.e., NDP-AI and MaxPlus-AI).  The poor performance of these methods in those experiments underscores the disadvantages of converting to CGSGs.

CGBGs can also be converted to non-collaborative graphical SGs, for which a host of solution algorithms have recently emerged \citep{Vickrey:02:AAAI,Ortiz:03:NIPS,Daskalakis:06:EC}. However, to do so, the CGBG must first be converted to a CGSG, again forgoing the chance to exploit type independence.  Furthermore, in the resulting CGSG, all the payoff functions in which a given agent participates must then be combined into an individual payoff function.  This process, which corresponds to converting from an edge-based decomposition to an agent-based one, results in the worst case in yet another exponential increase in the size of the payoff function \citep{Kok:06:JMLR}.

Another option is to convert the CGBG into a non-collaborative graphical BG \citep{Singh:04:EC} by combining the local payoff functions into individual payoff functions directly at the level of the Bayesian game. Again, this may lead to an exponential increase in the size of the payoff functions.
Each BNE in the resulting graphical BG corresponds to a local optimum of the original CGBG. \citet{Soni:07:AAMAS} recently proposed a solution method for graphical BGs, which can then be applied to find locally optimal solutions. However, this method converts the graphical BG into a CGSG and thus suffers an exponential blow-up in the size of the payoff functions, just like the other conversion approaches.

The second strategy is to cast the problem of maximization over the CGBG's factor graph into a \emph{(distributed) constraint optimization problem ((D)COP)} \citep{Modi:05:AI}.  As such, any algorithm, exact or approximate, for (D)COPs
can be used to find a solution for the CGBG \citep{Liu:95:ICMAS,Yokoo:01,Modi:05:AI,Pearce:07:IJCAI,Marinescu:09:AI}. \citet{Oliehoek:10:AAMAS} propose a heuristic search algorithm for CBGs that uses this approach and exploits type independence (additivity of the value function) at the level of joint types.  \citet{Kumar:10:AAMAS} employ state-of-the-art methods for weighted constraint
satisfaction problems to instances of CBGs in the context of solving Dec-POMDPs.  The Dec-POMDPs are solved backwards using dynamic programming, resulting in CBGs with few types but many actions.  This approach exploits type independence but has been tested only as part of a Dec-POMDP solution method.  Thus, our results in \sec{FactDPOMDPs:Experiments} provide additional confirmation of the advantage of exploiting type independence in Dec-POMDPs, while our results in \sec{experiments} isolate and quantify this advantage in individual CGBGs.  Furthermore, the approach presented in this article differs from both these alternatives in that it makes the use of type independence explicit and simultaneously exploits agent independence as well.

Finally, our work is also related to the framework of action-graph games (AGGs),
which was recently extended to handle imperfect information and can model
any BG \citep{Jiang:10:NIPS}.  This work proposes two solution methods for
general-sum Bayesian AGGs (BAGGs): 
the Govindan-Wilson algorithm and simplicial subdivision.
Both involve computation of the expected payoff of each agent given a
current profile as a key step in an inner loop of the algorithm.
\citet{Jiang:10:NIPS} show how this expected payoff can be computed efficiently
for each (agent, type)-pair, thereby exploiting type (and possibly agent)
independence in this inner loop. As such, this approach may compute a sample
Nash equilibrium more efficiently than without using this structure.  

On the
one hand, BAGGs are more
general than CGBGs since they additionally allow representation of
context-specific independence and anonymity. Furthermore, the solution method is more
general since it works for general-sum games. On the other hand, in the context of collaborative games, a sample Nash
equilibrium is not guaranteed to be a PONE (but only a local optimum).  
In contrast, we solve for the global optimum and thus a PONE.  
In addition, their approach does not exploit the synergy that
independence brings in the identical payoff setting.  In contrast, NDP and
$\<MP>$, by operating directly on the factor graph, exploit independence not just
within an inner loop but throughout the computation of the solution. 
Finally, note that our algorithms also work for collaborative
BAGGs that possess the same form of structure as CGBGs (i.e., agent and type independence). 
In cases where there is no anonimity or context-specific independence (e.g., as in the 
CGBGs generated for Dec-POMDPs), the BAGG framework offers no advantages.

\section{Future Work}
\label{sec:future}

The work presented in this article opens several interesting avenues for future work.  A straightforward extension of our methods would replace $\<MP>$ with more recent message passing algorithms 
for belief propagation that are
guaranteed to converge \citep{Globerson:08:NIPS07}. Since these algorithms are guaranteed to compute the exact MAP
configuration for perfect graphs with binary variables \citep{Jebara:09:UAI}, the resulting approach would be able to efficiently compute optimal solutions for CGBGs with two actions and
perfect interaction graphs.

Another avenue would be to investigate whether our algorithms can be extended to work on
a broader class of collaborative Bayesian AGGs. Doing so could enable our approach to also exploit context-specific independence and anonymity. A complementary idea is to extend our algorithms to
the non-collaborative case by rephrasing the task of finding a sample Nash
equilibrium as one of minimizing regret, as suggested by
\citet{Vickrey:02:AAAI}.

Our approach to solving Dec-POMDPs with CGBGs could be integrated with methods for clustering histories \citep{Emery-Montemerlo:05:ICRA,Oliehoek:09:AAMAS} to allow
scaling to larger horizons.  In addition, there is great potential for further improvement in the accuracy and efficiency of computing approximate value functions.  In particular, the transfer planning approach could be extended to transfer tasks with different action and/or observation spaces, as done in transfer learning \citep{Taylor:07:JMLR}.  Furthermore, it may be possible to automatically identify suitable source tasks and mappings between tasks, e.g., using qualitative DBNs \citep{Liu:06:AAAI}.

\section{Conclusions}
\label{sec:conclusions}

In this article, we considered the interaction of several agents under
uncertainty. In particular, we focused on settings in which multiple collaborative agents, each possessing some private information, must coordinate their actions. Such settings can be formalized by the Bayesian game framework. We presented an overview of game-theoretic models used for collaborative decision
making and delineated two different types of structure in
collaborative games: 1) agent independence, and 2) type independence.

Subsequently, we proposed the \emph{collaborative graphical Bayesian game (CGBG)}
as a model that facilitates more efficient decision making by decomposing the global payoff function as the sum of local payoff functions that depend on only a few agents. We showed how CGBGs can be represented as factor graphs (FGs) that capture both agent and type independence. Since a maximizing configuration of the factor graph corresponds to a solution of the CGBG, this representation also makes it possible to effectively exploit this independence.

We considered two solution methods: non-serial
dynamic programming (NDP) and \<MP> message passing. The former has a computational complexity that is exponential in the 
induced tree width of the FG, which we proved to be exponential in the number of individual types. The latter is tractable when there is enough independence between agents; we showed that it is exponential only in $k$, the maximum number of agents that participate in the same local payoff function.
An empirical evaluation showed that exploiting both agent and type
agent independence can lead to a large performance increase, compared to
exploiting just one form of independence, without sacrificing solution quality.  For example, the experiments showed
that this approach allows for the solution of coordination problems with imperfect information for up to 750 agents, limited only by a 1GB memory constraint.

We also showed that CGBGs and their solution methods provide a key missing component in the approximate solution of Dec-POMDPs with many agents. In particular, we proposed $\FSPCF$, which approximately solves Dec-POMDPs by
representing them as a series of CGBGs.  To estimate the payoff functions of these CGBGs, we computed approximate factored
value functions given predetermined scope structures via a 
method we call \emph{transfer planning}. It uses value functions for smaller source problems as components of the
factored Q-value function for the original target problem.
An empirical evaluation showed that \<FSPCF> significantly outperforms 
state-of-the-art methods for solving Dec-POMDPs with more than two agents and scales
well with respect to the number of agents. 
In particular, \<FSPCF> found (near-)optimal solutions on problem instances for which the optimum can be computed.
For larger problem instances it found solutions as good as or better than comparison Dec-POMDP methods in almost all cases
and in all cases outperformed the baselines.
The most salient result from our experimental evaluation is that the proposed method is able to compute
solutions for problems that cannot be tackled by any other methods at all (not even the baselines).
In particular, it found good
solutions for up to 1000 agents, where previously only problems
with small to moderate  numbers of agents (up to 20) had been 
tackled.

\section*{Acknowledgements}
We would like to thank Nikos Vlassis for extensive discussions on the
topic, and Kevin Leyton-Brown, David Silver and Leslie Kaelbling for
their valuable input.
This research was partly performed under the Interactive
Collaborative Information Systems (ICIS) project, supported by the
Dutch Ministry of Economic Affairs, grant nr: BSIK03024. 
The research is supported in part by AFOSR MURI project \#FA9550-09-1-0538.
This work was partly funded by Funda\c{c}\~{a}o para a Ci\^encia e a
Tecnologia (ISR/IST pluriannual funding) through the PIDDAC Program
funds and was supported by project PTDC/EEA-ACR/73266/2006.

\section{Appendix}
\label{appendix}

\begin{lemma}[Complexity of \<MP>{}] 
    \label{lem:MPcomplexity}
    The complexity of one iteration of $\<MP>$ is
\begin{equation}
O\left( 
m^k \mult k^2 \mult l \mult F 
\right), 
\label{eq:MPcomplexity_app}
\end{equation} 
where, 
$F$ is the number of factors, 
the maximum degree of a factor is $k$,
the maximum degree of a variable is $l$,
the maximum number of values a variable can take is $m$.

\begin{proof}
We can directly derive that 
the number of edges is bounded by $e=F \mult k$.
Messages sent by a variable are constructed by summing over incoming messages. 
As a variable has 
at most $l$ neighbors, 
this involves adding 
at most $l-1$ incoming messages of size $m$. 
The cost of constructing one message for one variable therefore is:
$ O(m \mult (l-1)). $
This means that the total cost of constructing all $e=O( F \mult k)$ messages sent by variables,
one over each edge, is 
\begin{equation}
    O \left(    m \mult (l-1) \mult F \mult k   \right) 
\label{eq:MPcomplexVars}
\end{equation}

Now we consider the messages sent by factors. Recall that the maximum size of a factor is 
$m^k$. The construction of each message entails factor-message addition 
(see, e.g., \cite{Oliehoek:10:PhD}, Sec. 5.5.3) with $k-1$ incoming messages, 
each one has cost $O(m^k)$. 
This leads to a cost of
\[ O( (k-1) \mult m^k) = O( k  \mult  m^k )\]
per factor message, and a total cost of
\begin{equation}
O( F \mult k^2 \mult m^k ).
\label{eq:MPcomplexFactors}
\end{equation} 
The complexity of a single iteration of \<MP>{} is the sum of
\eq{MPcomplexVars} and \eq{MPcomplexFactors}, which can be reduced to
\eq{MPcomplexity_app}.
\end{proof}
\end{lemma}

{
\small
\bibliographystyle{abbrvnat}
\bibliography{bib}
}

\end{document}

%% file: preamble_U14.tex
\usepackage{mathptmx}
\usepackage{eucal}
\usepackage{authblk}
\usepackage[scale=0.72]{geometry}
\usepackage{verbatim}
\usepackage{float}
\usepackage{graphicx}
\usepackage{amsmath}
\usepackage{amssymb}
\usepackage{amsbsy}
\usepackage{amsthm}

\usepackage[hang,tabbotcap]{subfigure}
\let\subfloat\subfigure

\usepackage{colortbl}
\usepackage{multirow}
\usepackage{hhline}

\usepackage[round]{natbib}

\usepackage{algorithmic}

\usepackage{psfrag}
\usepackage{longtable}
\usepackage{verbatim}
\usepackage{icomma} 

\numberwithin{equation}{section}

\graphicspath{{figures/}, {figs/}}

\theoremstyle{plain}
\newtheorem{theorem}{Theorem}[section]
\newtheorem{lemma}{Lemma}[section]
\newtheorem{corollary}{Corollary}

\theoremstyle{plain}

\theoremstyle{definition}
\newtheorem{definition}{Definition}[section]
\newtheorem{example}{Example}[section]
\let\oex\endexample
\renewcommand{\endexample}{\qed \endexample}
\renewcommand{\endexample}{\qed \oex}

\theoremstyle{definition}
\newtheorem{defn}{Definition}[section]

\newtheorem{observation}{Observation}
\theoremstyle{remark}

\renewcommand{\sec}[1]{Sec.~\ref{sec:#1}}
\newcommand{\eq}[1]{\eqref{eq:#1}}

\newcommand{\fig}[1]{Fig.~\ref{fig:#1}}

\newcommand{\tab}[1]{Table~\ref{tab:#1}}

\newcommand{\lem}[1]{Lemma~\ref{lem:#1}}
\newcommand{\thm}[1]{Theorem~\ref{thm:#1}}

\providecommand{\sfe}{\ed} 
\providecommand{\age}{\ed}  

\providecommand{\problemName}[1]{\textsc{#1}}
\providecommand{\FFGfull}{\problemName{Sequential Fire Fighting}}
\providecommand{\GFF}{\problemName{Generalized Fire Fighting}}
\providecommand{\FFG}{\FFGfull}

\providecommand{\Aloha}{\problemName{Aloha}}

\newcommand{\labsize}{\footnotesize}
\psfrag{SimValue}{\labsize Value}
\psfrag{cpu time (s)}{\labsize time (s)}
\psfrag{CPUTimeGMAA}{\labsize\hspace{-2.5em}$\GMAA$ Time(s)}
\psfrag{horizon}[Bc][Bc]{\labsize Horizon}
\psfrag{FFG.agents}[Bc][Bc]{\labsize Number of agents}
\psfrag{Aloha.island}[Bc][Bc]{\labsize Island configuration}

\def\<#1>{%
    \expandafter\ifx\csname<#1>\endcsname\relax
        \errmessage{abbreviation <#1> undefined!}
    \else
        \csname<#1>\endcsname
    \fi
}
\def\abbr#1#2{%
    \expandafter\def\csname<#1>\endcsname{#2}%
}
\abbr{dectig}{Dec-Tiger}
\abbr{sdectig}{Skewed Dec-Tiger}
\abbr{AOH}{action-observation history}
\abbr{JAOH}{joint action-observation history}
\abbr{GFF}{\problemName{Generalized Fire Fighting}}
\abbr{FFG}{\problemName{Sequential Fire Fighting}}
\abbr{Aloha}{\problemName{Aloha}}
\abbr{FSPCF}{\FSPCF}

\providecommand{\tabularnewline}{\\}
\floatstyle{ruled}
\newfloat{algorithm}{tbp}{loa}
\floatname{algorithm}{Algorithm}

\providecommand{\BAGABAB}{\algName{BaGaBaB}}

\providecommand{\ourvec}[1]{\bar{#1}}

\providecommand{\Co}{C}
\providecommand{\CoJT}[1]{\Co_{#1}}
\providecommand{\CoJTGE}[2]{\Co_{#1}^{#2}}

\providecommand{\bSymbol}{\boldsymbol{b}}

\providecommand{\bO}{\bSymbol^0} 

\providecommand{\agentS}{\mathcal{D}}
\providecommand{\agentI}[1]{{#1}}
\providecommand{\nrA}{n} 

\providecommand{\sS}{\mathcal{S}} 
\providecommand{\s}{s} 

\providecommand{\nrSF}{|\mathcal{X}|} 

\providecommand{\factorSymb}{x}
\providecommand{\factorSetSymb}{\mathcal{X}}

\providecommand{\sfacS}{\factorSetSymb}
\providecommand{\sfacI}[1]{\factorSymb_{#1}} 
\providecommand{\sfacG}[1]{{\mathbf\factorSymb}_{#1}} 
\providecommand{\sfacIS}[1]{\factorSetSymb_{#1}} 
\providecommand{\sfacGT}[2]{{\mathbf\factorSymb}_{#1}^{#2}} 

\providecommand{\agScopeSymb}{\mathbb{A}}    

\providecommand{\agSC}[1]{\agScopeSymb(#1)}    

\providecommand{\ed}{e}     
\providecommand{\edS}{\mathcal{E}}     
\providecommand{\nrEd}{\nrR}

\providecommand{\nrR}{\rho} 
\providecommand{\nrRA}[1]{\nrR_{#1}} 
\providecommand{\REd}[1]{R^{#1}} 
\providecommand{\QEd}[1]{Q^{#1}} 
\providecommand{\QEdP}[2]{Q^{#1}_{#2}} 

\providecommand{\ts}{t}             

\providecommand{\aA}[1]{a_{#1}} 	
\providecommand{\aAS}[1]{\mathcal{A}_{#1}} 	
\providecommand{\aAT}[2]{\aA{#1}^{#2}} 	
\providecommand{\oA}[1]{o_{#1}} 	
\providecommand{\oAS}[1]{\mathcal{O}_{#1}} 	
\providecommand{\oAT}[2]{\oA{#1}^{#2}} 	

\providecommand{\jo}{\mathbf{o}}    
\providecommand{\joS}{\mathcal{O}}    

\providecommand{\ja}{\mathbf{a}}    
\providecommand{\jaS}{\mathcal{A}}    
\providecommand{\jaG}[1]{\ja_{#1}}  
\providecommand{\jaGT}[2]{\jaG{#1}^{#2}}  

\providecommand{\oaHist}{\ourvec{\btheta}}

\providecommand{\oaHistT}[1]{\ourvec{\btheta}{}^{#1}}

\providecommand{\oaHistAT}[2]{\ourvec{\theta}_{#1}^{\,#2}}

\providecommand{\oaHistATI}[3]{\oaHistAT{#1,#3}{#2}}

\providecommand{\oaHistGT}[2]{\oaHist{}_{#1}^{#2}}

\providecommand{\oHistAT}[2]{\ourvec{o}_{#1}^{\,#2}}
\providecommand{\oHistATS}[2]{\ourvec{\mathcal{O}}_{#1}^{#2}}

\def\bpi{\boldsymbol{\pi}} 

\def\btheta{\boldsymbol{\theta}}
\def\bTheta{\boldsymbol{\Theta}}

\providecommand{\jpol}{\bpi}	
\providecommand{\polA}[1]{\pi_{#1}}	
  
\providecommand{\bdelta}{{\boldsymbol{\delta}}}

\providecommand{\drAT}[2]{\delta_{#1}^{#2}}
\providecommand{\jdrT}[1]{\bdelta^{#1}}

\def\bvarphi{{\boldsymbol{\varphi}}}
\providecommand{\partJPol}{\bvarphi}

\providecommand{\pP}{\partJPol}	
	
\providecommand{\pJPolT}[1]{\pP^{#1}}

\providecommand{\Q}{Q}	

\providecommand{\jtype}{\btheta}
\providecommand{\jtypeG}[1]{\btheta_{#1}}

\providecommand{\jtypeS}{\bTheta}
\providecommand{\typeA}[1]{\theta_{#1}}
\providecommand{\typeAI}[2]{\typeA{#1}^{#2}} 
\providecommand{\typeAS}[1]{\Theta_{#1}}
\providecommand{\utF}{u}		
\providecommand{\utA}[1]{\utF_{#1}}	
\providecommand{\utI}[1]{\utF^{#1}}	
\providecommand{\utS}{\mathcal{U}}

\def\bbeta{\boldsymbol{\beta}}

\providecommand{\polBG}[1]{\beta_{#1}}
\providecommand{\polBGA}[1]{\polBG{#1}} 
\providecommand{\jpolBG}{ {\bbeta} }

\providecommand{\jpolBGG}[1]{\jpolBG_{#1}} 

\providecommand{\QBG}{\textrm{Q}_{\textrm{BG}}}

\providecommand{\QMDP}{\textrm{Q}_{\textrm{MDP}}} 
\providecommand{\QPOMDP}{\textrm{Q}_{\textrm{POMDP}}}

\providecommand{\TP}{\algName{TP}}
\providecommand{\QTP}{\textrm{Q}_{\TP}}

\providecommand{\DICE}{{\algName{Dice}}} 
 
\providecommand{\MAA}{{\algName{MAA}^{*}}} 
 
\newcommand{\FSPC}{\algName{FSPC}} 
\providecommand{\GMAAfull}{\text{generalized }\algName{MAA}^{*}} 
\providecommand{\GMAA}{\GMAAfull}

\providecommand{\FSPCF}{{\algName{Factored } \FSPC}}

\providecommand{\MP}{{\algName{Max-Plus}}}
\abbr{MP}{\algName{Max-Plus}}

\providecommand{\CE}{{\algName{CE}}}
\providecommand{\CEnormal}{{\algName{CEnormal}}}
\providecommand{\CEfast}{{\algName{CEfast}}}
\providecommand{\BAGABAB}{{\algName{BaGaBaB}}}
\providecommand{\AltMax}{{\algName{AltMax}}}

\providecommand{\tpSP}{\text{s}}

\providecommand{\house}{H}
\providecommand{\nrHouses}{{N_{\house}}}
\providecommand{\nrFLs}{{N_{f}}}
\providecommand{\fl}[1]{\sfacI{#1}}
\providecommand{\flG}[1]{\sfacG{#1}}
\providecommand{\ffA}[1]{{\house_#1}}
\providecommand{\ffOfl}{F}
\providecommand{\ffOnf}{N}

\providecommand{\GFFdensity}{{N_{d}}}

\providecommand{\ffnrAcs}{{N_A}}
\providecommand{\ffnrObs}{{N_O}}

\providecommand{\argmax}{\operatornamewithlimits{arg\,max}}
\providecommand{\defas}{\ \operatorname{\equiv}\ }

\providecommand{\reals}{\mathbb {R} }

\providecommand{\algName}[1]{\textsc{#1}}

\providecommand{\fig}{Figure~}
\providecommand{\lem}{Lemma~}
\providecommand{\thm}{Theorem~}

\providecommand{\tab}{Table~}

\providecommand{\excl}[1]{\neq{#1}}

\providecommand{\mult}{\cdot}

\providecommand{\eps}{\epsilon} 

\newlength{\thickertableline}
\definecolor{tabledark}{rgb}{0.25, 0.25, 0.60}
\definecolor{tablemid}{rgb}{0.80, 0.80, 0.95}
\definecolor{tablelight}{rgb}{0.90, 0.90, 1.0}

\definecolor{examtablelight}{rgb}{0.95,1.0,1.0}
\definecolor{examtabledark}{rgb}{0.2,0.4,0.2}